\documentclass[]{fairmeta}
\usepackage[table,xcdraw,dvipsnames]{xcolor}
\usepackage{wrapfig}
\usepackage{mathpazo}
\usepackage{tgpagella}
\usepackage{bbding}
\usepackage{graphicx}
\usepackage{pgfplots}
\usepackage{makecell}
\usepackage{enumitem}
\usepackage{mathtools}
\usepackage{xfrac}
\usepackage{tikz}
\usetikzlibrary{spy}

\newcolumntype{L}[1]{>{\raggedright\let\newline\\\arraybackslash\hspace{0pt}}m{#1}}
\newcolumntype{C}[1]{>{\centering\let\newline\\\arraybackslash\hspace{0pt}}m{#1}}
\newcolumntype{R}[1]{>{\raggedleft\let\newline\\\arraybackslash\hspace{0pt}}m{#1}}
\newcolumntype{Y}{>{\centering\arraybackslash}X}

\title{Scaling Sequence-to-Sequence Generative\\ Neural Rendering}
\author[1,\ast]{Shikun Liu}
\author[1,\ast]{Kam Woh Ng}
\author[1]{Wonbong Jang}
\author[1]{Jiadong Guo}
\author[1]{Junlin Han}
\author[1]{Haozhe Liu}
\author[1]{\\Yiannis Douratsos}
\author[1]{Juan~C.~Pérez}
\author[1]{Zijian Zhou}
\author[1]{Chi Phung}
\author[1]{Tao Xiang}
\author[1]{Juan-Manuel~Pérez-Rúa}

\affiliation[1]{Meta AI}
\contribution[\ast]{Core Contributors}

\abstract{
We present Kaleido, a family of generative models designed for photorealistic, unified object- and scene-level neural rendering. Kaleido operates on the principle that 3D can be regarded as a specialised sub-domain of video, expressed purely as a sequence-to-sequence image synthesis task.
Through a systemic study of scaling sequence-to-sequence generative neural rendering, we introduce key architectural innovations that enable our model to: i) perform generative view synthesis without explicit 3D representations; ii) generate any number of 6-DoF target views conditioned on any number of reference views via a masked autoregressive framework; and iii) seamlessly unify 3D and video modelling within a single decoder-only rectified flow transformer.
Within this unified framework, Kaleido leverages large-scale video data for pre-training, which significantly improves spatial consistency and reduces reliance on scarce, camera-labelled 3D datasets --- all without any architectural modifications.
Kaleido sets a new state-of-the-art on a range of view synthesis benchmarks. Its zero-shot performance substantially outperforms other generative methods in few-view settings, and, for the first time, matches the quality of per-scene optimisation methods in many-view settings.

}

\date{\today}
\correspondence{\email{shikun@meta.com} (model design, training), \email{kamwohng@meta.com} (data rendering, inference)}
\metadata[Blogpost]{\url{https://shikun.io/projects/kaleido}}

\begin{document}

\maketitle

\begin{figure}[ht!]
    \centering
    \includegraphics[width=\linewidth]{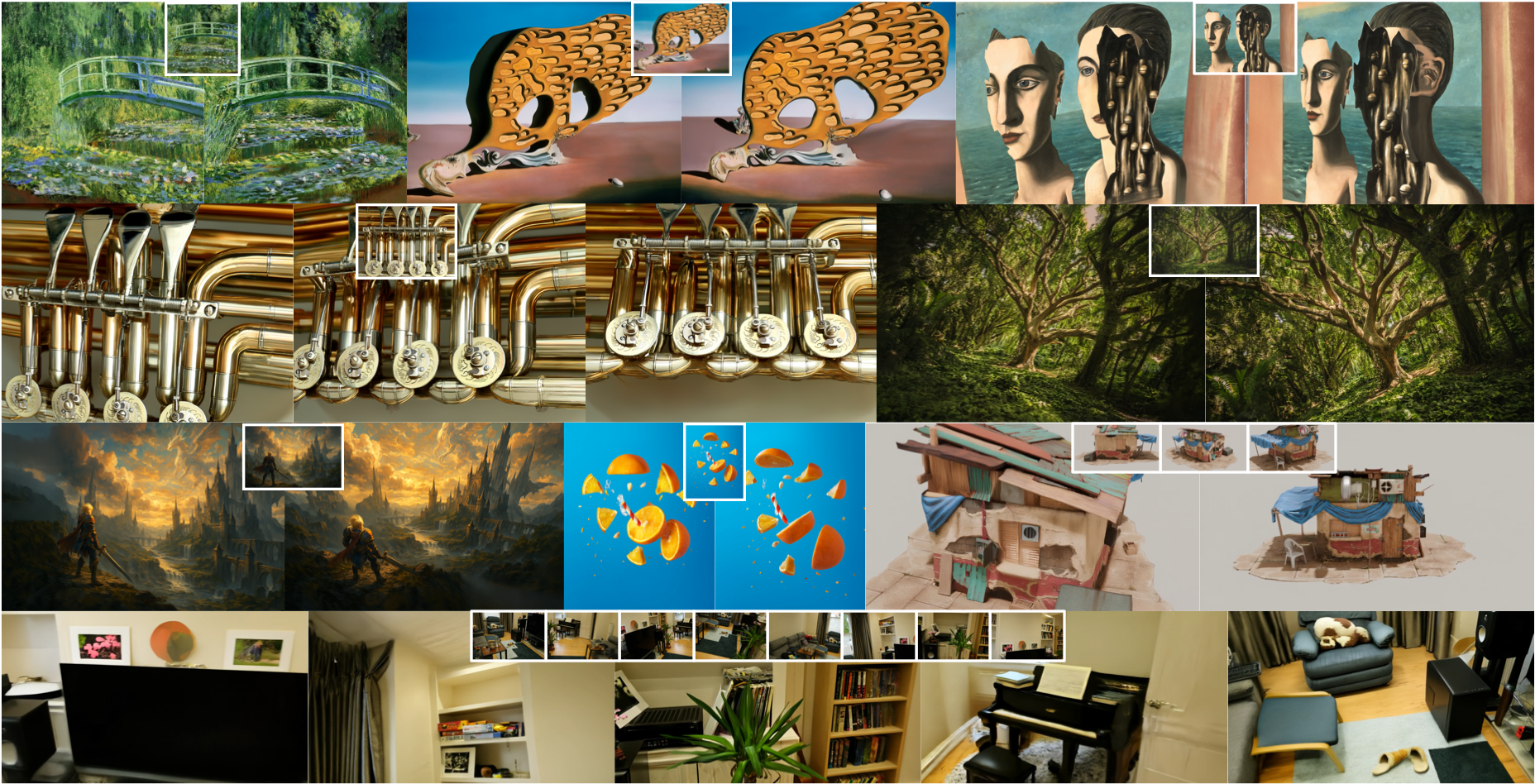}
    \caption{{\it Kaleido} is a generative rendering engine that can synthesise any number of photorealistic novel views across diverse artistic styles from any number of reference images (white boxes) with arbitrary 6-DoF camera poses. }
    \label{fig:highlight}
\end{figure}

\section{Introduction}
Rendering and view synthesis are foundational to 3D computer vision and graphics, driving applications across virtual reality, cinematic effects, robotics and autonomous driving. By allowing a scene to be rendered from arbitrary viewpoints based on a limited set of reference views, view synthesis mimics the adaptability of human vision — the ability to construct and reconstruct a coherent 3D understanding of our surroundings.

While deep learning, fuelled by massive datasets and scalable architecture designs, has achieved remarkable success in language modelling and 2D vision, its progress in 3D vision for {\it general-purpose rendering} has been comparatively slow. We argue this stems from two persistent and interconnected bottlenecks:

\begin{enumerate}
    \item {\it A Fragmented Landscape of 3D Representations.} 3D vision lacks a consensus on the {\it right 3D representation}, with methods spanning explicit structures like voxels \citep{wu20153dshapenet} and point clouds \citep{qi2017pointnet,guo2020pointcloud_survey} to implicit ones like neural fields \citep{mildenhall2020nerf,xie2022neuralfield_survey}. This fragmentation has prevented the focused, collective effort required to scale a powerful architecture for any single representation, as development remains divided across incompatible data formats.
    \item {\it The High Cost of 3D Data.} 3D datasets are scarce and difficult to obtain primarily because their creation is guided by the principle of {\it strict 3D consistency}. Achieving this level of precision requires either hand-crafting 3D synthetic object meshes \citep{deitke2023objaverse,deitke2023objaversexl} or employing bundle adjustment and global alignments \citep{hartley2003mvg} for slow multi-view camera labelling, making the data acquisition process slow, costly, and fundamentally difficult to scale.
\end{enumerate}

As a direct consequence of these challenges, the 3D vision community has yet to converge on a scalable paradigm for 3D modelling. The combination of fragmented research efforts and restrictive data requirements has prevented the kind of focused, large-scale investment that enabled the dramatic architectural scaling and performance gains seen in language and 2D vision.

We believe these limitations, taken together, point to a fundamental oversight:
\begin{center}
{\it 3D perception is not a geometric problem, but a form of visual common sense.}
\end{center}

The human ability to perceive 3D structure emerges from extensive observation of the world, not from maintaining a precise 3D model in the mind. For example, humans can interpret 3D geometry in optical illusions ({\it e.g.} in M.C. Escher’s impossible structures and the Ponzo illusion), without having a physically accurate 3D representation or even a correct sense of depth. Accordingly, we argue that an ideal rendering system should not aim to {\it explicitly model perfect geometric consistency}, but instead to learn an {\it implicit representation} by capturing the statistical patterns of the extensive visual experience of the world. 

Building on this insight, we introduce {\it Kaleido}, a scalable architecture for generative neural rendering. We design Kaleido as a type of {\it spatial generative model} that does not encode any explicit 3D structures. Instead, Kaleido inherits spatial perception and visual common sense directly from large-scale video data, purely in a data-driven way, similar to how modern large language models acquire textual common sense from large-scale corpora without relying on explicit linguistic rules. This leads to our central hypothesis, inspired by the success of domain-specific fine-tuning in pre-trained language models (e.g., coding in \cite{roziere2023code_llama,anil2023palm2,chen2021codex_openai}), we believe that a powerful general-purpose rendering model can be created by treating {\it 3D as a specialised sub-domain of video}. To put it simply,
\begin{align*}
    &\text{{\it We observed}} &\text{\it large-scale corporus data}  \to \text {\it structured code data} &= \text{\it a general-purpose coding model} \\
    \Longrightarrow\quad  &\text{{\it We hypothesise}} &\text{\it large-scale video data}  \to \text {\it structured 3D data} &= \text{\it a general-purpose rendering model}
\end{align*}

\begin{figure}[t!]
    \centering
    \includegraphics[width=\linewidth]{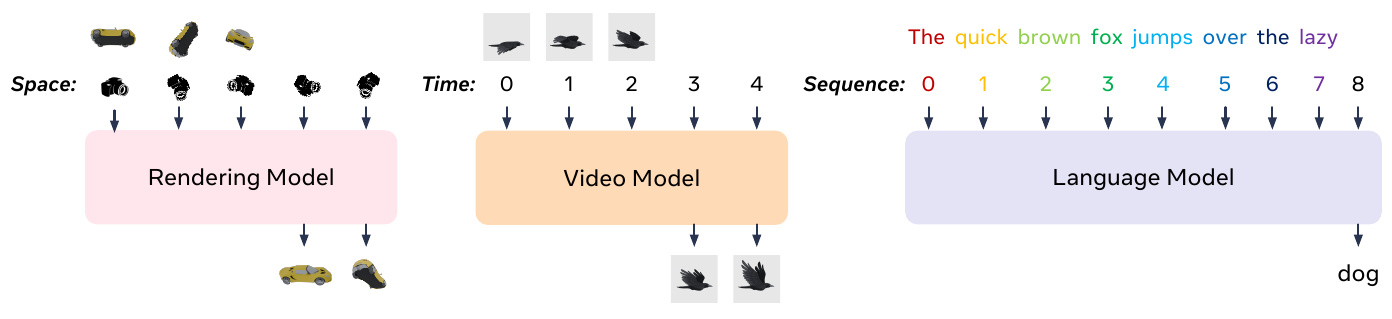}
    \caption{\textbf{Rendering as Sequence-to-Sequence Image Modelling.} We propose that neural rendering can be framed as a sequence-to-sequence task, unifying its design with language and video generation. In this formulation, a transformer \citep{vaswani2017transformer} learns to generate image tokens conditioned on their spatial positions, similar to how language models condition on token positions in a sequence, and video models condition on temporal positions across frames.}
    \label{fig:teaser}
\end{figure}

To realise this hypothesis, we reformulate rendering as a sequence-to-sequence problem, specifically as a pose-conditioned, image-to-image synthesis task. We first establish {\it a unified, geometrically consistent representation of space and time} as the core of our model design. This is achieved with a positional encoding design that extends the original Rotary Positional Encoding (RoPE) \citep{su2021roformer} to parametrise all 2D, 3D, and temporal positions {\it relatively}, within the dot-product self-attention of a transformer \citep{vaswani2017transformer}. This foundational design enables Kaleido to learn rich world representations from large-scale, unstructured video data and then perform efficient transfer learning with much smaller-scale, structured multi-view 3D data, all within the same model {\it without any task-specific architectural changes}.

Building on this unified representation, Kaleido naturally benefits from scalable architectures and powerful generative techniques developed for language and vision. Specifically, Kaleido adopts a scalable transformer architecture inspired by Diffusion Transformer (DiT) \citep{peebles2023dit} and Llama-3 \citep{dubey2024llama3}, performing generative modelling via a rectified flow objective \citep{liu2023flow,esser2024sd3} within a masked autoregressive framework \citep{li2024mar,fan2025fluid,liu2025mardini}.

Finally, we identify that rectified flow SNR samplers commonly used for text-to-image/video generation are suboptimal for the precise pose conditioning required in rendering. We therefore introduce an improved, noise-biased sampling strategy and other key architectural adjustments to ensure stable and efficient scaling. Through extensive systemic studies, we validate these designs and highlight our primary contributions:

\begin{enumerate}
    \item We introduce the {\it Kaleido family of Spatial Generative Models (SGMs)}, which can perform unified object- and scene-level view synthesis from any number of reference views to any number of target views with full 6-DoF camera control. This is enabled by the following designs:
    \begin{enumerate}
        \item A simple decoder-only rectified flow transformer that considers generative rendering as a sequence-to-sequence task.
        \item A unified positional encoding design that seamlessly processes both 3D and video data within a single, unchanged architecture.
        \item An effective scaling recipe for both model size and resolution, supported with a tailored SNR sampler and solutions for training instability.
    \end{enumerate}
    \item Kaleido generates high-resolution images (up to 1024px) across diverse aspect ratios, achieving state-of-the-art results on numerous view synthesis and 3D reconstruction benchmarks. Most notably, in many-view settings, Kaleido is the first zero-shot generative model to match the rendering quality of per-scene optimisation methods like Instant-NGP \citep{muller2022instant}.
\end{enumerate}

\section{Related Work}

\paragraph{From 2D to 3D and Camera Parameters.} 
Reconstructing 3D geometry and camera parameters from 2D images is a foundational problem in computer vision. Classical approaches like Structure from Motion (SfM)~\citep{hartley2003mvg,schoenberger2016colmap2} and Simultaneous Localisation and Mapping (SLAM) \citep{davison2007monoslam, mur2015orb,izadi2011kinectfusion} have been highly successful, but they are limited by their need to optimise each scene from scratch and their struggles with non-overlapping views. More recently, learning-based methods have emerged to address these limitations. Models like DUSt3R \citep{wang2024dust3r} and VGGT \citep{wang2025vggt} introduce feed-forward pointmap regression, enabling end-to-end 3D reconstruction that generalises across scenes. While these methods represent a significant step forward, their reliance on direct geometric regression means they cannot effectively infer content in occluded regions. 
Notably, Kaleido’s fully generative design allows it to predict plausible, spatially consistent content for occluded regions, a key advantage over both classical and modern regression-based techniques.

\paragraph{Multi-View Stereo, Neural Rendering, and Novel View Synthesis.}
Traditional Multi-view Stereo (MVS) \citep{furukawa2015multi, schoenberger2016colmap2} reconstructs 3D surfaces by triangulating features across multiple viewpoints. This principle was revolutionised by Neural Radiance Fields (NeRF) \citep{mildenhall2020nerf}, which uses volume rendering and coordinate MLPs to achieve photorealistic novel view synthesis. A plethora of follow-up works have focused on improving the speed and quality of this per-scene optimisation paradigm \citep{muller2022instant, fridovich2022plenoxels, chen2022tensorf,kerbl20233gaussian}. However, these methods require numerous, dense input views. To handle synthesis from only a few views, a learned prior is necessary. Early works in this area used category-specific priors and pre-trained image features \citep{sitzmann2019scene,yu2021pixelnerf,jang2021codenerf}, but a performance gap remained compared to scene-specific methods with dense views. More recently, feed-forward transformer-based models have emerged \citep{hong2023lrm, jang2024nvist, jin2024lvsm}, which can directly predict 3D primitives or render novel views from limited inputs. However, as deterministic models, they still fundamentally struggle with the inherently probabilistic nature of inferring large, occluded regions.

\paragraph{Generative 3D Modelling and View Synthesis}
Generative 3D modelling has rapidly evolved from synthesising isolated objects to composing entire, complex scenes. Pioneering text-to-3D works like Shap-E \citep{jun2023shape} and Score Distillation Sampling (SDS) based methods like DreamFusion \citep{poole2022dreamfusion} laid the groundwork for single-object synthesis, inspiring a wave of research focused on high-fidelity object generation \citep{liang2024luciddreamer,tang2023dreamgaussian,wang2023prolificdreamer,shi2023mvdream}. More recently, the frontier has expanded to scene generation, with approaches ranging from procedural construction \citep{sun20233d,raistrick2023infinite} to direct compositional scene optimisation \citep{li2024dreamscene}. A common thread in many of these works is the reliance on SDS to refine an explicit 3D representation.

Generative view synthesis models \citep{liu2023zero,liu2023one, liu2023syncdreamer,shi2023zero123++} have emerged alongside this trend, but often face their own limitations. These methods typically struggle with multi-view consistency, are designed for a fixed number of reference (often one) and target views, and frequently rely on the same complex, two-stage SDS pipelines to enforce geometric coherence. Conversely, Kaleido's sequence-to-sequence design naturally handles an arbitrary number of both reference and target views, allowing it to generate spatially consistent views directly without requiring any post-processing or optimisation stages like SDS.

\paragraph{Sequence-to-Sequence Generative View Synthesis.}
Our work formulates generative view synthesis as a sequence-to-sequence modelling problem, built upon a pure transformer architecture. A critical challenge when applying transformers to this domain is effectively encoding camera positions. Recent advancements have introduced RoPE-style encodings \citep{su2021roformer} to parameterise 6-DoF camera extrinsics, with notable examples including CaPE \citep{kong2024eschernet}, GTA \citep{Miyato2024GTA}, and also camera intrinsics in a more recent work \citep{li2025cameras}. Kaleido builds directly on this direction, leveraging a GTA-based framework to create a unified representation for both multi-view 3D poses and temporal video positions.

While other sequence-to-sequence methods like CAT3D \citep{gao2025cat3d}, EscherNet \citep{kong2024eschernet} and SEVA \citep{zhou2025seva} have shown impressive results, their foundations lie in text-to-image latent diffusion models that use U-Net backbones. This reliance on a convolutional architecture is known to scale less effectively than pure transformers. Furthermore, these models often require 3D-specific learnable components, such as Plücker ray encodings for camera poses and a separate vision encoder for reference views. In contrast, Kaleido adheres to a pure transformer design from first principles, which results in a simpler, cleaner design that unifies 3D and video modelling, without any 3D-specific architectural modifications.


\paragraph{Generative Video and World Models}
Kaleido's methodology is deeply connected to recent advancements in generative video and the emerging paradigm of world models. The field of video generation has seen milestone progress with models like OpenAI's Sora \citep{brooks2024video} and DeepMind's Veo \citep{veo}, which have set a new standard for generative realism and temporal consistency. This progress is largely driven by a dominant technical stack combining diffusion or rectified flow models with transformer architectures, a foundation shared by many other generative video models \citep{blattmann2023svd,chen2024gentron,chen2024videocrafter2,yang2024cogvideox}. While Kaleido is built upon this same foundation and is pre-trained on large-scale video data, its goal is not to be a standalone video generator. Instead, it leverages video pre-training specifically to build a robust world representation for high-fidelity generative rendering.

The increasing capabilities of video generation have also positioned it as a stepping stone towards building world models --- systems that learn an internal model of the world to simulate physical interactions and predict future states. This trajectory is evident in models that focus on controllability and interactivity. For instance, the Navigation World Model \citep{bar2025navigation} predicts future observations to facilitate planning, while frameworks like WonderWorld \citep{yu2025wonderworld} and GameFactory \citep{yu2025gamefactory} generate explorable 3D environments. Most notably, the Genie series \citep{bruce2024genie, genie3} creates interactive environments with persistent spatial memory and real-time promptable world events, marking a significant advance toward truly immersive and dynamic virtual worlds.

Kaleido contributes to this broader pursuit of world modelling from a different perspective. Instead of focusing on temporal dynamics or agent-based interactivity, Kaleido approaches world modelling through the lens of neural rendering, prioritising {\it spatial consistency} and {\it generation flexibility}. This unique approach allows it to operate across a spectrum of realities: with many reference views, it produces a grounded reality through faithful reconstruction; while with few views, it creates a generated reality with plausible unseen details. This unique capability to seamlessly transition between reconstruction and creative generation marks a distinct and intriguing path toward creating truly versatile and navigable virtual worlds.

\section{Kaleido: Scaling Rectified Flow Transformers for Generative Rendering}

\subsection{Background and Notations}
Kaleido considers rendering and video generation within a unified sequence-to-sequence framework. The goal is to estimate the conditional distribution of a set of target views given a set of reference views:
\begin{align}
\label{eq:cond}
\mathcal{X}^T \sim p(\mathcal{X}^T| \mathcal{X}^R, \mathcal{P}^R, \mathcal{P}^T)
\end{align}
Here, the conditioning set consists of $N$ reference views $\mathcal{X}^R=\{x^R_{i=1:N}\}$ and their corresponding positions $\mathcal{P}^R=\{P^R_{i=1:N}\}$. The target set consists of $M$ target views $\mathcal{X}^T=\{x^T_{j=1:M}\}$ and their positions $\mathcal{P}^T=\{P^T_{j=1:M}\}$.

The positions $P$ are defined flexibly depending on their data modality. For 3D data, each $P \in SE(3)$ represents a 6-DoF camera pose. For video data, each $P \in \mathbb{N}$ represents a temporal position ({\it i.e.,} a frame index). 

This {\it "any-to-any view prediction"} can be seen as a form of {\it "next set-of-tokens prediction"}, which is elegantly handled by a masked auto-regressive framework \citep{li2024mar}. A key advantage of this approach is its flexibility: the number of reference views, $N$, and target views, $M$, can be arbitrary during both training and inference. This allows for various inference strategies, such as generating all target views at once or generating long sequences autoregressively by treating previously generated frames as new reference views. For training efficiency with batched optimisations, within each iteration, we sample a fixed total of $V$ views, and choose $N$ reference and $M$ target views such that $N+M=V$.

Kaleido is a latent rectified flow model \citep{rombach2022ldm,ma2024sit,esser2024sd3} that operates on spatially compressed image tokens. We first use a pre-trained VAE \citep{kingma2014vae} (with an $8 \times 8$ compression rate and 16 latent channels) to encode all reference and target images into a latent space: $\{\mathcal{Z}^R, \mathcal{Z}^T\} = \mathcal{E}(\{\mathcal{X}^R, \mathcal{X}^T\})$.

Following the rectified flow formulation \citep{liu2023flow,lipman2023flow}, we then construct a linear interpolation path between each target latent $z^T\in\mathcal{Z}^T$ (from the data distribution $p_0$) and a standard normal noise latent $\epsilon \sim \mathcal{N}(0, I)$ (from the noise distribution $p_1$):
\begin{align}
\mathcal{Z}_t^T = (1-t)z^T + t\epsilon, \quad \text{where } t \in [0,1] \text{ and } \forall z^T\in \mathcal{Z}^T.
\end{align}

A vision transformer (ViT) \citep{dosovitskiy2020vit} then processes the combined sequence of clean reference latents $\mathcal{Z}^R$ and noised target latents $\mathcal{Z}_t^T$. We tokenise the latents using a patch size of $2\times2$ (for a combined spatial compression of $16\times16$), which we found provides an optimal trade-off between generation quality and inference speed. Kaleido is trained with a standard noise-prediction objective \citep{kingma2023understanding}, applied only to the target latents $\mathcal{Z}_t^T$, to estimate a velocity field between $p_0$ and $p_1$, conditioned as defined in Eq.~\ref{eq:cond}. To analyse the scalability, we present three model variants: \textbf{Kaleido-Small}, \textbf{Kaleido-Medium}, and \textbf{Kaleido-Large (Kaleido)}, with detailed architectures and training strategies introduced next.

\subsection{Kaleido Architecture Details and Training Strategies}
\begin{figure}[t]
    \centering
    \includegraphics[width=\linewidth]{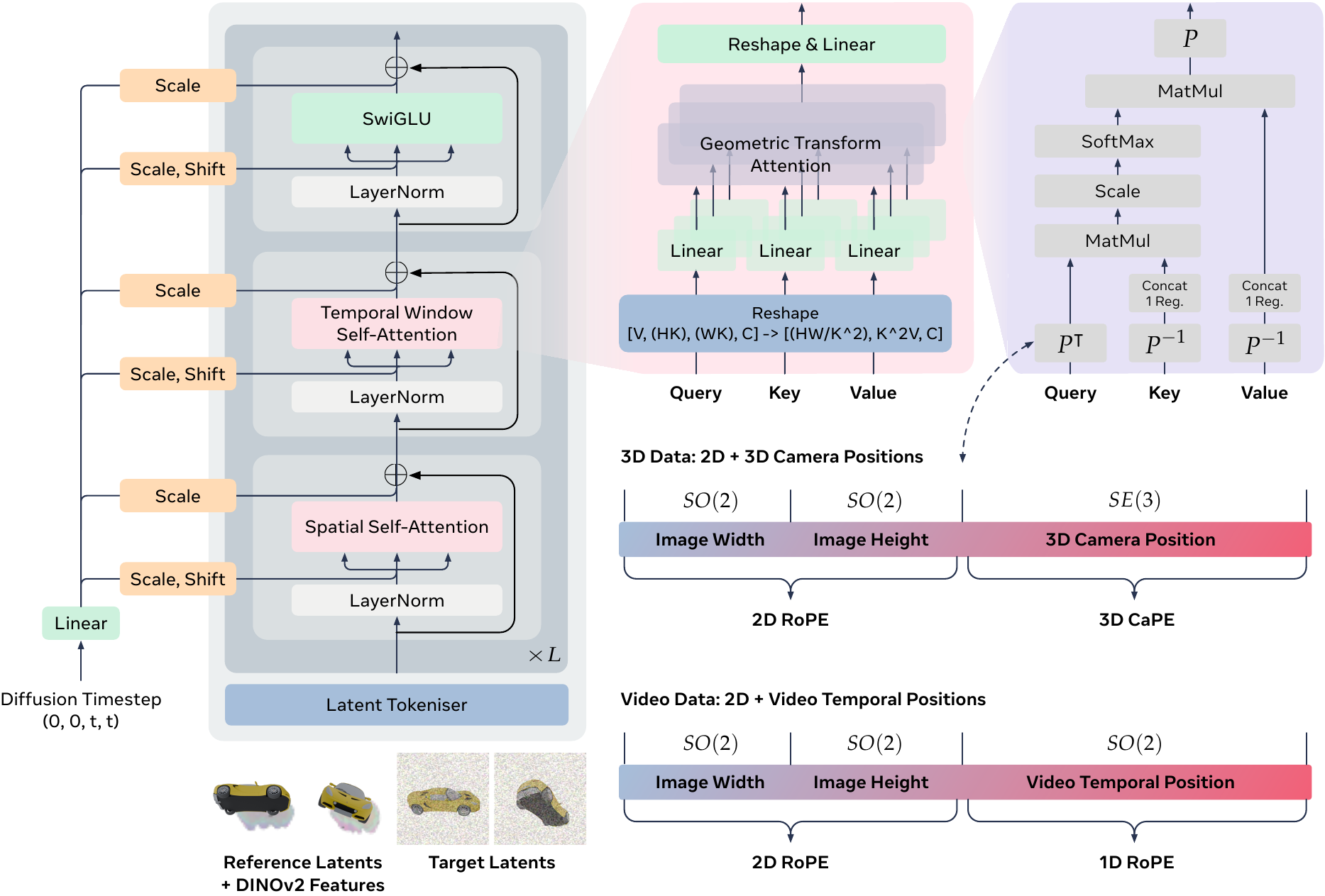}
    \caption{\textbf{Kaleido Architecture Design Details.} Kaleido is designed with a simple and scalable decoder-only transformer. It processes a sequence of tokens with clean reference latents (concatenated with their DINOv2 features) and noised target latents. During training, a single timestep $t$ is sampled per scene and integrated into the network via AdaIN layers, similar to DiT \citep{peebles2023dit}. The core of the model consists of repeating blocks of spatial self-attention (for within-frame interactions) followed by temporal window attention (for cross-frame interactions), and a SwiGLU feed-forward layer. Within each attention block, we encode a unified positional encoding design based on Geometric Transformation Attention (GTA) \citep{Miyato2024GTA}, which consistently represents all 2D, 3D, and temporal positions. This enables the same architecture to be trained on both video and multi-view 3D data without architectural changes.   }
    \label{fig:kaleido_arch}
\end{figure}

In this section, we present a comprehensive ablation study of the Kaleido architecture design and its training strategies. Our goal is to identify the key design decisions that address the unique scaling challenges of sequence-to-sequence generative neural rendering. Our main findings are summarised in Fig.~\ref{fig:ablation}, with full quantitative results and explorations of alternative designs detailed in Appendix~\ref{app:ablation}. An overview of the final Kaleido architecture is shown in Fig.~\ref{fig:kaleido_arch}.

To provide a holistic view of our design process, we conducted a series of controlled experiments. For efficiency, we used our \textbf{Kaleido-Small} for all ablations, allowing for rapid iteration. We trained each experimental configuration on two distinct datasets: \textbf{Objaverse} \citep{deitke2023objaverse}, which contains synthetic objects with ground-truth camera poses, and \textbf{uCO3D} \citep{liu2025uco3d}, which contains real-world objects with noisy, estimated camera poses. Each experiment is trained for 100K optimisation steps on $8\times$ H100 GPUs. We perform a greedy search over key design choices, organised by the four primary objectives introduced next.

\subsubsection{Designing Kaledio's Design Spaces} 
\label{subsec:kaleido_design}
We begin by exploring Kaleido’s architectural design spaces and training strategies. Our Kaleido-Small's starting point is a vanilla DiT-L/SiT-L architecture \citep{peebles2023dit,ma2024sit} within a rectified flow framework, whose scaling properties have been well established in image and video generation \citep{esser2024sd3,polyak2024moviegen,chen2025goku}.

\paragraph{\textbf{(i)} Improved Architecture Design with Llama 3.} We first incorporate recent architectural advances from state-of-the-art sequence-to-sequence language models like Llama-3~\citep{dubey2024llama3}. Specifically, we replace the standard GLU activations in our transformer's feed-forward layers with \textbf{SwiGLU} \citep{shazeer2020swiglu} and swap multi-head attention (MHA) for the more efficient \textbf{grouped-query attention (GQA)} \citep{ainslie2023gqa}. These simple modifications yield consistent performance gains across our experiments without increasing computational overhead.

\paragraph{\textbf{(ii)} Unified Positional Encoding for Space and Time} One of the critical design decisions in Kaleido is a unified positional encoding that seamlessly represents 2D, 3D, and temporal positions within a single, consistent design. Specifically, we introduce a parameter-free encoding scheme that extends the principles of \textbf{RoPE-style relative encodings} \citep{su2021roformer} and \textbf{Geometric Transformation Attention (GTA)} \citep{Miyato2024GTA}, which we adapt and generalise to create a unified representation for space and time. This design allows Kaleido to process both multi-view 3D and video data without any architectural modifications.

We represent different positions as follows: In 2D image positions, pixel coordinates are mapped to a pair of angles $(\theta_h, \theta_w)$, representing an element in $SO(2) \times SO(2)$, where $\theta_{h,w} \in [0, 2\pi)$ distributed uniformly from the top-left to the bottom-right patches; In temporal positions, frame indices are similarly mapped to a single angle $\theta_t \in SO(2)$, with values interpolated linearly from the start to the end of a clip. In 3D camera poses, 6-DoF camera extrinsics 
$c = 
\begin{bsmallmatrix}
\mathbf{R} & \mathbf{t} \\
0 &1
\end{bsmallmatrix}
$ 
(with rotation $\mathbf{R}$ and translation $\mathbf{t}$) 
are represented as an element in $SE(3)$, following the design in CaPE \citep{kong2024eschernet}.

This allows us to define a unified geometric attribute $g$ for each image token, depending on its data modality:
\begin{alignat}{2}
&\text{For 3D data:}\quad  && g := (\theta_h, \theta_w, c) \in SO(2) \times SO(2) \times SE(3) \\
&\text{For video data:}\quad  && g := (\theta_h, \theta_w, \theta_t) \in SO(2) \times SO(2) \times SO(2). 
\end{alignat}

Within the GTA framework, these components are used to construct a block-diagonal transformation matrix $P_g$ that is applied to each token's feature vector $v \in \mathbb{R}^d$. The construction of $P_g$ varies for different attention blocks, allocating the feature dimension $d$ as follows: In Spatial Attention, we apply only the 2D position embeddings $(\theta_h, \theta_w)$, which are expanded into $d/4$ distinct frequency bands, with dimensions allocated to image height and width components based on a 1:1 ratio; In Temporal/3D Attention, the 2D embeddings $(\theta_h, \theta_w)$ are expanded into $d/8$ frequency bands.  For video data, the temporal embedding $\theta_t$ is expanded into $d/4$ frequency bands. For 3D data, the pose embedding $c$ is repeated to fill the remaining dimensions. The total dimensions are allocated to image height, width, and temporal/3D components based on a 1:1:2 ratio.

Finally, we normalise the camera translation element $\mathbf{t}$ such that its maximum norm across all views in a given scene is $1$. This ensures all positional transformations remain within a {\it bounded} range, which we found is crucial for stable training to handle different scene scales.

Our ablations confirm that this unified design outperforms (more significantly in multi-view settings) both a simpler baseline (2D RoPE + 3D CaPE without value-transformation) and the Plücker embeddings used in other leading sequence-to-sequence rendering models \citep{gao2025cat3d,zhou2025seva,jin2024lvsm}. 

\begin{figure}[t]
    \centering
    \includegraphics[width=\linewidth]{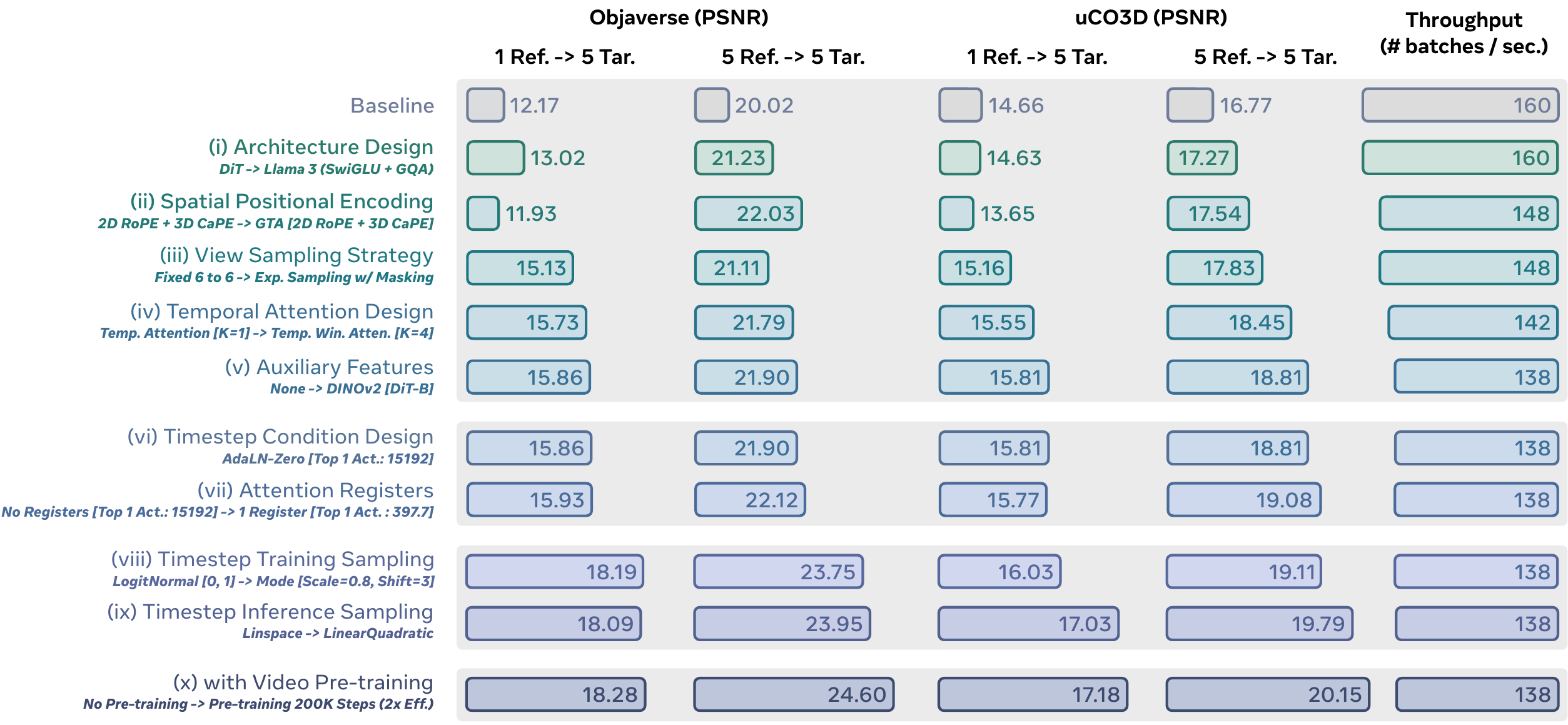}
    \footnotesize
    \caption{\textbf{Kaleido Design Ablations.} We extensively ablate various architectural designs and training strategies to explore effective scaling strategies for generative neural rendering. Each ablation experiment was conducted with Kaleido-Small, trained for 100K steps in total, on a mixture of Objaverse and uCO3D sampled randomly. We report PSNR and training throughput for each configuration and evaluate performance in two settings: 5 target views conditioned on 1 reference view, and 5 target views conditioned on 5 reference views. We broadly split our designs into four categories: the Kaleido architecture design spaces (i–v); scaling stability techniques to handle large activations (vi–vii); training and inference timestep sampling strategies (viii–ix); and the role of video pre-training (x). The arrow $(\to)$ indicates the progression from our initial baseline design to our final, optimised design choice. }
    \label{fig:ablation}
\end{figure}

\paragraph{\textbf{(iii)} Principled View Sampling Improves Generalisation.} The strategy for sampling reference and target views is an {\it often-overlooked design aspect} of sequence-to-sequence rendering models, which are typically trained with a fixed number of reference and target views \citep{jin2024lvsm,kong2024eschernet,gao2025cat3d}. While such fixed-view training can generalise to other view configurations, we found that performance can be further improved by adopting a more principled sampling strategy.

Our key insight is that {\it the rendering task becomes easier and more constrained as the number of reference views increases}. Therefore, the training process should place more emphasis on the challenging, less-constrained scenarios involving fewer reference views. To achieve this, we designed a sampling distribution  $\pi(n)$ where its probability density halves as the number of reference views $n$ increases, {\it i.e.}, $\pi(n+1) = \frac{1}{2}\pi(n), n\geq 0$. We design an \textbf{exponential distribution} which elegantly provides this property:
\begin{align}
\pi(n) = \lambda e^{-\lambda n}, \quad \text{where } \lambda = \ln(2)
\end{align}
In each training step, we sample the number of reference views $N$ from this distribution and set the remaining $M=V-N$ as target views. By combining this with random attention masking, our model is exposed to all possible combinations of $(n,m),n\in[1,N], m\in[1,M]$ view pairs, such that $n+m\in[2,V]$.

Our experiments show this sampling strategy offers the best trade-off between single- and multi-view conditioning. It significantly outperforms both fixed-view sampling and uniform sampling, which tend to degrade single-view performance by over-emphasising multi-view settings.

\paragraph{\textbf{(iv)} Expanded Perception Field with Window Attention}
Our baseline model processes a token sequence of shape $V \times H \times W$ (a sequence of $V$ image latents with height $H$ and width $W$) using a standard factorised attention mechanism: Spatial Attention (within-frame) followed by Temporal Attention (cross-frame). This approach has a computational complexity of $O(H^2W^2) + O(V^2)$, which is efficient but limits cross-view interactions to a single token at each spatial location. To improve this, we redesign the Temporal Attention layer by expanding its receptive field. Instead of attending to a single token across frames, each query token now attends to a local $K \times K$ window around the corresponding spatial location in all other frames. This \textbf{windowed cross-view attention} design significantly improves feature exchange between views while maintaining computational efficiency. The complexity only increases by a small, constant factor from $O(V^2)$ to $O(V^2K^4)$, which is far more scalable than the full attention's complexity of $O(V^2H^2W^2)$ cost, given that $K \ll H,W$. 

Our experiments show that this design consistently boosts performance with larger window sizes. In practice, we use a window size of $K=4$ for our Small and Medium models and $K=8$ for our Large model.

\paragraph{\textbf{(v)} Integration of Auxiliary Visual Features}
We study the integration of auxiliary visual features from pre-trained networks to enhance 3D perception. Our findings show that features from \textbf{DINOv2} \citep{oquab2024dinov2} further improve Kaleido's depth estimation on in-the-wild images, leading to more accurate renderings. These pre-trained semantic features performed similarly to, and sometimes slightly better than, pre-computed depth or surface normals, which encode explicit scene geometry built on top of the same DINOv2 model.

We also observed that larger DINOv2 models provide additional, albeit marginal, performance gains. Based on this trade-off, we pair the feature extractor with our model size: we use DINOv2 with ViT-B backbone for Kaleido-Small and Medium, and DINOv2 with ViT-L backbone for our Large model.

\subsubsection{Massive Activations in Rectified Flow Transformers}
During our initial scaling experiments with Kaleido, we consistently observed severe instability in training convergence on high-resolution images. A deeper analysis revealed that this instability arises from massive activations emerging within the transformer layers.

\begin{figure}[ht!]
    \vspace{-0.2cm}
    \centering
    \begin{subfigure}[b]{0.32\linewidth}
      \centering
        \resizebox{\linewidth}{!}{{
\begin{tikzpicture}

\definecolor{c1}{RGB}{243,141,59}
\definecolor{c2}{RGB}{65,183,196}
\definecolor{c3}{RGB}{120,198,121}
\definecolor{c4}{RGB}{140,150,198}

\begin{axis}[
    every axis y label/.style={at={(current axis.north west)},above=0mm},
    legend cell align={left},
    legend style={fill opacity=0.0, draw opacity=1, text opacity=1, draw=none, legend columns=1, column sep=2pt, very thick, anchor=north west, at={(0.0,1.0)}, font=\footnotesize},
    width=8cm, height=5cm,
    tick align=outside,
    tick pos=left,
    xmin=-1.15, xmax=25,
    ymin=-0.1, ymax=28000,
    xlabel={Layer Index},
    ymode=log,
    xtick={0, 5, 11, 17, 23},
    xticklabels={1, 6, 12, 18, 24},
    ytick={10, 100, 1000, 10000},
    ylabel={Magnitude},
]
\addplot [thick, mark=*, opacity=1.0, c1]
table {%
0 79.875
1 96.875
2 106.9375
3 115.1875
4 137.625
5 148.625
6 178.375
7 185.25
8 193.375
9 220.75
10 15080
11 15080
12 15080
13 15192
14 15192
15 15192
16 15192
17 15192
18 15192
19 15192
20 15192
21 15192
22 15192
23 15192
};
\addlegendentry{Top 1}
\addplot [thick, mark=*, opacity=1.0, c2]
table {%
0 30.6875
1 34.0625
2 36.21875
3 38.78125
4 46.21875
5 50.125
6 60.28125
7 64.6875
8 66.75
9 69.25
10 156.625
11 160.125
12 161.25
13 156.25
14 161.5
15 163
16 161.625
17 162.125
18 165.25
19 166.625
20 168.625
21 177
22 182
23 219.75
};
\addlegendentry{Top 0.001\%}
\addplot [thick, mark=*, opacity=1.0, c3]
table {%
0 17.859375
1 19.96875
2 22.390625
3 23.015625
4 23.21875
5 24.15625
6 25.265625
7 27.8125
8 32.125
9 34.96875
10 64.875
11 67.875
12 66.0625
13 72.375
14 82.8125
15 100.125
16 118.125
17 120.9375
18 134.5
19 140.75
20 142.875
21 151.875
22 157
23 174.125
};
\addlegendentry{Top 0.01\%}
\addplot [thick, mark=*, opacity=1.0, c4]
table {%
0 5.125
1 5.8203125
2 6.5078125
3 6.63671875
4 6.77734375
5 7.23828125
6 8.1796875
7 8.5625
8 9.4375
9 9.96875
10 11.3125
11 11.9609375
12 12.5078125
13 13.5390625
14 13.5546875
15 13.7578125
16 14.171875
17 14.828125
18 15.421875
19 16.109375
20 17.578125
21 18.5
22 19.34375
23 19.75
};
\addlegendentry{Top 0.1\%}

\draw[-, thick, shorten <=2pt] (axis cs:23,15192) -- (axis cs:23,11000) node[pos=1, anchor=north] {\footnotesize \textcolor{c1}{\bf 15192}};
\draw[-, thick, shorten <=2pt] (axis cs:23,219.75) -- (axis cs:23,300) node[pos=1, anchor=south] {\footnotesize \textcolor{c2}{\bf 219.75}};
\draw[-, thick, shorten <=2pt] (axis cs:23,174.125) -- (axis cs:23,120) node[pos=1, anchor=north] {\footnotesize \textcolor{c3}{\bf 174.13}};
\draw[-, thick, shorten <=2pt] (axis cs:23,19.75) -- (axis cs:23,14.5) node[pos=1, anchor=north] {\footnotesize \textcolor{c4}{\bf 19.75}};
\end{axis}

\end{tikzpicture}}}
      \caption{\centering Massive activations are very\\ few in numbers.}
      \label{fig:act_a}
    \end{subfigure}\hfill
    \begin{subfigure}[b]{0.32\linewidth}
      \centering
        \resizebox{\linewidth}{!}{{
\begin{tikzpicture}

\definecolor{c1}{RGB}{140,150,198}
\definecolor{c2}{RGB}{140,150,198}
\definecolor{c3}{RGB}{120,198,121}
\definecolor{c4}{RGB}{120,198,121}

\begin{axis}[
    every axis y label/.style={at={(current axis.north west)},above=0mm},
    legend cell align={left},
    legend style={fill opacity=0.0, draw opacity=1, text opacity=1, draw=none, legend columns=1, column sep=2pt, very thick, anchor=north west, at={(0.0,1.0)}, font=\footnotesize},
    width=8cm, height=5cm,
    tick align=outside,
    tick pos=left,
    xmin=5000, xmax=106500,
    ymin=-0.1, ymax=150000,
    xlabel={Training Steps},
    ymode=log,
    scaled x ticks=base 10:-4,
    xtick={10000,20000,30000,40000,50000,60000,70000,80000,90000,100000},
    ytick={10, 100, 1000, 10000, 100000},
    ylabel={Magnitude},
]
\addplot [thick, mark=*, opacity=1.0, c1]
table {%
10000 34.87
20000 55.78
30000 387.5
40000 1191
50000 3360
60000 6652
70000 10448
80000 14774
90000 18896
100000 24352
};
\addlegendentry{512 w/o Reg.}
\addplot [thick, mark=*, opacity=0.5, c2]
table {%
10000 34.53
20000 72.06
30000 141.625
40000 206
50000 215.75
60000 251.25
70000 276.5
80000 297
90000 322.25
100000 333
};
\addlegendentry{512 w/ 1 Reg.}
\addplot [thick, mark=*, opacity=1.0, c3]
table {%
10000 53.4
20000 77.625
30000 122.6875
40000 164.25
50000 296.25
60000 2244
70000 5500
80000 9472
90000 13136
100000 15104
};
\addlegendentry{256 w/o Reg.}
\addplot [thick, mark=*, opacity=0.5, c4]
table {%
10000 45.46
20000 80.187
30000 108.75
40000 143.87
50000 178.62
60000 211.25
70000 248.25
80000 301.75
90000 337.75
100000 363.75
};
\addlegendentry{256 w/ 1 Reg.}
\draw[-, thick, shorten <=2pt] (axis cs:100000,363.75) -- (axis cs:100000,250) node[pos=1, anchor=north] {\footnotesize \textcolor{c3}{\bf 363.8}};
\draw[-, thick, shorten <=2pt] (axis cs:100000,333) -- (axis cs:100000,500) node[pos=1, anchor=south] {\footnotesize \textcolor{c1}{\bf 333}};
\draw[-, thick, shorten <=2pt] (axis cs:100000,15104) -- (axis cs:100000,10000) node[pos=1, anchor=north] {\footnotesize \textcolor{c3}{\bf 15104}};
\draw[-, thick, shorten <=2pt] (axis cs:100000,24352) -- (axis cs:100000,36000) node[pos=1, anchor=south] {\footnotesize \textcolor{c1}{\bf 24352}};
\end{axis}

\end{tikzpicture}}}
      \caption{\centering The activations grow faster and larger with more tokens.}
      \label{fig:act_b}
    \end{subfigure}\hfill
    \begin{subfigure}[b]{0.32\linewidth}
      \centering
        \resizebox{\linewidth}{!}{{
\begin{tikzpicture}

\definecolor{c1}{RGB}{243,141,59}
\definecolor{c2}{RGB}{65,183,196}
\definecolor{c3}{RGB}{120,198,121}
\definecolor{c4}{RGB}{140,150,198}

\begin{axis}[
    every axis y label/.style={at={(current axis.north west)},above=0mm},
    legend cell align={left},
    legend style={fill opacity=0.0, draw opacity=1, text opacity=1, draw=none, legend columns=1, column sep=2pt, very thick, anchor=north west, at={(0.0,1.0)}, font=\footnotesize},
    width=8cm, height=5cm,
    tick align=outside,
    tick pos=left,
    xmin=-1.15, xmax=25,
    ymin=-10, ymax=22000,
    xlabel={Layer Index},
    ymode=log,
    xtick={0, 5, 11, 17, 23},
    xticklabels={1, 6, 12, 18, 24},
    ytick={10, 100, 1000, 10000},
    ylabel={Magnitude},
]
\addplot [thick, mark=*, opacity=1.0, c1]
table {%
0 79.9375
1 97.125
2 107.125
3 115.0625
4 138.125
5 149
6 178.625
7 183.625
8 192.625
9 219
10 15008
11 15008
12 15008
13 15120
14 15120
15 15120
16 15120
17 15120
18 15120
19 15120
20 15120
21 15120
22 15120
23 15120
};
\addlegendentry{w/o Reg.}
\addplot [thick, mark=*, opacity=1.0, c2]
table {%
0 88.1875
1 91.1875
2 95.1875
3 100.1875
4 120.6875
5 131.25
6 142.25
7 161.25
8 169.125
9 179.75
10 190.375
11 202.375
12 207.625
13 205.75
14 236.25
15 283.5
16 267
17 263.5
18 263.25
19 276.75
20 290.75
21 327.75
22 333.75
23 397.75
};
\addlegendentry{w/ 1 Reg.}
\addplot [thick, mark=*, opacity=1.0, c3]
table {%
0 82
1 87.3125
2 91.4375
3 98.75
4 115.5625
5 125.375
6 143.875
7 148.25
8 155.875
9 174.75
10 179.875
11 185
12 183.375
13 182.375
14 201.75
15 202.125
16 203.75
17 206.75
18 214.25
19 216.875
20 220.125
21 229
22 245.75
23 279.75
};
\addlegendentry{w/ 8 Reg.}
\addplot [thick, mark=*, opacity=1.0, c4]
table {%
0 57.1875
1 63.6875
2 67.9375
3 87
4 117.3125
5 139.75
6 156.625
7 171.5
8 195.75
9 214.25
10 187.75
11 177.875
12 184.875
13 197.5
14 230.5
15 259.25
16 267.75
17 249.5
18 242.875
19 235.5
20 227.875
21 222.75
22 217.375
23 238.75
};
\addlegendentry{w/ 32 Reg.}
\draw[-, thick, shorten <=2pt] (axis cs:23,15192) -- (axis cs:23,12000) node[pos=1, anchor=north] {\footnotesize \textcolor{c1}{\bf 15192}};
\draw[-, thick, shorten <=2pt] (axis cs:23,397.75) -- (axis cs:23,500) node[pos=1, anchor=south] {\footnotesize \textcolor{c2}{\bf 397.75}};
\draw[-, thick, shorten <=2pt] (axis cs:23,238.75) -- (axis cs:23,180) node[pos=1, anchor=north] {\footnotesize \textcolor{c4}{\bf 238.75}};
\end{axis}

\end{tikzpicture}}}
      \caption{\centering The use of learnable registers effectively reduces massive activations.}
      \label{fig:act_c}
    \end{subfigure}
    \vspace{-0.2cm}
    \caption{\textbf{Visual Analysis of Massive Activations in a Rectified Flow Transformer.} We provide an empirical analysis of massive activations emerging during training. (a) Visualisation of activation magnitudes across model layers at 100K training steps, showing they are sparse but grow suddenly at a middle layer. (b) The maximum activation magnitude (measured at the final layer) grows over training time and correlates positively with image resolution (and thus, number of tokens). (c) The same training configuration as (a), but with learnable register tokens applied, demonstrating a significant and consistent reduction in activation magnitudes. }
    \label{fig:activations}
\end{figure}
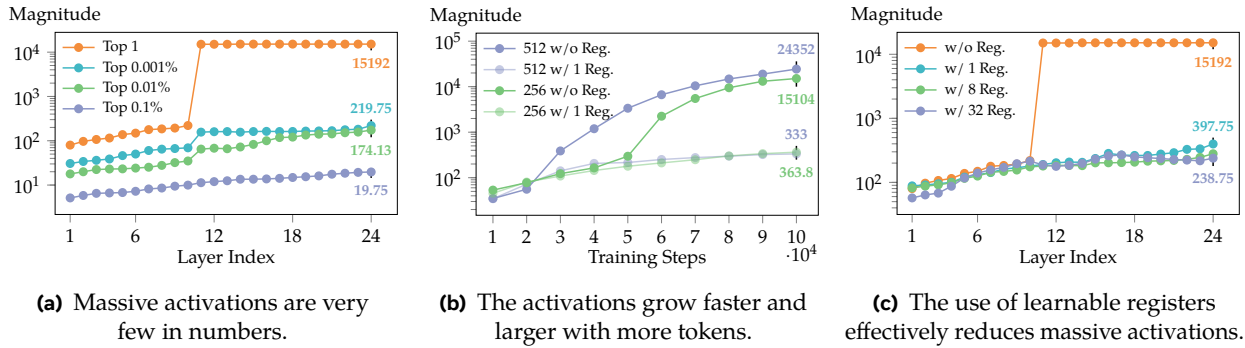

While massive activations have been studied extensively in autoregressive language models \citep{sun2024massive} (where they are sometimes called ``attention sinks'' \citep{xiao2023attentionsink,gu2024attentionsink}) and in visual representation learning \citep{da2023register}, their behaviour within diffusion or rectified flow models remains largely unexplored. In other contexts, they are known to act as {\it attention biases} or {\it global information aggregators}. In this section, we provide the first empirical analysis of this issue in the context of rectified flow transformers, comparing our findings to those observed in LLMs \citep{sun2024massive} and ViTs \citep{da2023register} (illustrated in Fig.~\ref{fig:activations}). Our key observations are:

\begin{enumerate}
    \item Similar to the observations in language models, massive activations are very sparse in numbers, with only a few tokens exhibiting this behaviour (Fig. \ref{fig:act_a}).
    \item The magnitude of these activations grows with model depth. We observe a sudden jump at a middle layer, after which the magnitude remains constantly high until the final layer. Unlike in language models, these activations do not diminish towards the end of the model depth.
    \item The activation magnitudes in rectified flow transformers are significantly higher than those reported in other domains. While language transformers report values around 1K-2K and vision transformers around 200, our activations can reach as high as 15K for 256px resolution and 24K for 512px resolution (Fig. \ref{fig:act_b}). These values seem to positively correlate with the number of training tokens and they continue to grow during training, directly causing precision overflow in our {\tt fp16} mixed-precision training.
\end{enumerate}

We also found that these massive activations emerge most prominently when training on a mixture of synthetic and real-world data. This supports the hypothesis in ViTs that they act as {\it global information aggregators}, perhaps in our context to reconcile the {\it different rendering logic} required for synthetic scenes ({\it e.g.}, maintaining clean/solid colour backgrounds) and real scenes ({\it e.g.}, generating semantically consistent textures).

To resolve this instability, we adopted the solution from \cite{sun2024massive}, appending learnable "register" tokens to the keys and values in each attention layer. This simple design proved highly effective, consistently reducing activation magnitudes to a stable level ($\sim$300) for both low and high-resolution training, as shown in Fig.~\ref{fig:act_b} and \ref{fig:act_c}. Additionally, our \textbf{ablation (vii)} shows that having 1 register token is already optimal, adding more provides no benefit and can even degrade performance.

In \textbf{ablation (vi)}, we also explored alternative solutions inspired by recent studies \citep{bozic2021transformerfusion,sun2025noise,tang2025exploring}, such as removing the scaling factor from the timestep conditioning or removing timestep conditioning entirely. While these modifications slightly reduced activation magnitudes, they were far less effective than using register tokens and consistently resulted in lower overall performance.

Finally, we empirically observed that massive activations are a pervasive phenomenon. They appear across all our model sizes, persist in both diffusion and rectified flow frameworks for image and video generation, and are independent of the inference timestep or input data. While our register token solution effectively stabilises training, we believe we have only scratched the surface of understanding this issue in visual generative models. A deeper analysis could inspire more stable and efficient architectures, which we consider an important direction for future work.

\subsubsection{Tailored Rectified Flow SNR Samplers for Generative Neural Rendering}
Prior rectified flow models for text-to-image/video generation \citep{esser2024sd3,polyak2024moviegen} typically use logit-normal sampling \citep{atchison1980logitnormal} to focus on intermediate timesteps $t\in[0,1]$. However, we found this approach to be suboptimal for rendering tasks, as generation quality becomes highly sensitive to the exact inference timesteps chosen, especially near the start ($t\approx1$) and end ($t\approx0$) of the trajectory.

We hypothesise this discrepancy arises from a fundamental difference between generation tasks: Text-conditioned image and video synthesis explores a {\it vast, unconstrained solution space}, whereas image-to-3D rendering is a {\it highly constrained problem}, as the output must be spatially consistent with the provided reference images. We argue this insight implies that rendering models should focus more heavily on the {\it early, high-noise timesteps} where the initial scene structure is formed.

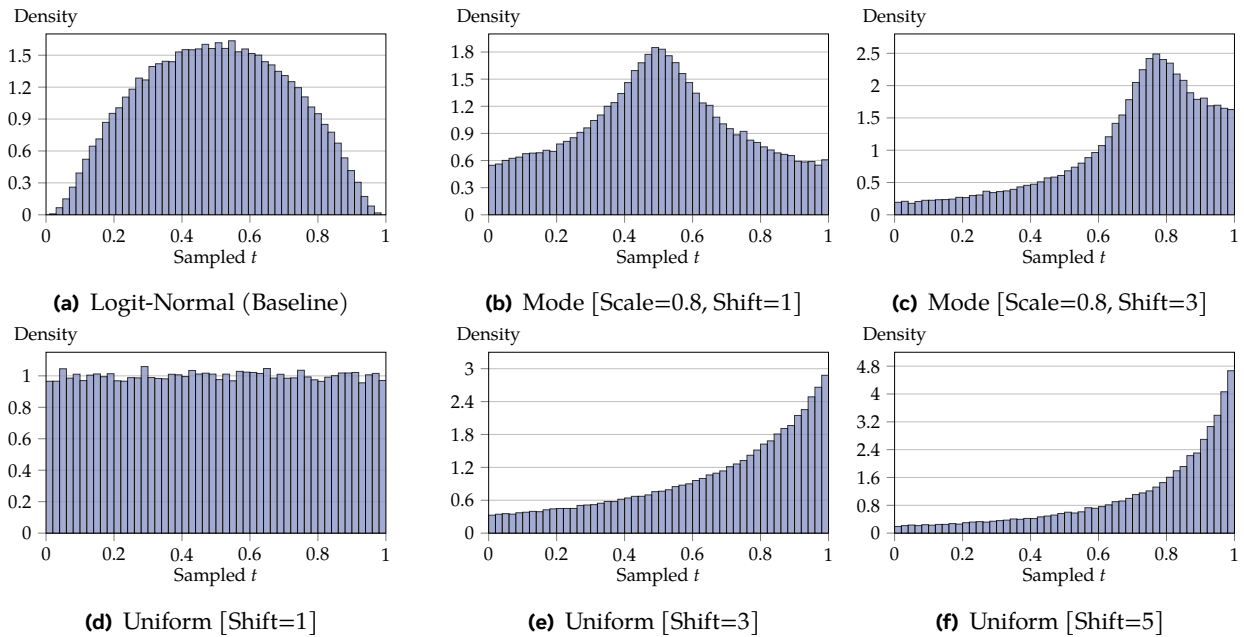
\begin{figure}[ht!]
    \centering
    \begin{subfigure}[b]{0.32\linewidth}
      \centering
        \resizebox{\linewidth}{!}{{
\begin{tikzpicture}
\definecolor{c2}{RGB}{140,150,198}

\begin{axis}[
    every axis y label/.style={at={(current axis.north west)},above=0mm},
    width=8cm, height=5cm,
    tick align=outside,
    tick pos=left,
    xlabel={Sampled $t$},
    xmin=0, xmax=1.0,
    ylabel={Density},
    ymajorgrids,
    ymin=0, ymax=1.7,
    xtick={0, 0.2, 0.4, 0.6, 0.8, 1},
    ytick={0, 0.3, 0.6, 0.9, 1.2, 1.5}
]
\draw[draw=black,fill=c2,opacity=0.8] (axis cs:0.0102377291768789,0) rectangle (axis cs:0.0297465939074755,0.00871398732563777);
\draw[draw=black,fill=c2,opacity=0.8] (axis cs:0.0297465939074755,0) rectangle (axis cs:0.0492554567754269,0.0666363800288741);
\draw[draw=black,fill=c2,opacity=0.8] (axis cs:0.0492554567754269,0) rectangle (axis cs:0.0687643215060234,0.149675547005072);
\draw[draw=black,fill=c2,opacity=0.8] (axis cs:0.0687643215060234,0) rectangle (axis cs:0.0882731825113297,0.25988190692532);
\draw[draw=black,fill=c2,opacity=0.8] (axis cs:0.0882731825113297,0) rectangle (axis cs:0.107782050967216,0.392641942167003);
\draw[draw=black,fill=c2,opacity=0.8] (axis cs:0.107782050967216,0) rectangle (axis cs:0.127290919423103,0.522326552308325);
\draw[draw=black,fill=c2,opacity=0.8] (axis cs:0.127290919423103,0) rectangle (axis cs:0.146799772977829,0.644835431498369);
\draw[draw=black,fill=c2,opacity=0.8] (axis cs:0.146799772977829,0) rectangle (axis cs:0.166308641433716,0.713009062081334);
\draw[draw=black,fill=c2,opacity=0.8] (axis cs:0.166308641433716,0) rectangle (axis cs:0.185817509889603,0.868835629207664);
\draw[draw=black,fill=c2,opacity=0.8] (axis cs:0.185817509889603,0) rectangle (axis cs:0.20532637834549,0.953412548865048);
\draw[draw=black,fill=c2,opacity=0.8] (axis cs:0.20532637834549,0) rectangle (axis cs:0.224835231900215,1.00569723100143);
\draw[draw=black,fill=c2,opacity=0.8] (axis cs:0.224835231900215,0) rectangle (axis cs:0.244344100356102,1.1061635916402);
\draw[draw=black,fill=c2,opacity=0.8] (axis cs:0.244344100356102,0) rectangle (axis cs:0.263852953910828,1.18100225291911);
\draw[draw=black,fill=c2,opacity=0.8] (axis cs:0.263852953910828,0) rectangle (axis cs:0.283361822366714,1.28505659140036);
\draw[draw=black,fill=c2,opacity=0.8] (axis cs:0.283361822366714,0) rectangle (axis cs:0.302870690822601,1.26762862007702);
\draw[draw=black,fill=c2,opacity=0.8] (axis cs:0.302870690822601,0) rectangle (axis cs:0.322379559278488,1.3937251184753);
\draw[draw=black,fill=c2,opacity=0.8] (axis cs:0.322379559278488,0) rectangle (axis cs:0.341888427734375,1.41986707546031);
\draw[draw=black,fill=c2,opacity=0.8] (axis cs:0.341888427734375,0) rectangle (axis cs:0.361397296190262,1.44293350809414);
\draw[draw=black,fill=c2,opacity=0.8] (axis cs:0.361397296190262,0) rectangle (axis cs:0.380906134843826,1.438835006966);
\draw[draw=black,fill=c2,opacity=0.8] (axis cs:0.380906134843826,0) rectangle (axis cs:0.400415003299713,1.53058595210271);
\draw[draw=black,fill=c2,opacity=0.8] (axis cs:0.400415003299713,0) rectangle (axis cs:0.4199238717556,1.55262720995281);
\draw[draw=black,fill=c2,opacity=0.8] (axis cs:0.4199238717556,0) rectangle (axis cs:0.439432740211487,1.55108944777722);
\draw[draw=black,fill=c2,opacity=0.8] (axis cs:0.439432740211487,0) rectangle (axis cs:0.458941608667374,1.55929084604703);
\draw[draw=black,fill=c2,opacity=0.8] (axis cs:0.458941608667374,0) rectangle (axis cs:0.47845047712326,1.60234818696352);
\draw[draw=black,fill=c2,opacity=0.8] (axis cs:0.47845047712326,0) rectangle (axis cs:0.497959345579147,1.56287895779007);
\draw[draw=black,fill=c2,opacity=0.8] (axis cs:0.497959345579147,0) rectangle (axis cs:0.517468214035034,1.61721322132754);
\draw[draw=black,fill=c2,opacity=0.8] (axis cs:0.517468214035034,0) rectangle (axis cs:0.536977052688599,1.56390652164349);
\draw[draw=black,fill=c2,opacity=0.8] (axis cs:0.536977052688599,0) rectangle (axis cs:0.556485950946808,1.63566386874628);
\draw[draw=black,fill=c2,opacity=0.8] (axis cs:0.556485950946808,0) rectangle (axis cs:0.575994789600372,1.53212605479921);
\draw[draw=black,fill=c2,opacity=0.8] (axis cs:0.575994789600372,0) rectangle (axis cs:0.595503628253937,1.55878063989441);
\draw[draw=black,fill=c2,opacity=0.8] (axis cs:0.595503628253937,0) rectangle (axis cs:0.615012526512146,1.5162311888911);
\draw[draw=black,fill=c2,opacity=0.8] (axis cs:0.615012526512146,0) rectangle (axis cs:0.63452136516571,1.50085817612984);
\draw[draw=black,fill=c2,opacity=0.8] (axis cs:0.63452136516571,0) rectangle (axis cs:0.65403026342392,1.43934319756802);
\draw[draw=black,fill=c2,opacity=0.8] (axis cs:0.65403026342392,0) rectangle (axis cs:0.673539102077484,1.40910489282135);
\draw[draw=black,fill=c2,opacity=0.8] (axis cs:0.673539102077484,0) rectangle (axis cs:0.693048000335693,1.34861536780679);
\draw[draw=black,fill=c2,opacity=0.8] (axis cs:0.693048000335693,0) rectangle (axis cs:0.712556838989258,1.30863761053944);
\draw[draw=black,fill=c2,opacity=0.8] (axis cs:0.712556838989258,0) rectangle (axis cs:0.732065677642822,1.25174032312468);
\draw[draw=black,fill=c2,opacity=0.8] (axis cs:0.732065677642822,0) rectangle (axis cs:0.751574575901031,1.19483938516063);
\draw[draw=black,fill=c2,opacity=0.8] (axis cs:0.751574575901031,0) rectangle (axis cs:0.771083414554596,1.10821563415051);
\draw[draw=black,fill=c2,opacity=0.8] (axis cs:0.771083414554596,0) rectangle (axis cs:0.790592312812805,1.01338372563817);
\draw[draw=black,fill=c2,opacity=0.8] (axis cs:0.790592312812805,0) rectangle (axis cs:0.81010115146637,0.950338476278933);
\draw[draw=black,fill=c2,opacity=0.8] (axis cs:0.81010115146637,0) rectangle (axis cs:0.829610049724579,0.849868597424421);
\draw[draw=black,fill=c2,opacity=0.8] (axis cs:0.829610049724579,0) rectangle (axis cs:0.849118888378143,0.776571084985213);
\draw[draw=black,fill=c2,opacity=0.8] (axis cs:0.849118888378143,0) rectangle (axis cs:0.868627727031708,0.673540861828759);
\draw[draw=black,fill=c2,opacity=0.8] (axis cs:0.868627727031708,0) rectangle (axis cs:0.888136625289917,0.535140419608622);
\draw[draw=black,fill=c2,opacity=0.8] (axis cs:0.888136625289917,0) rectangle (axis cs:0.907645463943481,0.41570900985017);
\draw[draw=black,fill=c2,opacity=0.8] (axis cs:0.907645463943481,0) rectangle (axis cs:0.927154362201691,0.30601420546585);
\draw[draw=black,fill=c2,opacity=0.8] (axis cs:0.927154362201691,0) rectangle (axis cs:0.946663200855255,0.172742214943905);
\draw[draw=black,fill=c2,opacity=0.8] (axis cs:0.946663200855255,0) rectangle (axis cs:0.966172099113464,0.0825264440201036);
\draw[draw=black,fill=c2,opacity=0.8] (axis cs:0.966172099113464,0) rectangle (axis cs:0.985680937767029,0.0179405861217706);
\end{axis}

\end{tikzpicture}}}
      \caption{Logit-Normal (Baseline)}
    \end{subfigure}\hfill
    \begin{subfigure}[b]{0.32\linewidth}
      \centering
        \resizebox{\linewidth}{!}{{
\begin{tikzpicture}
\definecolor{c2}{RGB}{140,150,198}

\begin{axis}[
    every axis y label/.style={at={(current axis.north west)},above=0mm},
    width=8cm, height=5cm,
    tick align=outside,
    tick pos=left,
    xlabel={Sampled $t$},
    xmin=0, xmax=1.0,
    ylabel={Density},
    ymajorgrids,
    ymin=0, ymax=2.0,
    xtick={0, 0.2, 0.4, 0.6, 0.8, 1},
    ytick={0, 0.3, 0.6, 0.9, 1.2, 1.5, 1.8}
]
\draw[draw=black,fill=c2,opacity=0.8] (axis cs:9.22679919312941e-06,0) rectangle (axis cs:0.0200090054422617,0.547506080104503);
\draw[draw=black,fill=c2,opacity=0.8] (axis cs:0.0200090054422617,0) rectangle (axis cs:0.0400087833404541,0.561506235577495);
\draw[draw=black,fill=c2,opacity=0.8] (axis cs:0.0400087833404541,0) rectangle (axis cs:0.0600085631012917,0.601506623765751);
\draw[draw=black,fill=c2,opacity=0.8] (axis cs:0.0600085631012917,0) rectangle (axis cs:0.0800083428621292,0.624506877043577);
\draw[draw=black,fill=c2,opacity=0.8] (axis cs:0.0800083428621292,0) rectangle (axis cs:0.100008122622967,0.638007025706649);
\draw[draw=black,fill=c2,opacity=0.8] (axis cs:0.100008122622967,0) rectangle (axis cs:0.120007894933224,0.67600769600093);
\draw[draw=black,fill=c2,opacity=0.8] (axis cs:0.120007894933224,0) rectangle (axis cs:0.140007674694061,0.682007510238142);
\draw[draw=black,fill=c2,opacity=0.8] (axis cs:0.140007674694061,0) rectangle (axis cs:0.160007461905479,0.686007298726012);
\draw[draw=black,fill=c2,opacity=0.8] (axis cs:0.160007461905479,0) rectangle (axis cs:0.180007234215736,0.713508122924058);
\draw[draw=black,fill=c2,opacity=0.8] (axis cs:0.180007234215736,0) rectangle (axis cs:0.200007006525993,0.701507986308657);
\draw[draw=black,fill=c2,opacity=0.8] (axis cs:0.200007006525993,0) rectangle (axis cs:0.220006793737411,0.783008330761614);
\draw[draw=black,fill=c2,opacity=0.8] (axis cs:0.220006793737411,0) rectangle (axis cs:0.240006566047668,0.812009244308809);
\draw[draw=black,fill=c2,opacity=0.8] (axis cs:0.240006566047668,0) rectangle (axis cs:0.260006338357925,0.852509705385788);
\draw[draw=black,fill=c2,opacity=0.8] (axis cs:0.260006338357925,0) rectangle (axis cs:0.280006140470505,0.912009023755666);
\draw[draw=black,fill=c2,opacity=0.8] (axis cs:0.280006140470505,0) rectangle (axis cs:0.300005912780762,0.961010940616707);
\draw[draw=black,fill=c2,opacity=0.8] (axis cs:0.300005912780762,0) rectangle (axis cs:0.320005685091019,1.04301187415528);
\draw[draw=black,fill=c2,opacity=0.8] (axis cs:0.320005685091019,0) rectangle (axis cs:0.340005457401276,1.10351256292459);
\draw[draw=black,fill=c2,opacity=0.8] (axis cs:0.340005457401276,0) rectangle (axis cs:0.360005229711533,1.20001366154011);
\draw[draw=black,fill=c2,opacity=0.8] (axis cs:0.360005229711533,0) rectangle (axis cs:0.380005031824112,1.24051227408871);
\draw[draw=black,fill=c2,opacity=0.8] (axis cs:0.380005031824112,0) rectangle (axis cs:0.400004804134369,1.34151527246338);
\draw[draw=black,fill=c2,opacity=0.8] (axis cs:0.400004804134369,0) rectangle (axis cs:0.420004576444626,1.4620166443097);
\draw[draw=black,fill=c2,opacity=0.8] (axis cs:0.420004576444626,0) rectangle (axis cs:0.440004348754883,1.59501815846373);
\draw[draw=black,fill=c2,opacity=0.8] (axis cs:0.440004348754883,0) rectangle (axis cs:0.46000412106514,1.68101913754077);
\draw[draw=black,fill=c2,opacity=0.8] (axis cs:0.46000412106514,0) rectangle (axis cs:0.480003923177719,1.77401755278789);
\draw[draw=black,fill=c2,opacity=0.8] (axis cs:0.480003923177719,0) rectangle (axis cs:0.500003695487976,1.8495210558487);
\draw[draw=black,fill=c2,opacity=0.8] (axis cs:0.500003695487976,0) rectangle (axis cs:0.520003497600555,1.83101811677261);
\draw[draw=black,fill=c2,opacity=0.8] (axis cs:0.520003497600555,0) rectangle (axis cs:0.54000324010849,1.75902264671872);
\draw[draw=black,fill=c2,opacity=0.8] (axis cs:0.54000324010849,0) rectangle (axis cs:0.560003042221069,1.68251664744398);
\draw[draw=black,fill=c2,opacity=0.8] (axis cs:0.560003042221069,0) rectangle (axis cs:0.580002784729004,1.56202011038922);
\draw[draw=black,fill=c2,opacity=0.8] (axis cs:0.580002784729004,0) rectangle (axis cs:0.600002586841583,1.46251447066081);
\draw[draw=black,fill=c2,opacity=0.8] (axis cs:0.600002586841583,0) rectangle (axis cs:0.620002388954163,1.34501330806071);
\draw[draw=black,fill=c2,opacity=0.8] (axis cs:0.620002388954163,0) rectangle (axis cs:0.640002131462097,1.23701592608929);
\draw[draw=black,fill=c2,opacity=0.8] (axis cs:0.640002131462097,0) rectangle (axis cs:0.660001933574677,1.21051197725464);
\draw[draw=black,fill=c2,opacity=0.8] (axis cs:0.660001933574677,0) rectangle (axis cs:0.680001676082611,1.07951389831317);
\draw[draw=black,fill=c2,opacity=0.8] (axis cs:0.680001676082611,0) rectangle (axis cs:0.70000147819519,1.00550994888851);
\draw[draw=black,fill=c2,opacity=0.8] (axis cs:0.70000147819519,0) rectangle (axis cs:0.72000128030777,0.952509424481658);
\draw[draw=black,fill=c2,opacity=0.8] (axis cs:0.72000128030777,0) rectangle (axis cs:0.740001022815704,0.884011381295822);
\draw[draw=black,fill=c2,opacity=0.8] (axis cs:0.740001022815704,0) rectangle (axis cs:0.760000824928284,0.92250912764759);
\draw[draw=black,fill=c2,opacity=0.8] (axis cs:0.760000824928284,0) rectangle (axis cs:0.780000567436218,0.825510628121833);
\draw[draw=black,fill=c2,opacity=0.8] (axis cs:0.780000567436218,0) rectangle (axis cs:0.800000369548798,0.800007915575146);
\draw[draw=black,fill=c2,opacity=0.8] (axis cs:0.800000369548798,0) rectangle (axis cs:0.820000171661377,0.751007430746168);
\draw[draw=black,fill=c2,opacity=0.8] (axis cs:0.820000171661377,0) rectangle (axis cs:0.839999914169312,0.717009231209394);
\draw[draw=black,fill=c2,opacity=0.8] (axis cs:0.839999914169312,0) rectangle (axis cs:0.859999716281891,0.684506772763984);
\draw[draw=black,fill=c2,opacity=0.8] (axis cs:0.859999716281891,0) rectangle (axis cs:0.879999458789825,0.668508606783096);
\draw[draw=black,fill=c2,opacity=0.8] (axis cs:0.879999458789825,0) rectangle (axis cs:0.899999260902405,0.655006480877151);
\draw[draw=black,fill=c2,opacity=0.8] (axis cs:0.899999260902405,0) rectangle (axis cs:0.919999063014984,0.591505852578373);
\draw[draw=black,fill=c2,opacity=0.8] (axis cs:0.919999063014984,0) rectangle (axis cs:0.939998805522919,0.584507525302497);
\draw[draw=black,fill=c2,opacity=0.8] (axis cs:0.939998805522919,0) rectangle (axis cs:0.959998607635498,0.589005827842201);
\draw[draw=black,fill=c2,opacity=0.8] (axis cs:0.959998607635498,0) rectangle (axis cs:0.979998350143433,0.549007068248197);
\draw[draw=black,fill=c2,opacity=0.8] (axis cs:0.979998350143433,0) rectangle (axis cs:0.999998152256012,0.608506020784345);
\end{axis}

\end{tikzpicture}}}
      \caption{Mode [Scale=0.8, Shift=1]}
    \end{subfigure}
    \begin{subfigure}[b]{0.32\linewidth}
      \centering
        \resizebox{\linewidth}{!}{{
\begin{tikzpicture}
\definecolor{c2}{RGB}{140,150,198}

\begin{axis}[
    every axis y label/.style={at={(current axis.north west)},above=0mm},
    width=8cm, height=5cm,
    tick align=outside,
    tick pos=left,
    xlabel={Sampled $t$},
    xmin=0, xmax=1.0,
    ylabel={Density},
    ymajorgrids,
    ymin=0, ymax=2.8,
    xtick={0, 0.2, 0.4, 0.6, 0.8, 1},
    ytick={0, 0.5, 1.0, 1.5, 2.0, 2.5}
]
\draw[draw=black,fill=c2,opacity=0.8] (axis cs:0.000149652332765982,0) rectangle (axis cs:0.0201464872807264,0.19353062700691);
\draw[draw=black,fill=c2,opacity=0.8] (axis cs:0.0201464872807264,0) rectangle (axis cs:0.0401433221995831,0.208533001193492);
\draw[draw=black,fill=c2,opacity=0.8] (axis cs:0.0401433221995831,0) rectangle (axis cs:0.0601401552557945,0.177528111077433);
\draw[draw=black,fill=c2,opacity=0.8] (axis cs:0.0601401552557945,0) rectangle (axis cs:0.0801369920372963,0.206532665397384);
\draw[draw=black,fill=c2,opacity=0.8] (axis cs:0.0801369920372963,0) rectangle (axis cs:0.100133821368217,0.224535596403631);
\draw[draw=black,fill=c2,opacity=0.8] (axis cs:0.100133821368217,0) rectangle (axis cs:0.120130658149719,0.225035591837343);
\draw[draw=black,fill=c2,opacity=0.8] (axis cs:0.120130658149719,0) rectangle (axis cs:0.140127494931221,0.233536936417865);
\draw[draw=black,fill=c2,opacity=0.8] (axis cs:0.140127494931221,0) rectangle (axis cs:0.160124331712723,0.236037331882724);
\draw[draw=black,fill=c2,opacity=0.8] (axis cs:0.160124331712723,0) rectangle (axis cs:0.180121153593063,0.243038620290857);
\draw[draw=black,fill=c2,opacity=0.8] (axis cs:0.180121153593063,0) rectangle (axis cs:0.200117990374565,0.270042710204812);
\draw[draw=black,fill=c2,opacity=0.8] (axis cs:0.200117990374565,0) rectangle (axis cs:0.220114827156067,0.26704223564698);
\draw[draw=black,fill=c2,opacity=0.8] (axis cs:0.220114827156067,0) rectangle (axis cs:0.240111663937569,0.297547060318265);
\draw[draw=black,fill=c2,opacity=0.8] (axis cs:0.240111663937569,0) rectangle (axis cs:0.26010850071907,0.306548483991758);
\draw[draw=black,fill=c2,opacity=0.8] (axis cs:0.26010850071907,0) rectangle (axis cs:0.280105322599411,0.365058009901905);
\draw[draw=black,fill=c2,opacity=0.8] (axis cs:0.280105322599411,0) rectangle (axis cs:0.300102174282074,0.344054159583775);
\draw[draw=black,fill=c2,opacity=0.8] (axis cs:0.300102174282074,0) rectangle (axis cs:0.320098996162415,0.359557135780095);
\draw[draw=black,fill=c2,opacity=0.8] (axis cs:0.320098996162415,0) rectangle (axis cs:0.340095818042755,0.370058804558095);
\draw[draw=black,fill=c2,opacity=0.8] (axis cs:0.340095818042755,0) rectangle (axis cs:0.360092669725418,0.394562110336625);
\draw[draw=black,fill=c2,opacity=0.8] (axis cs:0.360092669725418,0) rectangle (axis cs:0.380089491605759,0.431568578829238);
\draw[draw=black,fill=c2,opacity=0.8] (axis cs:0.380089491605759,0) rectangle (axis cs:0.400086343288422,0.456071792936632);
\draw[draw=black,fill=c2,opacity=0.8] (axis cs:0.400086343288422,0) rectangle (axis cs:0.420083165168762,0.473575253941238);
\draw[draw=black,fill=c2,opacity=0.8] (axis cs:0.420083165168762,0) rectangle (axis cs:0.440080016851425,0.510080294731759);
\draw[draw=black,fill=c2,opacity=0.8] (axis cs:0.440080016851425,0) rectangle (axis cs:0.460076838731766,0.571090749736953);
\draw[draw=black,fill=c2,opacity=0.8] (axis cs:0.460076838731766,0) rectangle (axis cs:0.480073660612106,0.583592736377429);
\draw[draw=black,fill=c2,opacity=0.8] (axis cs:0.480073660612106,0) rectangle (axis cs:0.500070512294769,0.608595802635834);
\draw[draw=black,fill=c2,opacity=0.8] (axis cs:0.500070512294769,0) rectangle (axis cs:0.52006733417511,0.681108232173143);
\draw[draw=black,fill=c2,opacity=0.8] (axis cs:0.52006733417511,0) rectangle (axis cs:0.54006415605545,0.737117132322477);
\draw[draw=black,fill=c2,opacity=0.8] (axis cs:0.54006415605545,0) rectangle (axis cs:0.560060977935791,0.799627065524858);
\draw[draw=black,fill=c2,opacity=0.8] (axis cs:0.560060977935791,0) rectangle (axis cs:0.580057859420776,0.885138015809617);
\draw[draw=black,fill=c2,opacity=0.8] (axis cs:0.580057859420776,0) rectangle (axis cs:0.600054681301117,0.964153209713524);
\draw[draw=black,fill=c2,opacity=0.8] (axis cs:0.600054681301117,0) rectangle (axis cs:0.620051503181458,1.07167029482162);
\draw[draw=black,fill=c2,opacity=0.8] (axis cs:0.620051503181458,0) rectangle (axis cs:0.640048325061798,1.20819198893562);
\draw[draw=black,fill=c2,opacity=0.8] (axis cs:0.640048325061798,0) rectangle (axis cs:0.660045206546783,1.41572074732035);
\draw[draw=black,fill=c2,opacity=0.8] (axis cs:0.660045206546783,0) rectangle (axis cs:0.680042028427124,1.54524554876286);
\draw[draw=black,fill=c2,opacity=0.8] (axis cs:0.680042028427124,0) rectangle (axis cs:0.700038850307465,1.78228321546629);
\draw[draw=black,fill=c2,opacity=0.8] (axis cs:0.700038850307465,0) rectangle (axis cs:0.720035672187805,2.04832549117562);
\draw[draw=black,fill=c2,opacity=0.8] (axis cs:0.720035672187805,0) rectangle (axis cs:0.740032494068146,2.24635695956076);
\draw[draw=black,fill=c2,opacity=0.8] (axis cs:0.740032494068146,0) rectangle (axis cs:0.760029375553131,2.41887716523792);
\draw[draw=black,fill=c2,opacity=0.8] (axis cs:0.760029375553131,0) rectangle (axis cs:0.780026197433472,2.48939557985162);
\draw[draw=black,fill=c2,opacity=0.8] (axis cs:0.780026197433472,0) rectangle (axis cs:0.800023019313812,2.40388199123076);
\draw[draw=black,fill=c2,opacity=0.8] (axis cs:0.800023019313812,0) rectangle (axis cs:0.820019841194153,2.34787309108143);
\draw[draw=black,fill=c2,opacity=0.8] (axis cs:0.820019841194153,0) rectangle (axis cs:0.840016663074493,2.17884623170219);
\draw[draw=black,fill=c2,opacity=0.8] (axis cs:0.840016663074493,0) rectangle (axis cs:0.860013544559479,2.08132453209018);
\draw[draw=black,fill=c2,opacity=0.8] (axis cs:0.860013544559479,0) rectangle (axis cs:0.880010366439819,1.88930022110876);
\draw[draw=black,fill=c2,opacity=0.8] (axis cs:0.880010366439819,0) rectangle (axis cs:0.90000718832016,1.78728401012248);
\draw[draw=black,fill=c2,opacity=0.8] (axis cs:0.90000718832016,0) rectangle (axis cs:0.9200040102005,1.806287029816);
\draw[draw=black,fill=c2,opacity=0.8] (axis cs:0.9200040102005,0) rectangle (axis cs:0.940000832080841,1.68726811699867);
\draw[draw=black,fill=c2,opacity=0.8] (axis cs:0.940000832080841,0) rectangle (axis cs:0.959997713565826,1.69626449131425);
\draw[draw=black,fill=c2,opacity=0.8] (axis cs:0.959997713565826,0) rectangle (axis cs:0.979994535446167,1.64926207761162);
\draw[draw=black,fill=c2,opacity=0.8] (axis cs:0.979994535446167,0) rectangle (axis cs:0.999991357326508,1.62975897845248);
\end{axis}

\end{tikzpicture}}}
      \caption{Mode [Scale=0.8, Shift=3]}
    \end{subfigure}\\
    \begin{subfigure}[b]{0.32\linewidth}
      \centering
        \resizebox{\linewidth}{!}{{
\begin{tikzpicture}
\definecolor{c2}{RGB}{140,150,198}

\begin{axis}[
    every axis y label/.style={at={(current axis.north west)},above=0mm},
    width=8cm, height=5cm,
    tick align=outside,
    tick pos=left,
    xlabel={Sampled $t$},
    xmin=0, xmax=1.0,
    ylabel={Density},
    ymajorgrids,
    ymin=0, ymax=1.15,
    xtick={0, 0.2, 0.4, 0.6, 0.8, 1},
    ytick={0, 0.2, 0.4, 0.6, 0.8, 1.0}
]
\draw[draw=black,fill=c2,opacity=0.8] (axis cs:4.17232513427734e-07,0) rectangle (axis cs:0.0200003180652857,0.965504787321657);
\draw[draw=black,fill=c2,opacity=0.8] (axis cs:0.0200003180652857,0) rectangle (axis cs:0.0400002188980579,0.96600478980085);
\draw[draw=black,fill=c2,opacity=0.8] (axis cs:0.0400002188980579,0) rectangle (axis cs:0.0600001215934753,1.04500508418918);
\draw[draw=black,fill=c2,opacity=0.8] (axis cs:0.0600001215934753,0) rectangle (axis cs:0.0800000205636024,0.985504978272135);
\draw[draw=black,fill=c2,opacity=0.8] (axis cs:0.0800000205636024,0) rectangle (axis cs:0.0999999195337296,1.01050510456011);
\draw[draw=black,fill=c2,opacity=0.8] (axis cs:0.0999999195337296,0) rectangle (axis cs:0.119999825954437,0.970004538616912);
\draw[draw=black,fill=c2,opacity=0.8] (axis cs:0.119999825954437,0) rectangle (axis cs:0.139999717473984,1.00450544846033);
\draw[draw=black,fill=c2,opacity=0.8] (axis cs:0.139999717473984,0) rectangle (axis cs:0.159999623894691,1.01200473513435);
\draw[draw=black,fill=c2,opacity=0.8] (axis cs:0.159999623894691,0) rectangle (axis cs:0.179999530315399,0.994504653252082);
\draw[draw=black,fill=c2,opacity=0.8] (axis cs:0.179999530315399,0) rectangle (axis cs:0.199999421834946,1.0135054972768);
\draw[draw=black,fill=c2,opacity=0.8] (axis cs:0.199999421834946,0) rectangle (axis cs:0.219999328255653,0.968504531598433);
\draw[draw=black,fill=c2,opacity=0.8] (axis cs:0.219999328255653,0) rectangle (axis cs:0.239999234676361,0.966004519900966);
\draw[draw=black,fill=c2,opacity=0.8] (axis cs:0.239999234676361,0) rectangle (axis cs:0.259999126195908,0.988505361675496);
\draw[draw=black,fill=c2,opacity=0.8] (axis cs:0.259999126195908,0) rectangle (axis cs:0.279999017715454,0.987005353539418);
\draw[draw=black,fill=c2,opacity=0.8] (axis cs:0.279999017715454,0) rectangle (axis cs:0.299998939037323,1.05900416602343);
\draw[draw=black,fill=c2,opacity=0.8] (axis cs:0.299998939037323,0) rectangle (axis cs:0.31999883055687,0.989005364387522);
\draw[draw=black,fill=c2,opacity=0.8] (axis cs:0.31999883055687,0) rectangle (axis cs:0.339998722076416,0.983505334555235);
\draw[draw=black,fill=c2,opacity=0.8] (axis cs:0.339998722076416,0) rectangle (axis cs:0.359998643398285,0.981003859177512);
\draw[draw=black,fill=c2,opacity=0.8] (axis cs:0.359998643398285,0) rectangle (axis cs:0.379998534917831,1.01000547829262);
\draw[draw=black,fill=c2,opacity=0.8] (axis cs:0.379998534917831,0) rectangle (axis cs:0.399998426437378,1.00650545930843);
\draw[draw=black,fill=c2,opacity=0.8] (axis cs:0.399998426437378,0) rectangle (axis cs:0.419998347759247,0.995503916219382);
\draw[draw=black,fill=c2,opacity=0.8] (axis cs:0.419998347759247,0) rectangle (axis cs:0.439998239278793,1.03400560846987);
\draw[draw=black,fill=c2,opacity=0.8] (axis cs:0.439998239278793,0) rectangle (axis cs:0.45999813079834,1.01150548642869);
\draw[draw=black,fill=c2,opacity=0.8] (axis cs:0.45999813079834,0) rectangle (axis cs:0.479998052120209,1.01700400079871);
\draw[draw=black,fill=c2,opacity=0.8] (axis cs:0.479998052120209,0) rectangle (axis cs:0.499997943639755,1.01100548371667);
\draw[draw=black,fill=c2,opacity=0.8] (axis cs:0.499997943639755,0) rectangle (axis cs:0.519997835159302,0.976005293874844);
\draw[draw=black,fill=c2,opacity=0.8] (axis cs:0.519997835159302,0) rectangle (axis cs:0.539997756481171,1.01100397719517);
\draw[draw=black,fill=c2,opacity=0.8] (axis cs:0.539997756481171,0) rectangle (axis cs:0.559997618198395,0.968506696389722);
\draw[draw=black,fill=c2,opacity=0.8] (axis cs:0.559997618198395,0) rectangle (axis cs:0.579997539520264,1.02850404603881);
\draw[draw=black,fill=c2,opacity=0.8] (axis cs:0.579997539520264,0) rectangle (axis cs:0.599997460842133,1.0235040263692);
\draw[draw=black,fill=c2,opacity=0.8] (axis cs:0.599997460842133,0) rectangle (axis cs:0.619997322559357,1.02150706284161);
\draw[draw=black,fill=c2,opacity=0.8] (axis cs:0.619997322559357,0) rectangle (axis cs:0.639997243881226,1.01500399293086);
\draw[draw=black,fill=c2,opacity=0.8] (axis cs:0.639997243881226,0) rectangle (axis cs:0.659997165203094,1.04650411684941);
\draw[draw=black,fill=c2,opacity=0.8] (axis cs:0.659997165203094,0) rectangle (axis cs:0.679997026920319,0.986006817387987);
\draw[draw=black,fill=c2,opacity=0.8] (axis cs:0.679997026920319,0) rectangle (axis cs:0.699996948242188,1.01000397326125);
\draw[draw=black,fill=c2,opacity=0.8] (axis cs:0.699996948242188,0) rectangle (axis cs:0.719996869564056,0.9850038749132);
\draw[draw=black,fill=c2,opacity=0.8] (axis cs:0.719996869564056,0) rectangle (axis cs:0.739996731281281,0.987006824302174);
\draw[draw=black,fill=c2,opacity=0.8] (axis cs:0.739996731281281,0) rectangle (axis cs:0.759996652603149,1.03550407357626);
\draw[draw=black,fill=c2,opacity=0.8] (axis cs:0.759996652603149,0) rectangle (axis cs:0.779996573925018,0.993003906384577);
\draw[draw=black,fill=c2,opacity=0.8] (axis cs:0.779996573925018,0) rectangle (axis cs:0.799996435642242,0.975006741331935);
\draw[draw=black,fill=c2,opacity=0.8] (axis cs:0.799996435642242,0) rectangle (axis cs:0.819996356964111,0.96500379623476);
\draw[draw=black,fill=c2,opacity=0.8] (axis cs:0.819996356964111,0) rectangle (axis cs:0.83999627828598,0.991503900483694);
\draw[draw=black,fill=c2,opacity=0.8] (axis cs:0.83999627828598,0) rectangle (axis cs:0.859996140003204,1.00100692110079);
\draw[draw=black,fill=c2,opacity=0.8] (axis cs:0.859996140003204,0) rectangle (axis cs:0.879996061325073,1.01800400473263);
\draw[draw=black,fill=c2,opacity=0.8] (axis cs:0.879996061325073,0) rectangle (axis cs:0.899995982646942,1.01800400473263);
\draw[draw=black,fill=c2,opacity=0.8] (axis cs:0.899995982646942,0) rectangle (axis cs:0.919995844364166,1.02150706284161);
\draw[draw=black,fill=c2,opacity=0.8] (axis cs:0.919995844364166,0) rectangle (axis cs:0.939995765686035,0.9555037588625);
\draw[draw=black,fill=c2,opacity=0.8] (axis cs:0.939995765686035,0) rectangle (axis cs:0.959995687007904,1.00650395949252);
\draw[draw=black,fill=c2,opacity=0.8] (axis cs:0.959995687007904,0) rectangle (axis cs:0.979995548725128,1.01550702135649);
\draw[draw=black,fill=c2,opacity=0.8] (axis cs:0.979995548725128,0) rectangle (axis cs:0.999995470046997,0.970503817871331);
\end{axis}

\end{tikzpicture}}}
      \caption{Uniform [Shift=1]}
    \end{subfigure}\hfill
    \begin{subfigure}[b]{0.32\linewidth}
      \centering
        \resizebox{\linewidth}{!}{{
\begin{tikzpicture}
\definecolor{c2}{RGB}{140,150,198}

\begin{axis}[
    every axis y label/.style={at={(current axis.north west)},above=0mm},
    width=8cm, height=5cm,
    tick align=outside,
    tick pos=left,
    xlabel={Sampled $t$},
    xmin=0, xmax=1.0,
    ylabel={Density},
    ymajorgrids,
    ymin=0, ymax=3.3,
    xtick={0, 0.2, 0.4, 0.6, 0.8, 1},
    ytick={0, 0.6, 1.2, 1.8, 2.4, 3.0}
]
\draw[draw=black,fill=c2,opacity=0.8] (axis cs:8.5110601503402e-05,0) rectangle (axis cs:0.0200832895934582,0.327029763818383);
\draw[draw=black,fill=c2,opacity=0.8] (axis cs:0.0200832895934582,0) rectangle (axis cs:0.0400814712047577,0.34453132459338);
\draw[draw=black,fill=c2,opacity=0.8] (axis cs:0.0400814712047577,0) rectangle (axis cs:0.0600796490907669,0.355532391027193);
\draw[draw=black,fill=c2,opacity=0.8] (axis cs:0.0600796490907669,0) rectangle (axis cs:0.0800778269767761,0.346531571001188);
\draw[draw=black,fill=c2,opacity=0.8] (axis cs:0.0800778269767761,0) rectangle (axis cs:0.100076012313366,0.371533710431499);
\draw[draw=black,fill=c2,opacity=0.8] (axis cs:0.100076012313366,0) rectangle (axis cs:0.120074190199375,0.381534759991208);
\draw[draw=black,fill=c2,opacity=0.8] (axis cs:0.120074190199375,0) rectangle (axis cs:0.140072375535965,0.396535978966593);
\draw[draw=black,fill=c2,opacity=0.8] (axis cs:0.140072375535965,0) rectangle (axis cs:0.160070553421974,0.391035625574213);
\draw[draw=black,fill=c2,opacity=0.8] (axis cs:0.160070553421974,0) rectangle (axis cs:0.180068731307983,0.425538769007231);
\draw[draw=black,fill=c2,opacity=0.8] (axis cs:0.180068731307983,0) rectangle (axis cs:0.200066909193993,0.442540317945241);
\draw[draw=black,fill=c2,opacity=0.8] (axis cs:0.200066909193993,0) rectangle (axis cs:0.220065087080002,0.449540955743244);
\draw[draw=black,fill=c2,opacity=0.8] (axis cs:0.220065087080002,0) rectangle (axis cs:0.240063264966011,0.450041001300245);
\draw[draw=black,fill=c2,opacity=0.8] (axis cs:0.240063264966011,0) rectangle (axis cs:0.26006144285202,0.450041001300245);
\draw[draw=black,fill=c2,opacity=0.8] (axis cs:0.26006144285202,0) rectangle (axis cs:0.280059635639191,0.505545681432071);
\draw[draw=black,fill=c2,opacity=0.8] (axis cs:0.280059635639191,0) rectangle (axis cs:0.300057798624039,0.513047123767002);
\draw[draw=black,fill=c2,opacity=0.8] (axis cs:0.300057798624039,0) rectangle (axis cs:0.320055991411209,0.519546946595373);
\draw[draw=black,fill=c2,opacity=0.8] (axis cs:0.320055991411209,0) rectangle (axis cs:0.340054154396057,0.545550109190837);
\draw[draw=black,fill=c2,opacity=0.8] (axis cs:0.340054154396057,0) rectangle (axis cs:0.360052347183228,0.578052233170598);
\draw[draw=black,fill=c2,opacity=0.8] (axis cs:0.360052347183228,0) rectangle (axis cs:0.380050539970398,0.576052052432983);
\draw[draw=black,fill=c2,opacity=0.8] (axis cs:0.380050539970398,0) rectangle (axis cs:0.400048702955246,0.618556814912068);
\draw[draw=black,fill=c2,opacity=0.8] (axis cs:0.400048702955246,0) rectangle (axis cs:0.420046895742416,0.641057926405456);
\draw[draw=black,fill=c2,opacity=0.8] (axis cs:0.420046895742416,0) rectangle (axis cs:0.440045058727264,0.671061637519802);
\draw[draw=black,fill=c2,opacity=0.8] (axis cs:0.440045058727264,0) rectangle (axis cs:0.460043251514435,0.672060727838481);
\draw[draw=black,fill=c2,opacity=0.8] (axis cs:0.460043251514435,0) rectangle (axis cs:0.480041414499283,0.69706402585887);
\draw[draw=black,fill=c2,opacity=0.8] (axis cs:0.480041414499283,0) rectangle (axis cs:0.500039577484131,0.756569491480969);
\draw[draw=black,fill=c2,opacity=0.8] (axis cs:0.500039577484131,0) rectangle (axis cs:0.520037770271301,0.765069132137556);
\draw[draw=black,fill=c2,opacity=0.8] (axis cs:0.520037770271301,0) rectangle (axis cs:0.540035963058472,0.791071481726545);
\draw[draw=black,fill=c2,opacity=0.8] (axis cs:0.540035963058472,0) rectangle (axis cs:0.560034155845642,0.849076723117366);
\draw[draw=black,fill=c2,opacity=0.8] (axis cs:0.560034155845642,0) rectangle (axis cs:0.580032348632812,0.875079072706355);
\draw[draw=black,fill=c2,opacity=0.8] (axis cs:0.580032348632812,0) rectangle (axis cs:0.600030481815338,0.899583967953552);
\draw[draw=black,fill=c2,opacity=0.8] (axis cs:0.600030481815338,0) rectangle (axis cs:0.620028674602509,0.960586799239376);
\draw[draw=black,fill=c2,opacity=0.8] (axis cs:0.620028674602509,0) rectangle (axis cs:0.640026867389679,0.996590052516437);
\draw[draw=black,fill=c2,opacity=0.8] (axis cs:0.640026867389679,0) rectangle (axis cs:0.660025060176849,1.06159592648891);
\draw[draw=black,fill=c2,opacity=0.8] (axis cs:0.660025060176849,0) rectangle (axis cs:0.680023193359375,1.09360207777344);
\draw[draw=black,fill=c2,opacity=0.8] (axis cs:0.680023193359375,0) rectangle (axis cs:0.700021386146545,1.13460252341184);
\draw[draw=black,fill=c2,opacity=0.8] (axis cs:0.700021386146545,0) rectangle (axis cs:0.720019578933716,1.2121095269944);
\draw[draw=black,fill=c2,opacity=0.8] (axis cs:0.720019578933716,0) rectangle (axis cs:0.740017771720886,1.26161400025036);
\draw[draw=black,fill=c2,opacity=0.8] (axis cs:0.740017771720886,0) rectangle (axis cs:0.760015964508057,1.32461969348522);
\draw[draw=black,fill=c2,opacity=0.8] (axis cs:0.760015964508057,0) rectangle (axis cs:0.780014097690582,1.4211326497632);
\draw[draw=black,fill=c2,opacity=0.8] (axis cs:0.780014097690582,0) rectangle (axis cs:0.800012290477753,1.51563695392741);
\draw[draw=black,fill=c2,opacity=0.8] (axis cs:0.800012290477753,0) rectangle (axis cs:0.820010483264923,1.62414675894299);
\draw[draw=black,fill=c2,opacity=0.8] (axis cs:0.820010483264923,0) rectangle (axis cs:0.840008676052094,1.68315209070262);
\draw[draw=black,fill=c2,opacity=0.8] (axis cs:0.840008676052094,0) rectangle (axis cs:0.860006868839264,1.80866343198794);
\draw[draw=black,fill=c2,opacity=0.8] (axis cs:0.860006868839264,0) rectangle (axis cs:0.88000500202179,1.90867815768689);
\draw[draw=black,fill=c2,opacity=0.8] (axis cs:0.88000500202179,0) rectangle (axis cs:0.90000319480896,1.96317739396866);
\draw[draw=black,fill=c2,opacity=0.8] (axis cs:0.90000319480896,0) rectangle (axis cs:0.92000138759613,2.14669397664479);
\draw[draw=black,fill=c2,opacity=0.8] (axis cs:0.92000138759613,0) rectangle (axis cs:0.939999580383301,2.25370364610717);
\draw[draw=black,fill=c2,opacity=0.8] (axis cs:0.939999580383301,0) rectangle (axis cs:0.959997713565826,2.4867321137482);
\draw[draw=black,fill=c2,opacity=0.8] (axis cs:0.959997713565826,0) rectangle (axis cs:0.979995906352997,2.66174051658053);
\draw[draw=black,fill=c2,opacity=0.8] (axis cs:0.979995906352997,0) rectangle (axis cs:0.999994099140167,2.87976021698051);
\end{axis}

\end{tikzpicture}}}
      \caption{Uniform [Shift=3]}
    \end{subfigure}
    \begin{subfigure}[b]{0.32\linewidth}
      \centering
        \resizebox{\linewidth}{!}{{
\begin{tikzpicture}
\definecolor{c2}{RGB}{140,150,198}

\begin{axis}[
    every axis y label/.style={at={(current axis.north west)},above=0mm},
    width=8cm, height=5cm,
    tick align=outside,
    tick pos=left,
    xlabel={Sampled $t$},
    xmin=0, xmax=1.0,
    ylabel={Density},
    ymajorgrids,
    ymin=0, ymax=5.2,
    xtick={0, 0.2, 0.4, 0.6, 0.8, 1},
    ytick={0, 0.8, 1.6, 2.4, 3.2, 4.0, 4.8}
]
\draw[draw=black,fill=c2,opacity=0.8] (axis cs:4.5595890696859e-05,0) rectangle (axis cs:0.0200446806848049,0.189008647200451);
\draw[draw=black,fill=c2,opacity=0.8] (axis cs:0.0200446806848049,0) rectangle (axis cs:0.040043767541647,0.216009862396395);
\draw[draw=black,fill=c2,opacity=0.8] (axis cs:0.040043767541647,0) rectangle (axis cs:0.060042854398489,0.232510615773897);
\draw[draw=black,fill=c2,opacity=0.8] (axis cs:0.060042854398489,0) rectangle (axis cs:0.0800419375300407,0.216509925555974);
\draw[draw=black,fill=c2,opacity=0.8] (axis cs:0.0800419375300407,0) rectangle (axis cs:0.100041024386883,0.242011049536702);
\draw[draw=black,fill=c2,opacity=0.8] (axis cs:0.100041024386883,0) rectangle (axis cs:0.120040111243725,0.228010410307306);
\draw[draw=black,fill=c2,opacity=0.8] (axis cs:0.120040111243725,0) rectangle (axis cs:0.140039190649986,0.250011507951439);
\draw[draw=black,fill=c2,opacity=0.8] (axis cs:0.140039190649986,0) rectangle (axis cs:0.160038277506828,0.251011460469885);
\draw[draw=black,fill=c2,opacity=0.8] (axis cs:0.160038277506828,0) rectangle (axis cs:0.18003736436367,0.273512487802843);
\draw[draw=black,fill=c2,opacity=0.8] (axis cs:0.18003736436367,0) rectangle (axis cs:0.200036451220512,0.259511848573447);
\draw[draw=black,fill=c2,opacity=0.8] (axis cs:0.200036451220512,0) rectangle (axis cs:0.220035538077354,0.300013697772771);
\draw[draw=black,fill=c2,opacity=0.8] (axis cs:0.220035538077354,0) rectangle (axis cs:0.240034624934196,0.318514542468758);
\draw[draw=black,fill=c2,opacity=0.8] (axis cs:0.240034624934196,0) rectangle (axis cs:0.260033696889877,0.330015313441844);
\draw[draw=black,fill=c2,opacity=0.8] (axis cs:0.260033696889877,0) rectangle (axis cs:0.280032783746719,0.317014473979894);
\draw[draw=black,fill=c2,opacity=0.8] (axis cs:0.280032783746719,0) rectangle (axis cs:0.300031870603561,0.341515592631337);
\draw[draw=black,fill=c2,opacity=0.8] (axis cs:0.300031870603561,0) rectangle (axis cs:0.320030957460403,0.360016437327325);
\draw[draw=black,fill=c2,opacity=0.8] (axis cs:0.320030957460403,0) rectangle (axis cs:0.340030044317245,0.373017030897478);
\draw[draw=black,fill=c2,opacity=0.8] (axis cs:0.340030044317245,0) rectangle (axis cs:0.360029131174088,0.406518560482104);
\draw[draw=black,fill=c2,opacity=0.8] (axis cs:0.360029131174088,0) rectangle (axis cs:0.38002821803093,0.398018172378543);
\draw[draw=black,fill=c2,opacity=0.8] (axis cs:0.38002821803093,0) rectangle (axis cs:0.400027304887772,0.423019313859607);
\draw[draw=black,fill=c2,opacity=0.8] (axis cs:0.400027304887772,0) rectangle (axis cs:0.420026391744614,0.418019085563394);
\draw[draw=black,fill=c2,opacity=0.8] (axis cs:0.420026391744614,0) rectangle (axis cs:0.440025478601456,0.46702132286628);
\draw[draw=black,fill=c2,opacity=0.8] (axis cs:0.440025478601456,0) rectangle (axis cs:0.460024565458298,0.491022418688101);
\draw[draw=black,fill=c2,opacity=0.8] (axis cs:0.460024565458298,0) rectangle (axis cs:0.48002365231514,0.519023697146893);
\draw[draw=black,fill=c2,opacity=0.8] (axis cs:0.48002365231514,0) rectangle (axis cs:0.500022709369659,0.563026545166804);
\draw[draw=black,fill=c2,opacity=0.8] (axis cs:0.500022709369659,0) rectangle (axis cs:0.520021796226501,0.601027441204784);
\draw[draw=black,fill=c2,opacity=0.8] (axis cs:0.520021796226501,0) rectangle (axis cs:0.540020883083344,0.578526413871826);
\draw[draw=black,fill=c2,opacity=0.8] (axis cs:0.540020883083344,0) rectangle (axis cs:0.560019969940186,0.612527966286074);
\draw[draw=black,fill=c2,opacity=0.8] (axis cs:0.560019969940186,0) rectangle (axis cs:0.580019056797028,0.730533354076697);
\draw[draw=black,fill=c2,opacity=0.8] (axis cs:0.580019056797028,0) rectangle (axis cs:0.60001814365387,0.714532623528816);
\draw[draw=black,fill=c2,opacity=0.8] (axis cs:0.60001814365387,0) rectangle (axis cs:0.620017230510712,0.762534815172459);
\draw[draw=black,fill=c2,opacity=0.8] (axis cs:0.620017230510712,0) rectangle (axis cs:0.640016317367554,0.812537098134587);
\draw[draw=black,fill=c2,opacity=0.8] (axis cs:0.640016317367554,0) rectangle (axis cs:0.660015404224396,0.904041275955283);
\draw[draw=black,fill=c2,opacity=0.8] (axis cs:0.660015404224396,0) rectangle (axis cs:0.680014491081238,0.923542166310513);
\draw[draw=black,fill=c2,opacity=0.8] (axis cs:0.680014491081238,0) rectangle (axis cs:0.70001357793808,0.999045613583327);
\draw[draw=black,fill=c2,opacity=0.8] (axis cs:0.70001357793808,0) rectangle (axis cs:0.720012664794922,1.10805059044077);
\draw[draw=black,fill=c2,opacity=0.8] (axis cs:0.720012664794922,0) rectangle (axis cs:0.740011751651764,1.15555275925479);
\draw[draw=black,fill=c2,opacity=0.8] (axis cs:0.740011751651764,0) rectangle (axis cs:0.760010838508606,1.21355540749086);
\draw[draw=black,fill=c2,opacity=0.8] (axis cs:0.760010838508606,0) rectangle (axis cs:0.780009925365448,1.32306040717792);
\draw[draw=black,fill=c2,opacity=0.8] (axis cs:0.780009925365448,0) rectangle (axis cs:0.80000901222229,1.45356636570907);
\draw[draw=black,fill=c2,opacity=0.8] (axis cs:0.80000901222229,0) rectangle (axis cs:0.820008099079132,1.60557330591394);
\draw[draw=black,fill=c2,opacity=0.8] (axis cs:0.820008099079132,0) rectangle (axis cs:0.840007185935974,1.79308186702193);
\draw[draw=black,fill=c2,opacity=0.8] (axis cs:0.840007185935974,0) rectangle (axis cs:0.860006272792816,1.91558746027914);
\draw[draw=black,fill=c2,opacity=0.8] (axis cs:0.860006272792816,0) rectangle (axis cs:0.880005359649658,2.22760170596282);
\draw[draw=black,fill=c2,opacity=0.8] (axis cs:0.880005359649658,0) rectangle (axis cs:0.9000044465065,2.30360517606526);
\draw[draw=black,fill=c2,opacity=0.8] (axis cs:0.9000044465065,0) rectangle (axis cs:0.920003533363342,2.69512305165872);
\draw[draw=black,fill=c2,opacity=0.8] (axis cs:0.920003533363342,0) rectangle (axis cs:0.940002620220184,3.06513994557847);
\draw[draw=black,fill=c2,opacity=0.8] (axis cs:0.940002620220184,0) rectangle (axis cs:0.960001707077026,3.3876546706842);
\draw[draw=black,fill=c2,opacity=0.8] (axis cs:0.960001707077026,0) rectangle (axis cs:0.980000793933868,4.06368553633218);
\draw[draw=black,fill=c2,opacity=0.8] (axis cs:0.980000793933868,0) rectangle (axis cs:0.99999988079071,4.66871316017393);
\end{axis}

\end{tikzpicture}}}
      \caption{Uniform [Shift=5]}
    \end{subfigure}
    \caption{\textbf{Probability Densities of Different Timestep Sampler.} We visualise the PDFs for the samplers evaluated in our ablation study, where $t=1$ is the noise end and $t=0$ is the data end. (a) Logit-Normal: The standard baseline, which concentrates probability mass on the middle of the timestep range and has diminished density near the endpoints; (b) Mode: Similar to Logit-Normal, but maintains a positive density at the endpoints. (d) Uniform: A standard uniform distribution, which samples all timesteps with equal probability; (c), (e), and (f) show the Mode and Uniform distributions shifted towards the high-noise end of the interval to study the effect of noise-biased sampling.
    }
    \label{fig:snr_sampler}
\end{figure}

This motivated our exploration of alternative SNR samplers. To test our hypothesis, we ablate three base distributions previously explored in SD3 \citep{esser2024sd3}: \textbf{Uniform}, \textbf{Logit-Normal}, and \textbf{Mode}, and apply a \textbf{modulation function}: $m(t,\sigma) = \sigma \cdot t / (1 + (\sigma - 1) \cdot t)$ to skew them towards the noise end of the trajectory (using a shifting factor $\sigma > 1$). The resulting probability densities are visualised in Fig.~\ref{fig:snr_sampler}.

In \textbf{ablation (viii)}, our results clearly show that distributions shifted towards the noise end outperform the unshifted baselines Logit-Normal and Mode samplers by a large margin, especially in multi-view settings. This validates our hypothesis that emphasising the early, high-noise timesteps is crucial for view synthesis, which operates in a highly constrained solution space. While a uniform distribution with a shift factor of 3 performs marginally better than a shift factor of 5, indicating performance was near saturation, we chose \textbf{shifted Mode sampling} as our final design. This approach provides a superior balance between the critical early timesteps and the intermediate steps compared to a heavily shifted uniform distribution.

In \textbf{ablation (ix)}, we align our inference process with our noise-biased training strategy by adopting a \textbf{linear-quadratic sampling} schedule. This schedule samples the first half of the timesteps linearly and the second half quadratically, effectively concentrating more computation on the initial, high-noise part of the trajectory. We found that this design combination, using a noise-biased sampler for both training and inference, delivered the single most significant performance improvement across all of our Kaleido design ablations.

\subsubsection{Video Pre-training Improves 3D Efficiency}
Finally, in \textbf{ablation (x)}, we validated our core motivation of treating 3D as a specialised sub-domain of video. The results confirm that pre-training on video data significantly improves the efficiency of subsequent 3D fine-tuning. Specifically, we observed that pre-training on large-scale video data for 100K and 200K steps resulted in 1.3x and 2.0x improvements in 3D training efficiency, respectively. This demonstrates that a more capable video foundation model directly translates to faster convergence on view synthesis tasks, successfully concluding our Kaleido design ablations.

\subsection{Frame Interpolation as Zero-shot Spatial Upsampler}
While many video generation models use a VAE with temporal compression to reduce memory usage, this method is incompatible with multi-view 3D datasets, which are typically sparsely captured and lack temporal consistency. Consequently, Kaleido must rely on a standard image-based VAE. This presents a practical challenge at inference time. When Kaleido renders a dense, video-like sequence from a continuous camera trajectory, generating every frame of such a sequence solely by Kaleido alone would be computationally expensive and memory-intensive. 

To address this, we train a separate, lightweight frame interpolation model using our video data. This model's role is to efficiently generate the intermediate frames between the sparse keyframes rendered by Kaleido. For this task, we adapt the FiLM architecture \citep{reda2022film}, a deterministic, convolutional model designed for fast prediction. This two-stage approach, sparse generation by Kaleido followed by deterministic interpolation by FiLM, mitigates the high memory cost of dense rendering. It effectively emulates the decoding stage of a temporal VAE, allowing us to produce smooth, high frame-rate video sequences efficiently.

\section{Experiments}

We designed three variations of our model: Small, Medium, and Large, with increasing parameter counts to demonstrate the scalability of our architecture. The design choices for each model are summarised in Table~\ref{tab:kaleido_family}. Hereafter, we refer to our largest model simply as {\it Kaleido} and the entire collection as the {\it Kaleido family}.

\begin{table}[ht!]
    \centering
    \footnotesize
    \begin{tabular}{lccccccc}
    \toprule
    \textbf{} & \textbf{Layers} & \textbf{Hidden Size} & \textbf{Query Heads} & \textbf{KV Heads} & \textbf{Window Size} & \textbf{Aux. Encoder} & \textbf{Total Params.} \\
    \midrule
    \textbf{Kaleido-\textit{Small}} & 24 & 1024 & 16 & 4 & 4 & DINOv2-B (86M) & 653M \\
    \textbf{Kaleido-\text{Medium}} & 32 & 1280 & 20 & 5 & 4 & DINOv2-B (86M) & 1.2B \\
    \textbf{Kaleido} & 40 & 1792 & 28 & 7 & 8 & DINOv2-L (300M) & 3.1B \\
    \bottomrule
    \end{tabular}
    \caption{\textbf{Kaleido Family Architecture Details.} We detail the key hyper-parameters for our three model variants: the number of layers, hidden embedding size, number of query and key/value heads, the window size used in temporal attention, the choice of auxiliary DINOv2 encoder, and the total parameter count.}
    \label{tab:kaleido_family}
\end{table}

\subsection{Training Configurations and Evaluation Strategy}
\label{subsec:training}

\paragraph{Training Datasets} The Kaleido family is trained on a diverse mixture of object-level and scene-level datasets. For object-level data, we use \textbf{ShutterStock 3D}, our licensed collection of synthetic 3D meshes, which we render with object-centric camera poses under varied lighting conditions; and \textbf{uCO3D} \citep{liu2025uco3d}, which includes real-world objects with estimated poses. For scene-level data, we combine several datasets: \textbf{RealEstate10K} \citep{zhou2018realestate10k}, which features indoor room scenes; \textbf{DL3DV} \citep{ling2024dl3dv}, which features both indoor and outdoor scenes; and a filtered subset of \textbf{ShutterStock Video}. This licensed video subset is curated to include only static scenes, and then labelled with a pose estimator VGGT \citep{wang2025vggt}.

In summary, our 3D fine-tuning dataset consists of approximately 1.5M object sequences and 2M scene sequences. For the initial video pre-training stage, we leverage the full, \text{unfiltered Shutterstock Video} dataset, comprising 34M video clips.

\paragraph{Training Strategy} Our training process follows a two-stage, progressive-resolution curriculum. First, we pre-train Kaleido exclusively on video data at a fixed 256px resolution. We then fine-tune on our combined multi-view 3D datasets, progressively increasing the resolution from 256px to 512px, and finally to 1024px. In our 1024px fine-tuning, we introduce multi-aspect-ratio training (including 1:1, 4:5, 5:4, 16:9, and 9:16) to enable flexible resolution generation. Detailed training hyper-parameters can be found in Appendix~\ref{app:training}.

\paragraph{Evaluation Strategy} We evaluate Kaleido on both view synthesis (Section \ref{subsec:exp_nvs}), camera-conditioned video generation (Section \ref{subsec:camera}), and 3D reconstruction (Section \ref{subsec:3d_recon}). All evaluation datasets were held out and not used during model training. For all experiments, we report the zero-shot performance of Kaleido without any per-dataset fine-tuning. Unless otherwise specified, all generations use a classifier-free guidance scale of 1.5. To ensure a fair comparison with prior work, the frame interpolation model is not used in these evaluations.

\subsection{Results on Novel View Synthesis}
\label{subsec:exp_nvs}

\paragraph{Compared to Generative NVS Methods} We first evaluate Kaleido's zero-shot performance on standard novel view synthesis (NVS) benchmarks. For object-level NVS, we compare against leading object-specific generative NVS methods: \textbf{SV3D} \citep{voleti2024sv3d} and \textbf{EscherNet} \citep{kong2024eschernet} on three synthetic object-level datasets: \textbf{OmniObject3D (OO3D)} \citep{wu2023omniobject3d}, the 30-object subset of \textbf{GSO (GSO-30)} \citep{downs2022gso}, and the multi-object \textbf{RTMV} dataset \citep{tremblay2022rtmv}.

For scene-level NVS, we compare against \textbf{SEVA} \citep{zhou2025seva}, the current state-of-the-art general-purpose generative NVS model, on three scene-level datasets: \textbf{LLFF} \citep{mildenhall2019llff}, \textbf{Mip-NeRF 360} \citep{barron2022mipnerf360}, and \textbf{Tanks and Temples} \citep{knapitsch2017tandt} datasets. To ensure a fair comparison, we match our model's resolution to the baselines, using our 256px checkpoint against EscherNet (evaluated at 256px) and our 512px checkpoint against SV3D and SEVA (both evaluated at 576px). For single-view evaluations on scene-level datasets, we address scale ambiguity by sweeping camera translations along each and all axes (from $0.1$ to $2.0$) and reporting the best result, following the protocol of SEVA.

\begin{table}[b!]
\centering
\scriptsize
\setlength{\tabcolsep}{0.28em}  
\renewcommand{\arraystretch}{0.9}
\begin{tabular}{lcccccccccccccccccccccc}
\toprule
 & \textbf{OO3D} &\multicolumn{5}{c}{\textbf{GSO-30}} & \multicolumn{5}{c}{\textbf{RTMV}} & \multicolumn{2}{c}{\textbf{LLFF}} & \multicolumn{3}{c}{\textbf{Mip-NeRF 360}} & \multicolumn{4}{c}{\textbf{Tanks and Temples}} \\
   \cmidrule(lr){2-2}  \cmidrule(lr){3-7} \cmidrule(lr){8-12} \cmidrule(lr){13-14}  \cmidrule(lr){15-17} \cmidrule(l){18-21}
\# Ref. Views & 1 & 1 & 2 & 3 & 5 & 10 &  1  & 2 & 3 & 5 & 10 & 1 & 3 & 1 &3 & 6  & 1 & 3 & 6 & 9 \\
\cmidrule(r){1-1} \cmidrule(lr){2-2}  \cmidrule(lr){3-7} \cmidrule(lr){8-12} \cmidrule(lr){13-14}  \cmidrule(lr){15-17} \cmidrule(l){18-21}
Eval. Data Type & Object & \multicolumn{5}{c}{Object} & \multicolumn{5}{c}{Multi-Object} & \multicolumn{2}{c}{Scene} & \multicolumn{3}{c}{Scene} & \multicolumn{4}{c}{Scene} \\
  Eval. Resolution & 512 & \multicolumn{5}{c}{256} & \multicolumn{5}{c}{256}  & \multicolumn{2}{c}{512} & \multicolumn{3}{c}{512} & \multicolumn{4}{c}{512}\\
  Eval. Tar. Views & 20 &\multicolumn{5}{c}{15}  & \multicolumn{5}{c}{10} & \multicolumn{2}{c}{5} & \multicolumn{3}{c}{27} & \multicolumn{4}{c}{35}\\
 SoTA Model & SV3D &\multicolumn{5}{c}{EscherNet} & \multicolumn{5}{c}{EscherNet}  & \multicolumn{2}{c}{SEVA} & \multicolumn{3}{c}{SEVA} & \multicolumn{4}{c}{SEVA} \\
 Results (PSNR\textuparrow) & 19.28 & 20.24 & 22.91 & 24.09 & 25.09 & 25.90 & 10.56 & 12.66 & 13.59 & 14.52 & 15.55 &  14.03  & 19.48  & 12.93 & 15.78 & 16.70 & 11.28 & 12.65 & 13.80 & 14.72\\
\cmidrule(r){1-1} \cmidrule(lr){2-2}  \cmidrule(lr){3-7} \cmidrule(lr){8-12} \cmidrule(lr){13-14}  \cmidrule(lr){15-17} \cmidrule(l){18-21}
\textbf{Kaleido-Small}  & 19.77 & 18.58 & 23.73 & 26.20 & 29.11 & 31.66 & 13.57 & 17.18 & 18.41 & 19.97 & 21.75 & 14.57 & 19.30 & 12.75 & 15.81 & 17.07 & 11.40 & 13.13 & 14.13 & 15.20  \\
\textbf{Kaleido-Medium} & 20.78 & 20.32 & 25.78 & 28.01 & 30.74 & 32.94 & 13.78 & 18.07 & 19.41 & 21.09 & 22.73 & 14.86 & 20.40 & \textbf{14.17} & 16.47 & 17.80 & 11.36 & 13.04 & 14.43 & 15.47 \\
\textbf{Kaleido} & \textbf{21.06} & \textbf{20.94} & \textbf{26.31} & \textbf{28.89} & \textbf{31.37} & \textbf{33.74} & \textbf{14.66} & \textbf{18.48} & \textbf{19.69} & \textbf{21.13} & \textbf{23.04} & \textbf{15.34} & \textbf{20.71} & 13.74 & \textbf{16.78} & \textbf{18.03} & \textbf{11.79} & \textbf{13.20} & \textbf{14.61} & \textbf{15.88} \\
\bottomrule
\end{tabular}
\caption{\textbf{Zero-shot PSNR Performance with Generative Methods.} Kaleido achieves state-of-the-art NVS performance across all object- and scene-level benchmarks, with particularly dominant results in many-view settings. Notably, our Kaleido-Small model consistently matches or outperforms all baselines despite having significantly fewer model parameters.}
\label{tab:few_shot_nvs}
\end{table}

In Table~\ref{tab:few_shot_nvs}, our PSNR results demonstrate that even our smallest model, Kaleido-Small (0.6B), performs on par with or surpasses all baselines across both object and scene-level benchmarks. This is particularly noteworthy given its efficiency, as it uses less than half the model parameters of SEVA (1.5B) and SV3D (2.3B).

Furthermore, Kaleido exhibits strong positive scaling, with performance consistently improving as model size increases. Our largest Kaleido model decisively outperforms all competing methods across every dataset, often by a remarkable margin. The benefits of scaling are most pronounced in multi-view settings, where Kaleido achieves an incredible +7.8 dB PSNR improvement on GSO-30 (10 views) over EscherNet; and +1.3 dB PSNR improvement on LLFF (3 views) over SEVA. However, we note that both Kaleido and SEVA struggle on the Tanks and Temples dataset ($<16$ PSNR with 9 views), which features unbounded scenes with extreme viewpoint changes, highlighting a clear direction for future improvements. 

\begin{figure}[b!]
    \centering
    \small
    \setlength{\tabcolsep}{0.1em}  
    \renewcommand{\arraystretch}{0.75}
    \begin{tabular}{C{0.192\textwidth}@{\hspace{4pt}}|@{\hspace{4pt}}C{0.192\textwidth} C{0.192\textwidth} C{0.192\textwidth} C{0.192\textwidth}}
    Reference Image & \multicolumn{4}{c}{Kaleido Generations} \\
    \includegraphics[width=\linewidth]{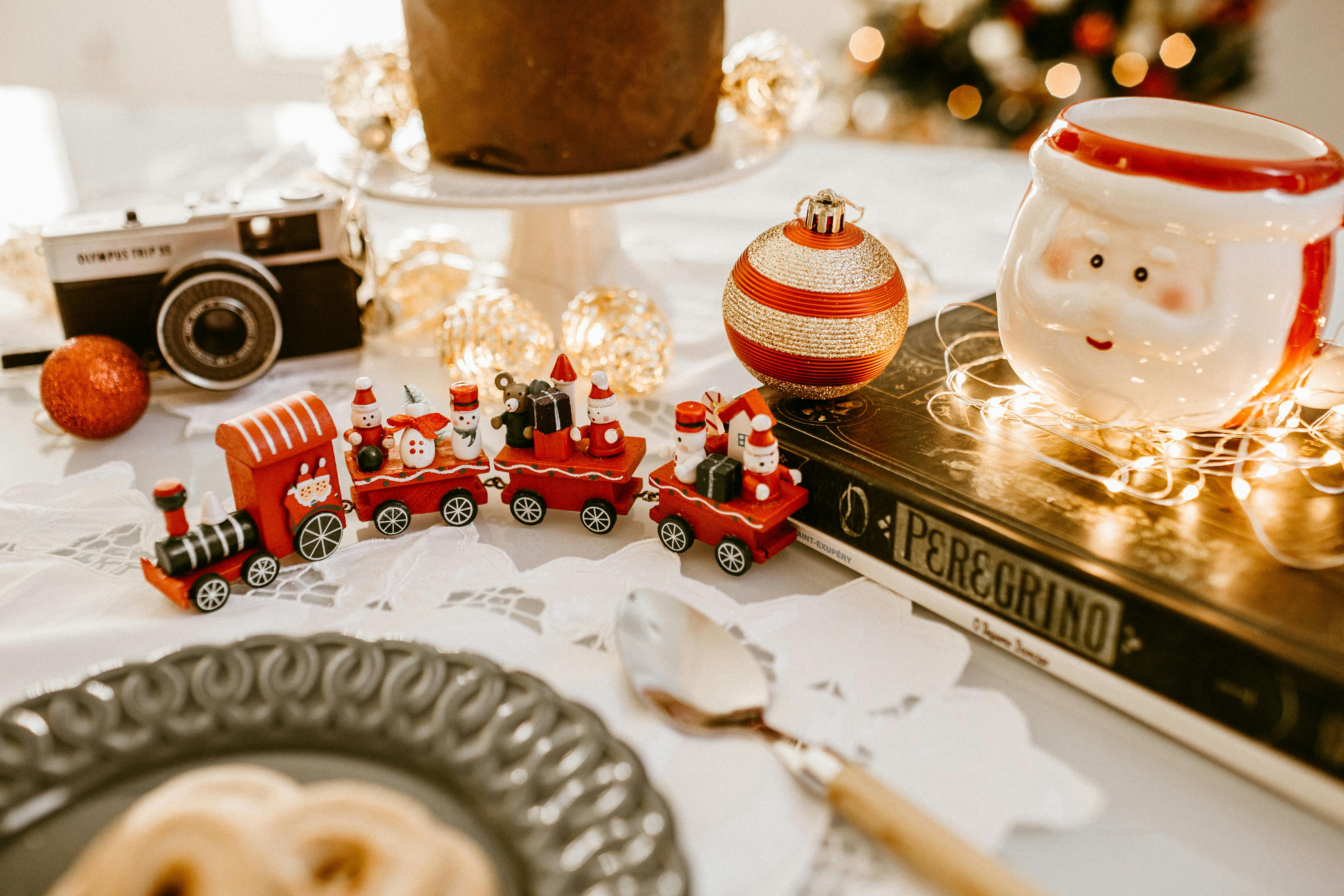} &
    \includegraphics[width=\linewidth]{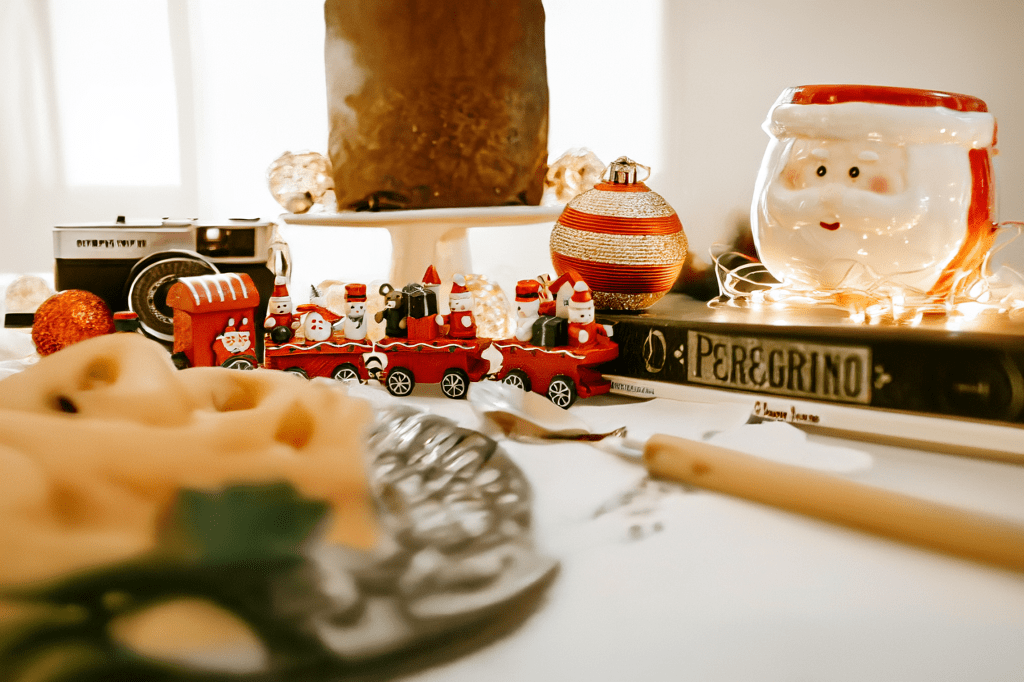} &
    \includegraphics[width=\linewidth]{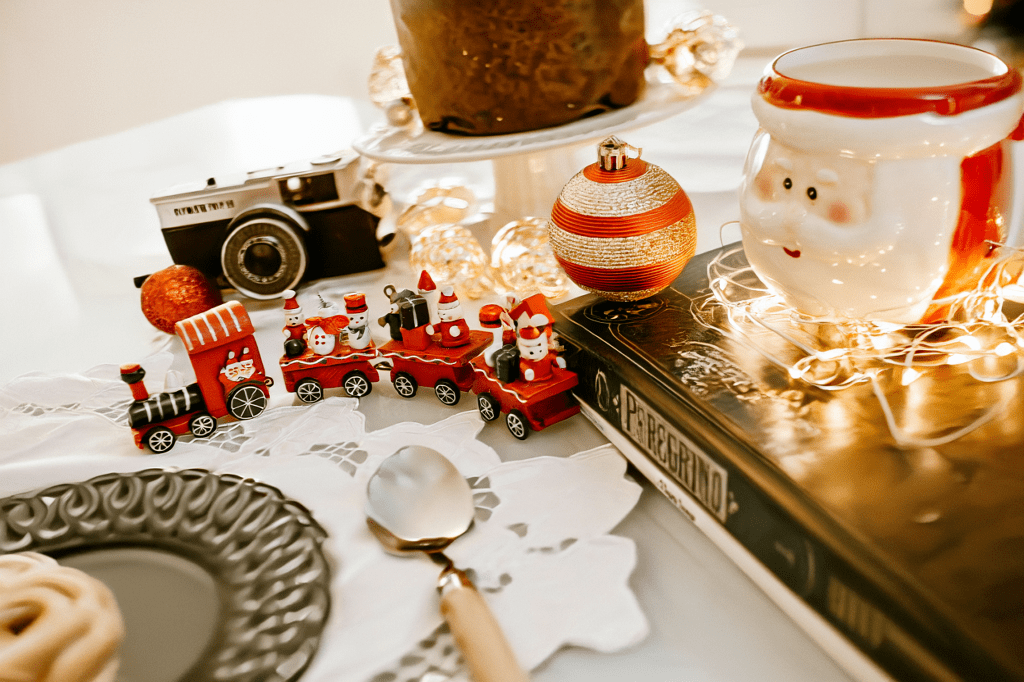} &
    \includegraphics[width=\linewidth]{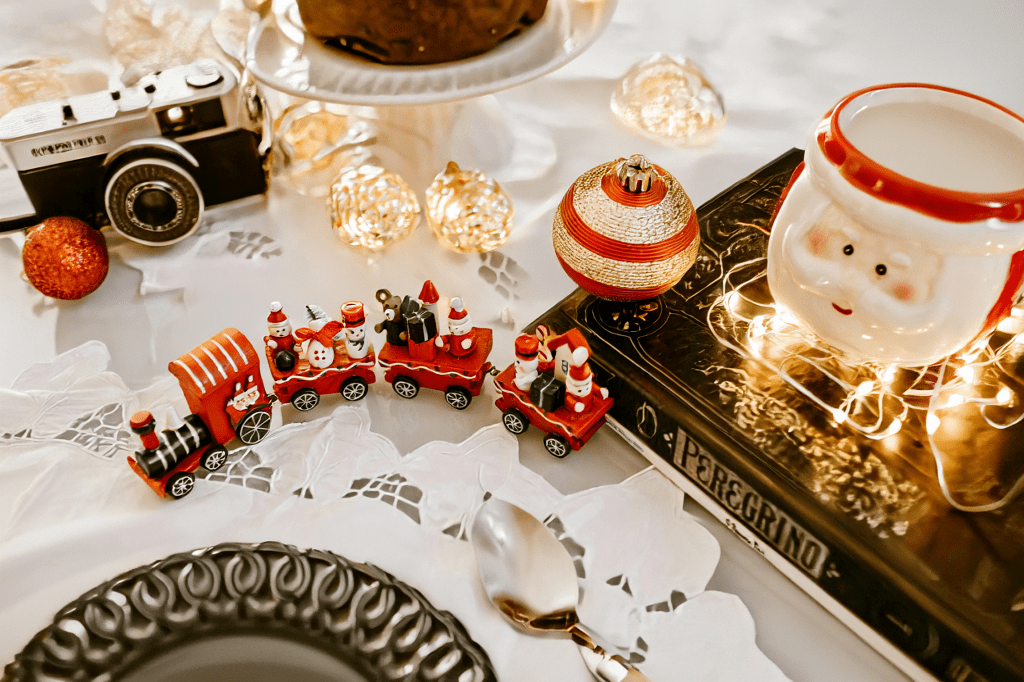} &
    \includegraphics[width=\linewidth]{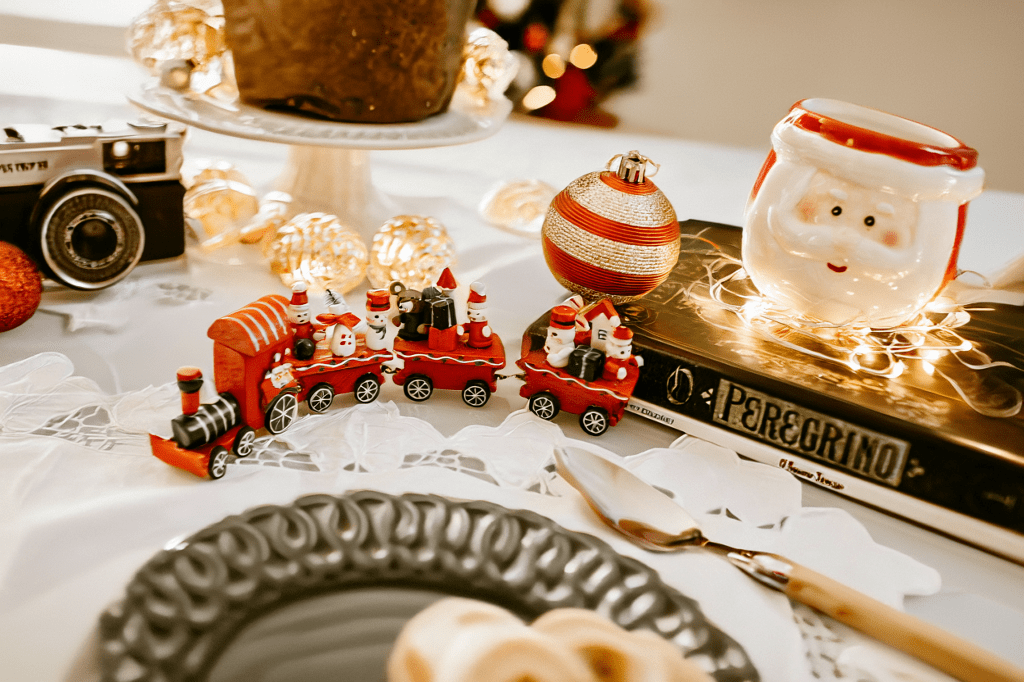} \\
    \includegraphics[width=\linewidth]{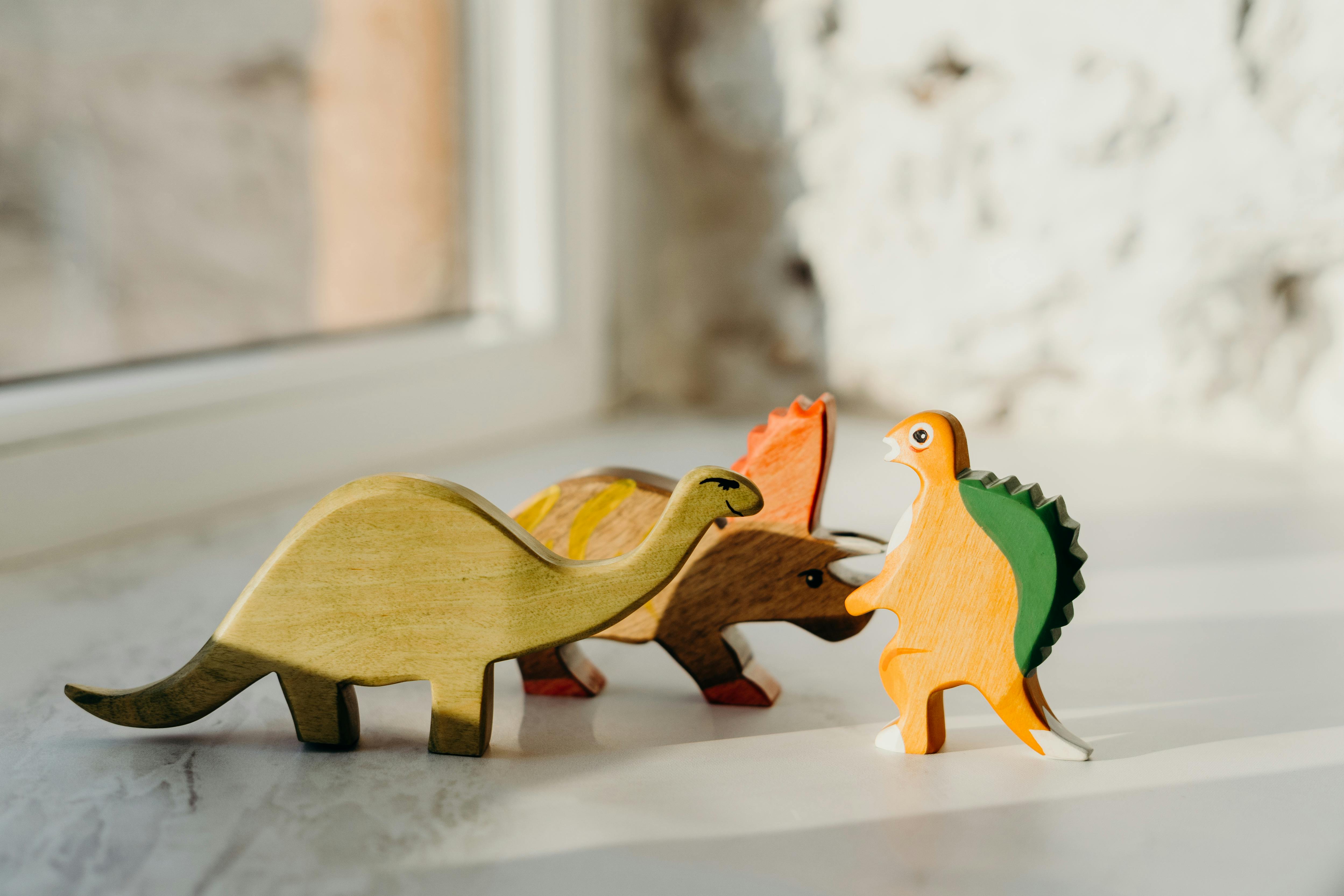} &
    \includegraphics[width=\linewidth]{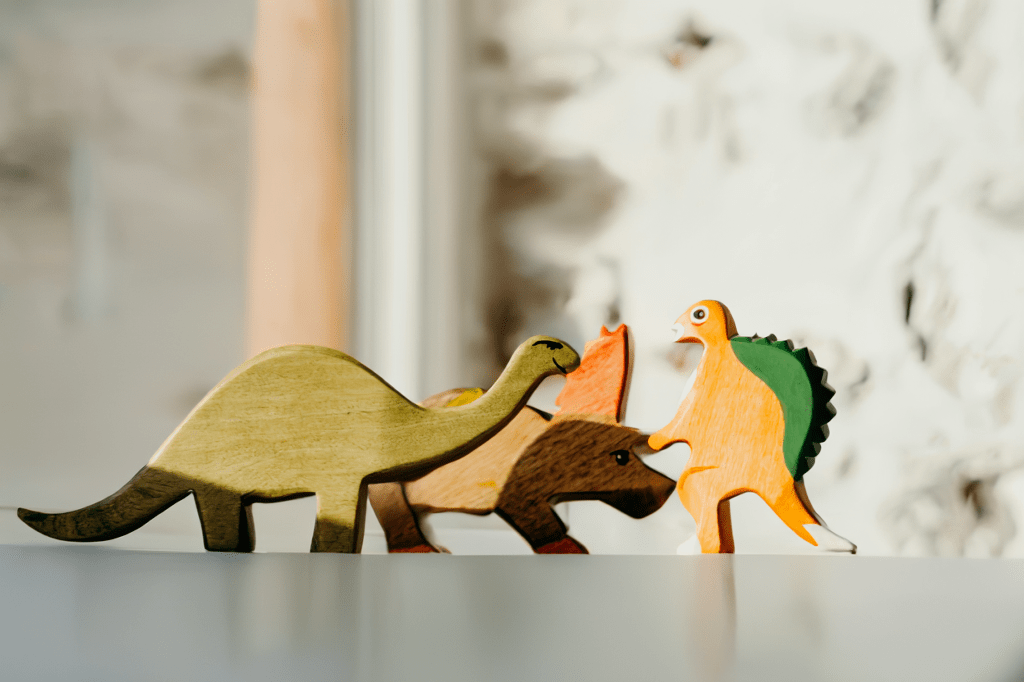} &
    \includegraphics[width=\linewidth]{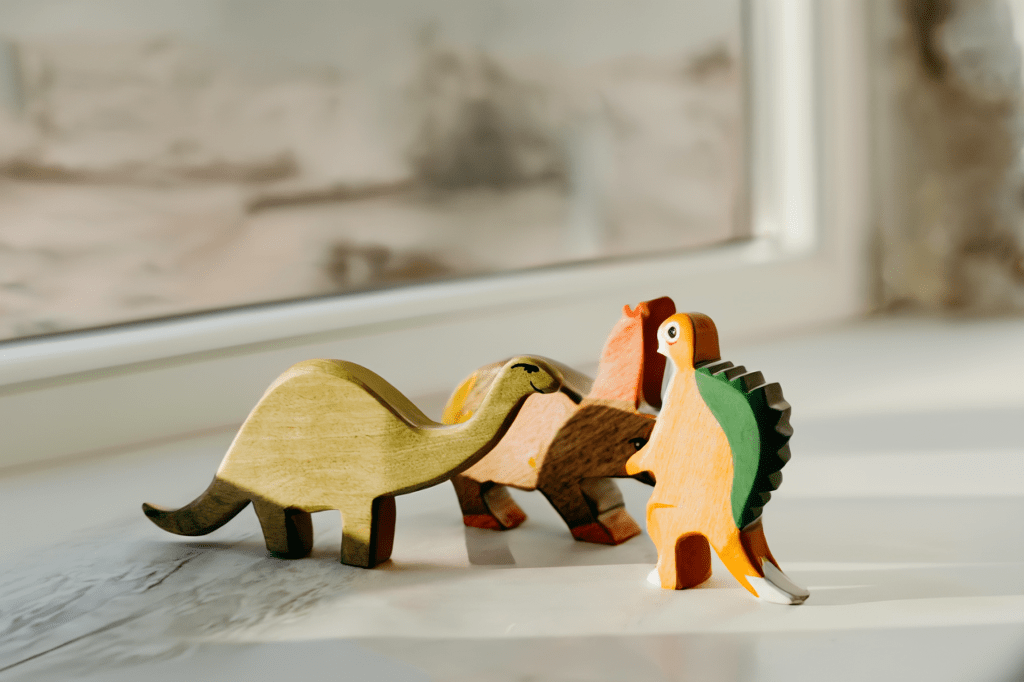} &
    \includegraphics[width=\linewidth]{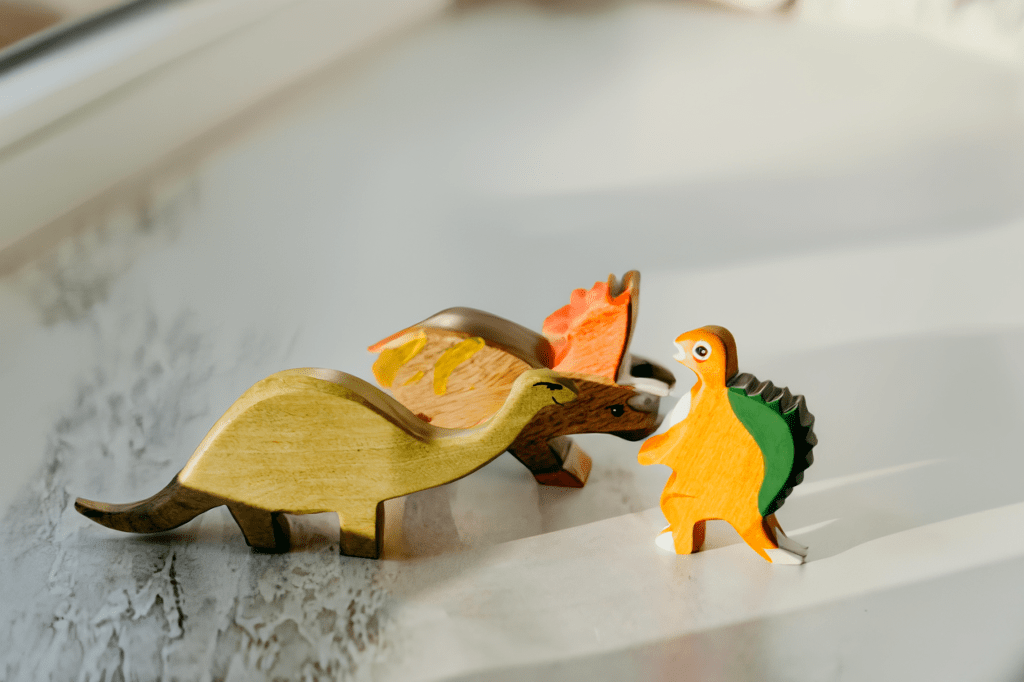} &
    \includegraphics[width=\linewidth]{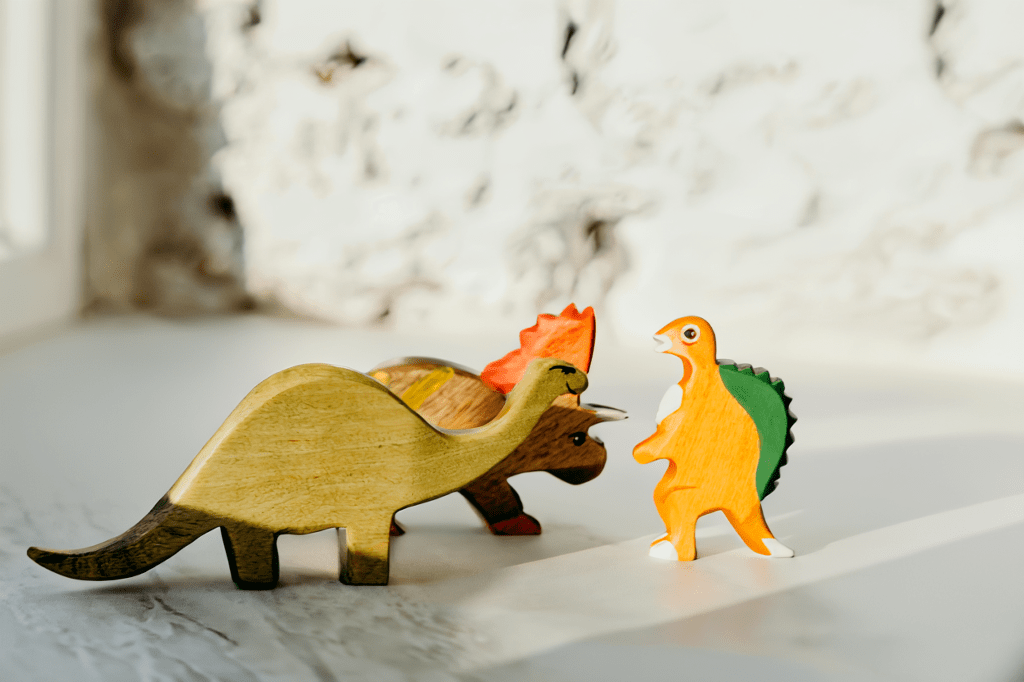} \\
    \includegraphics[width=\linewidth]{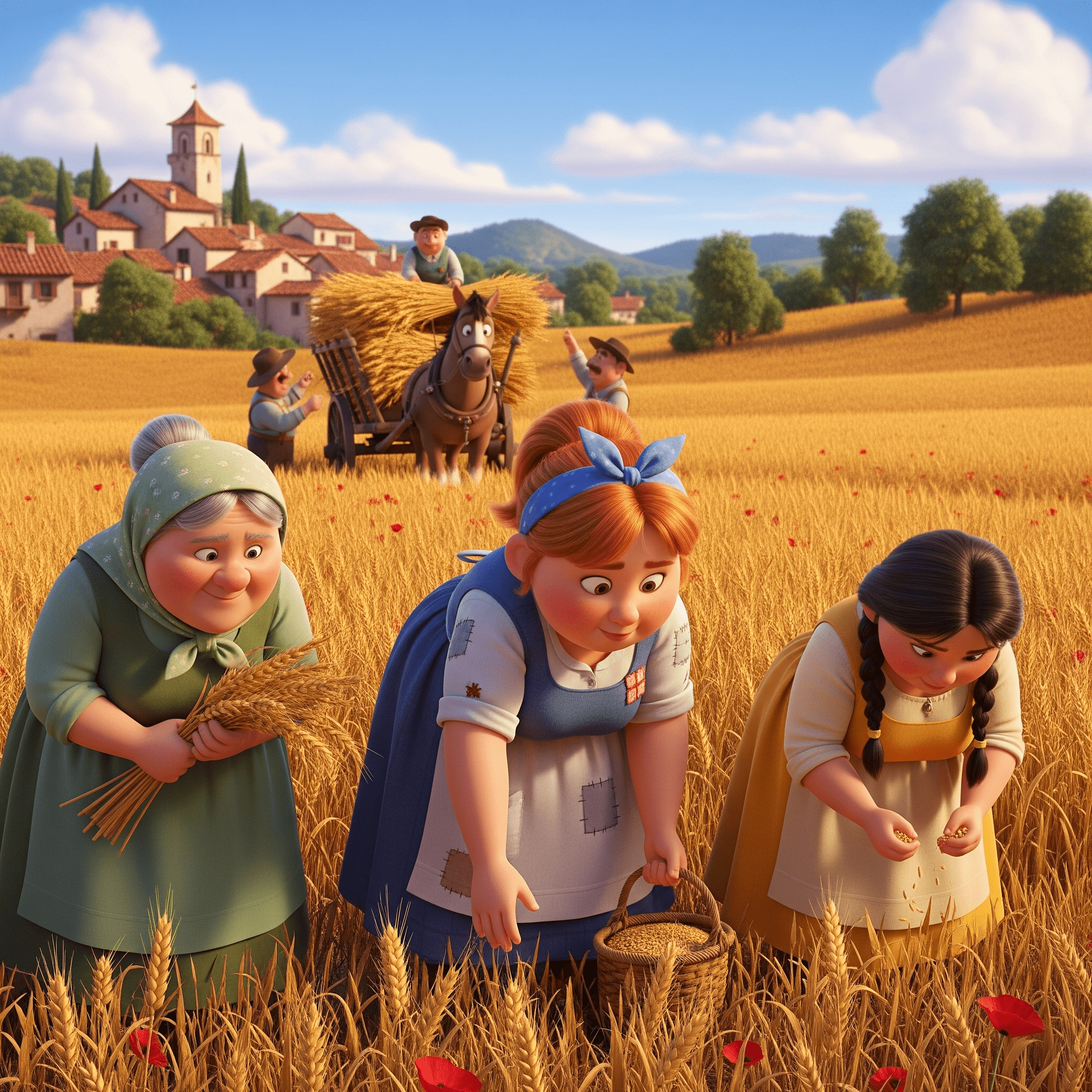} &
    \includegraphics[width=\linewidth]{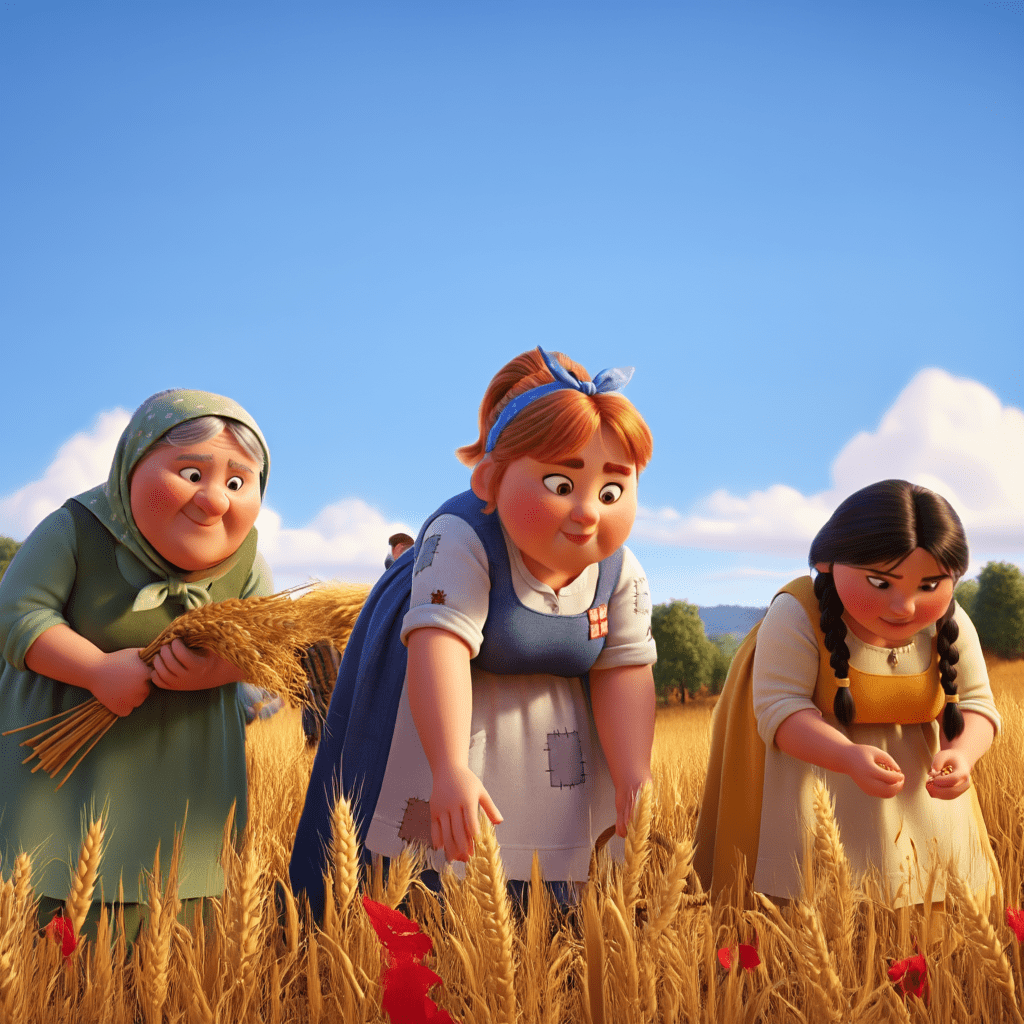} &
    \includegraphics[width=\linewidth]{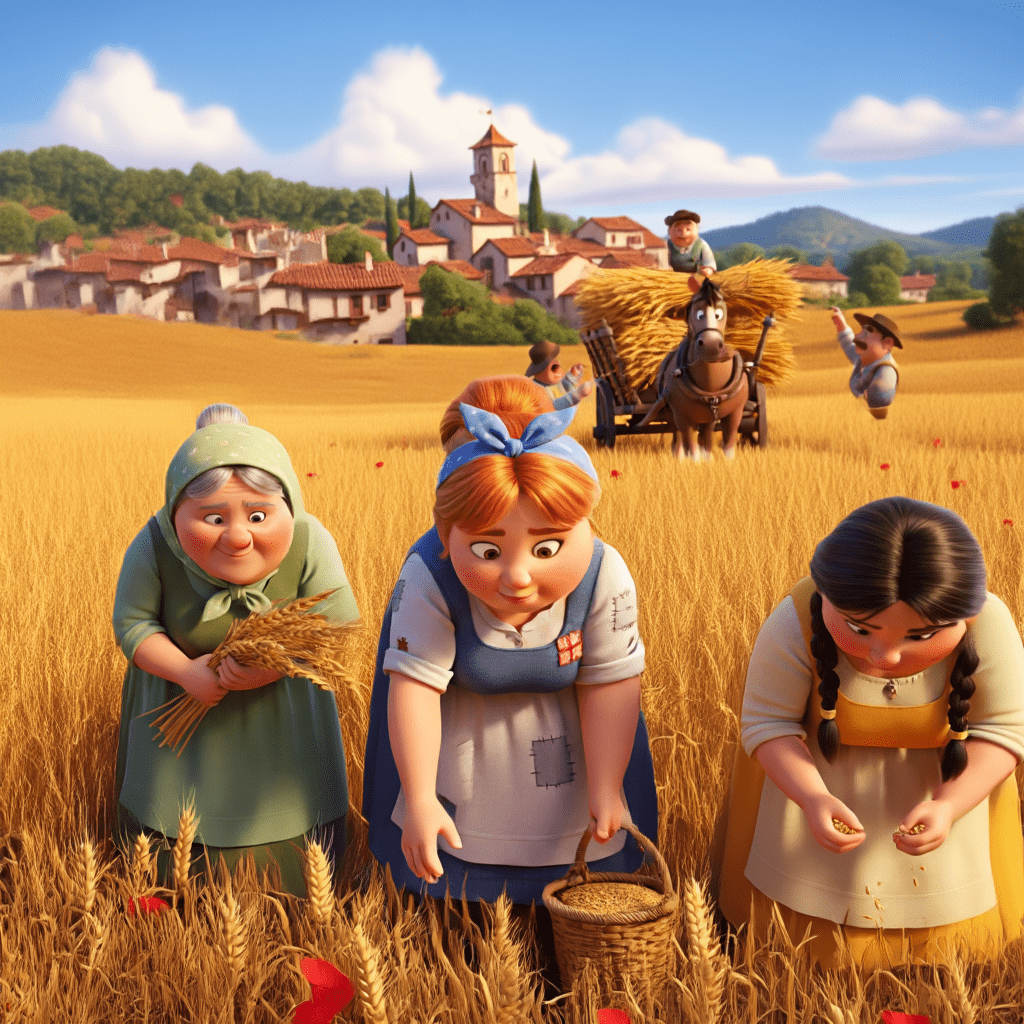} &
    \includegraphics[width=\linewidth]{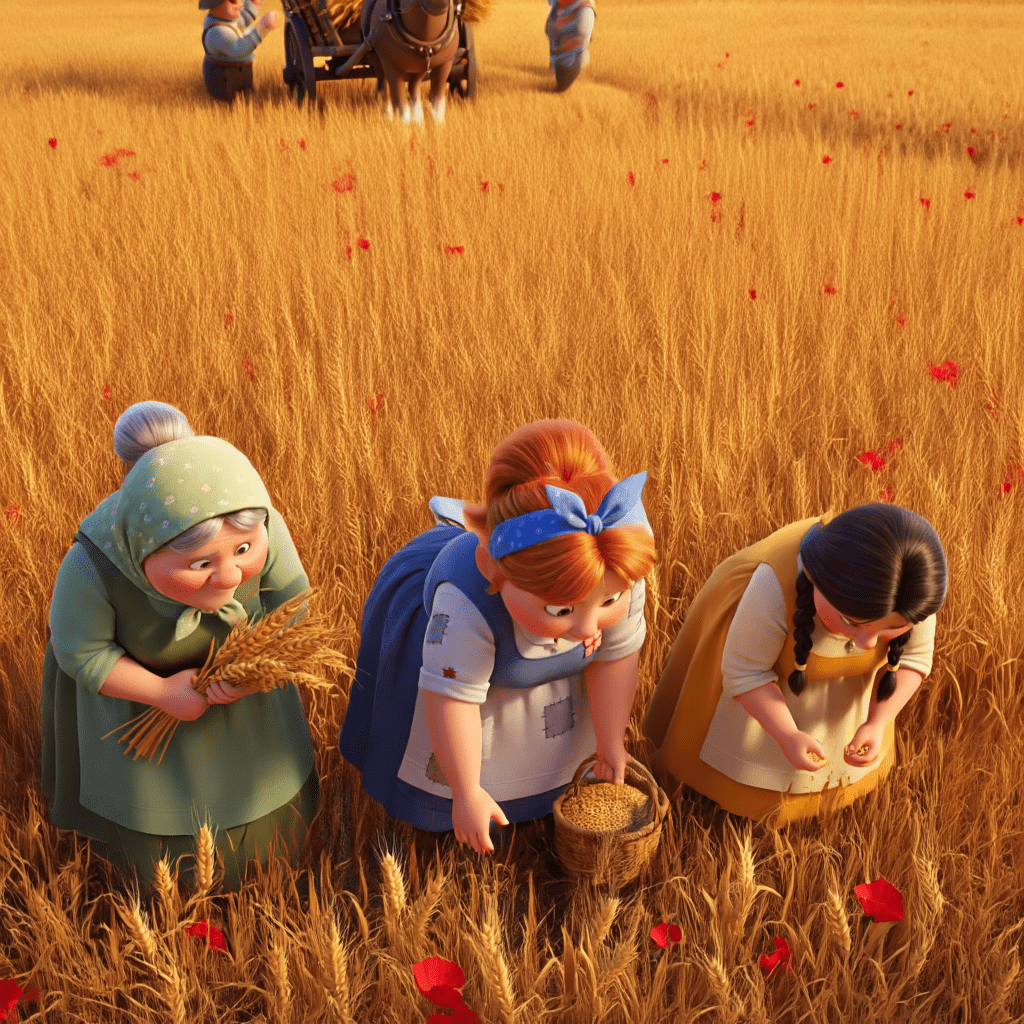} &
    \includegraphics[width=\linewidth]{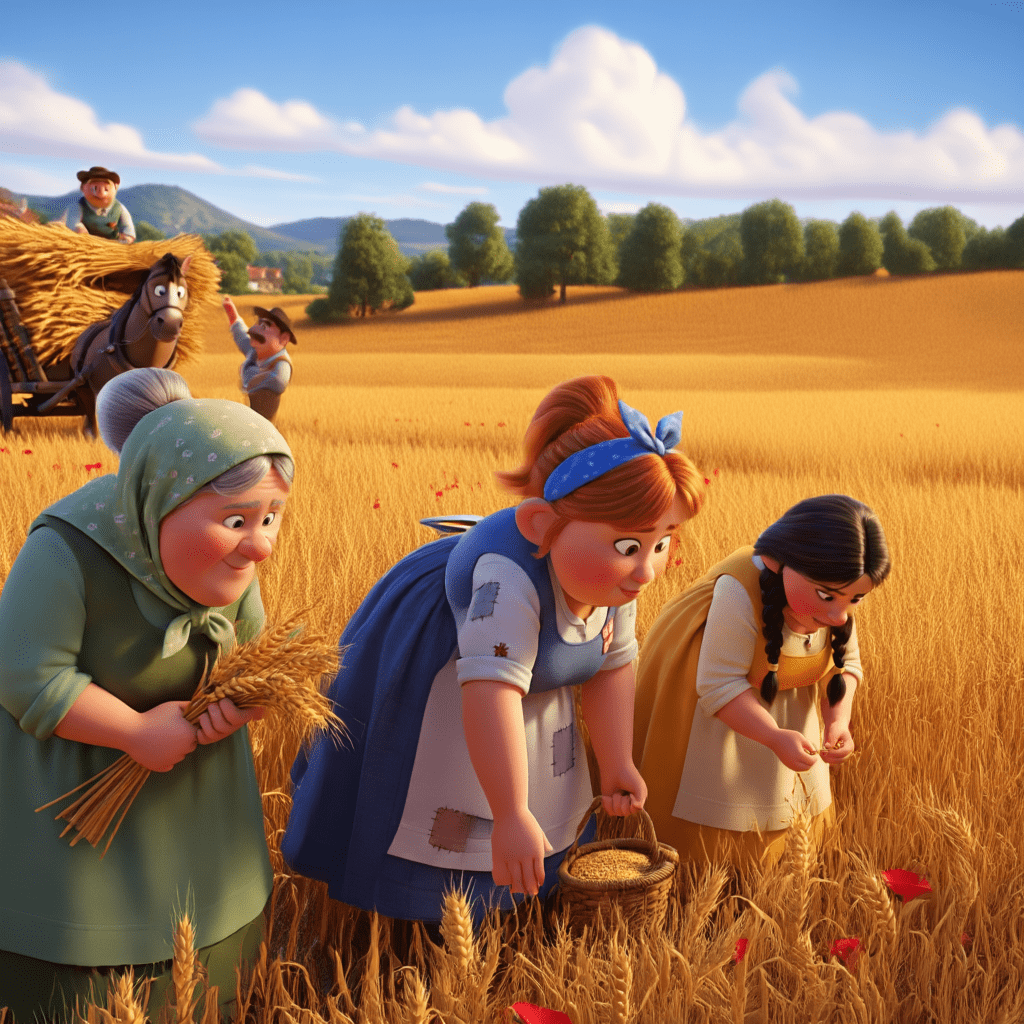} \\
    \includegraphics[width=\linewidth]{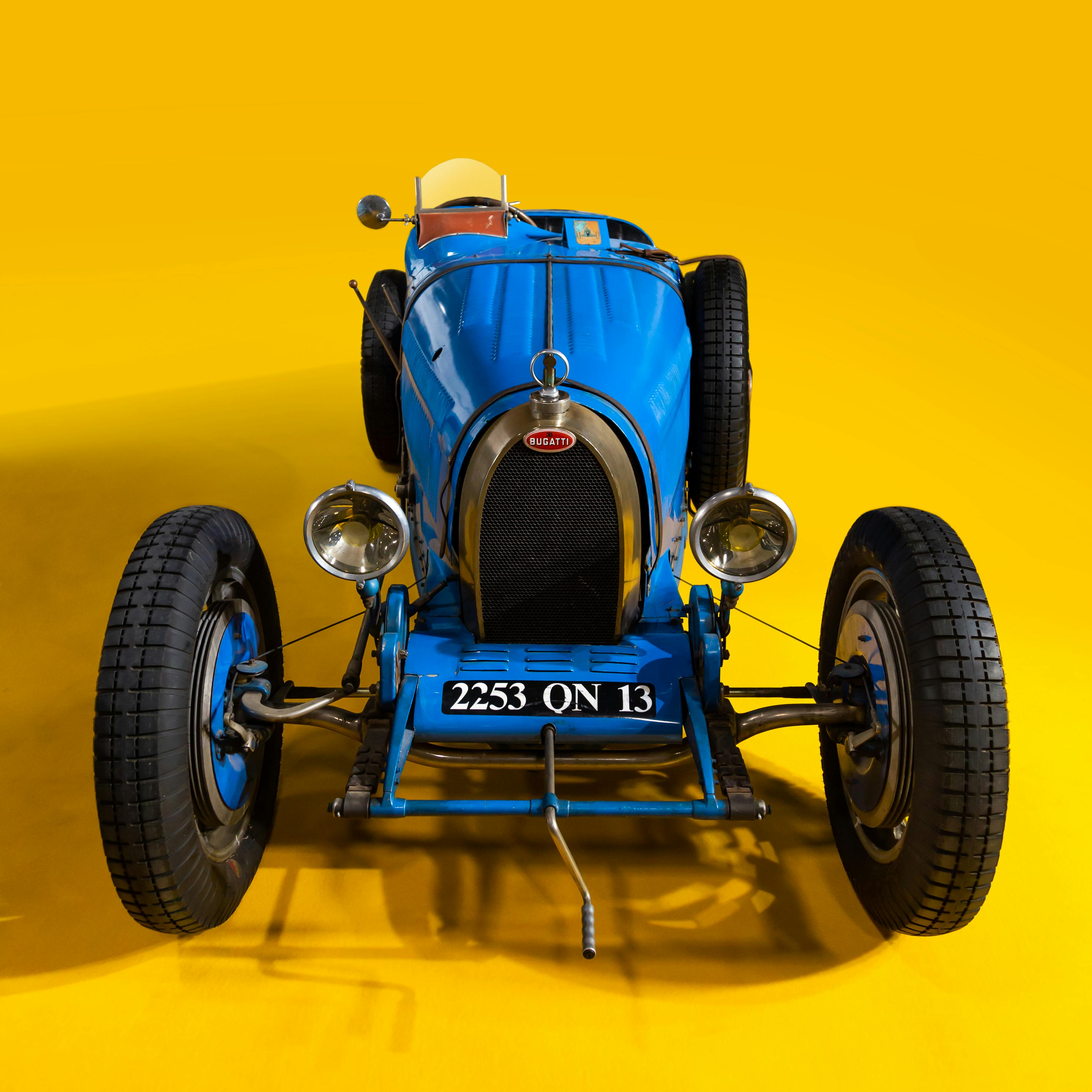} &
    \includegraphics[width=\linewidth]{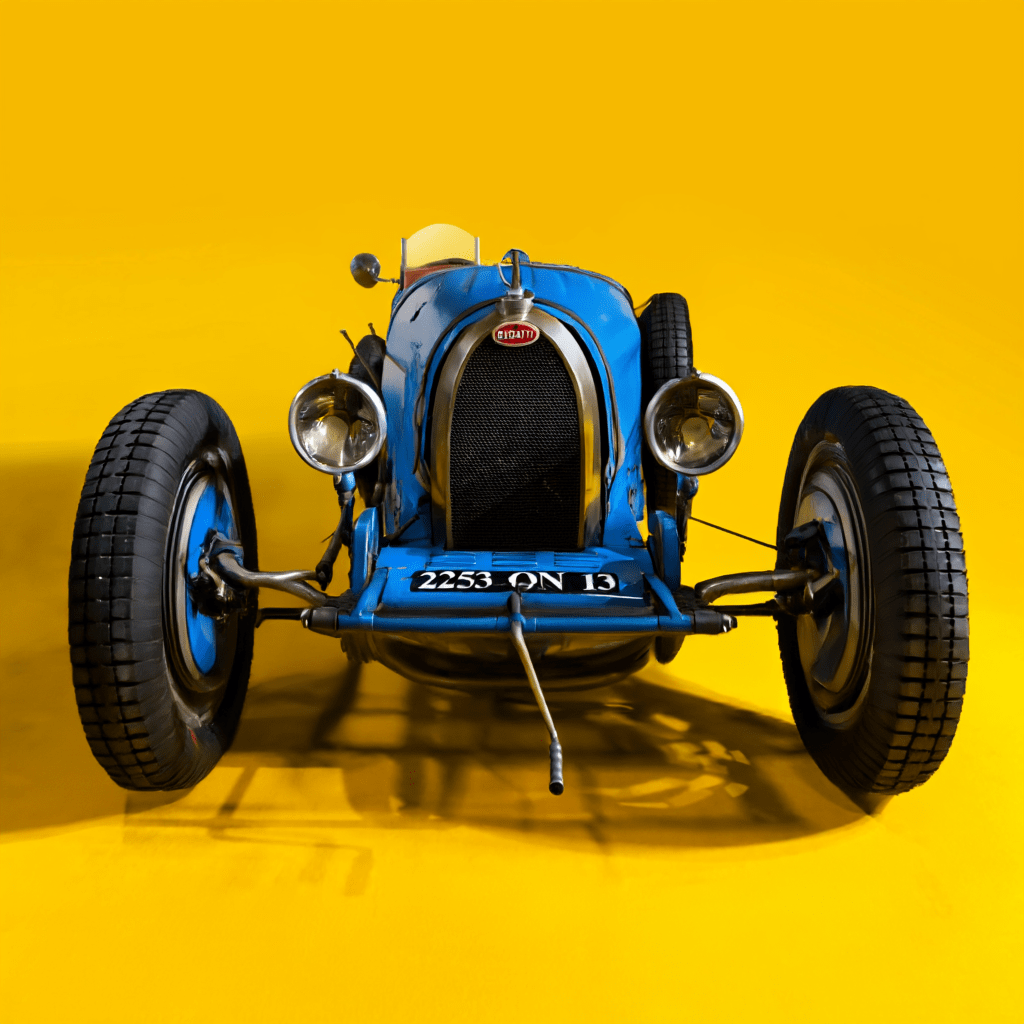} &
    \includegraphics[width=\linewidth]{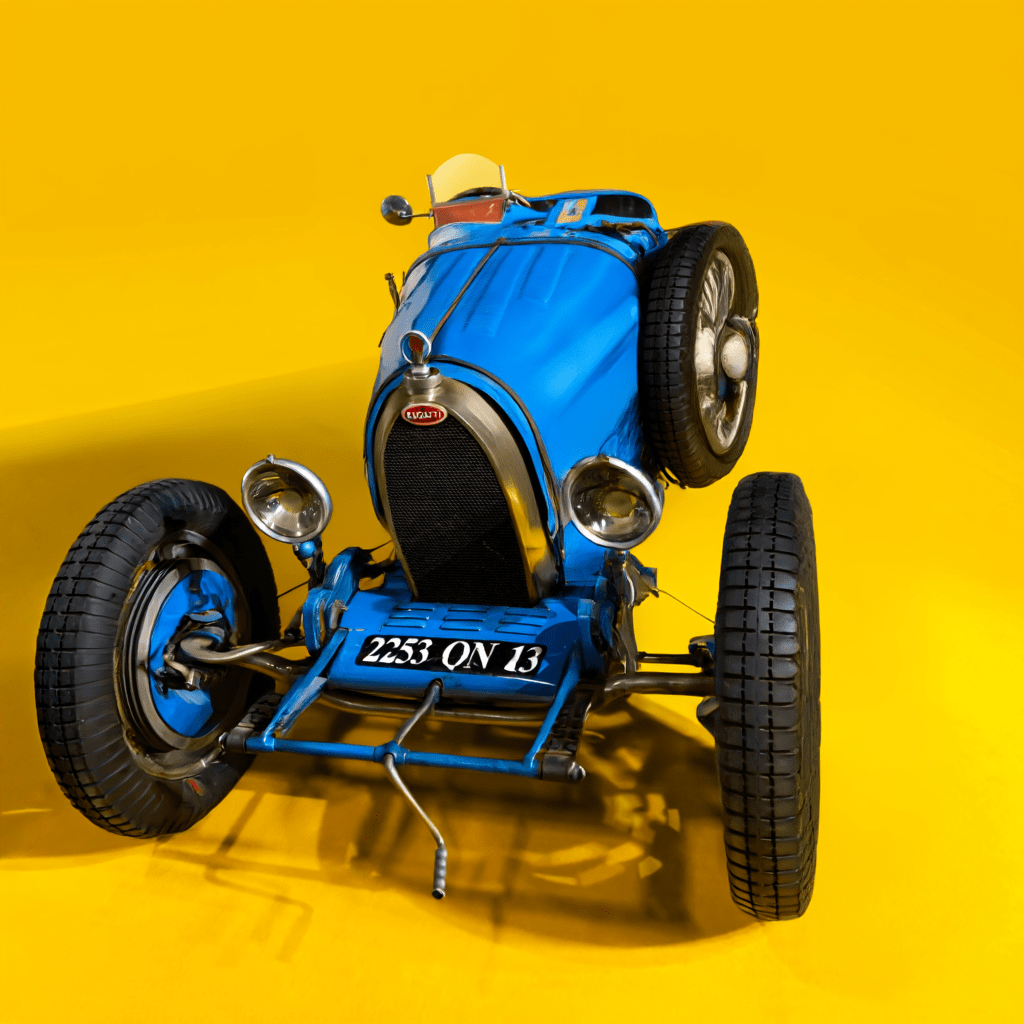} &
    \includegraphics[width=\linewidth]{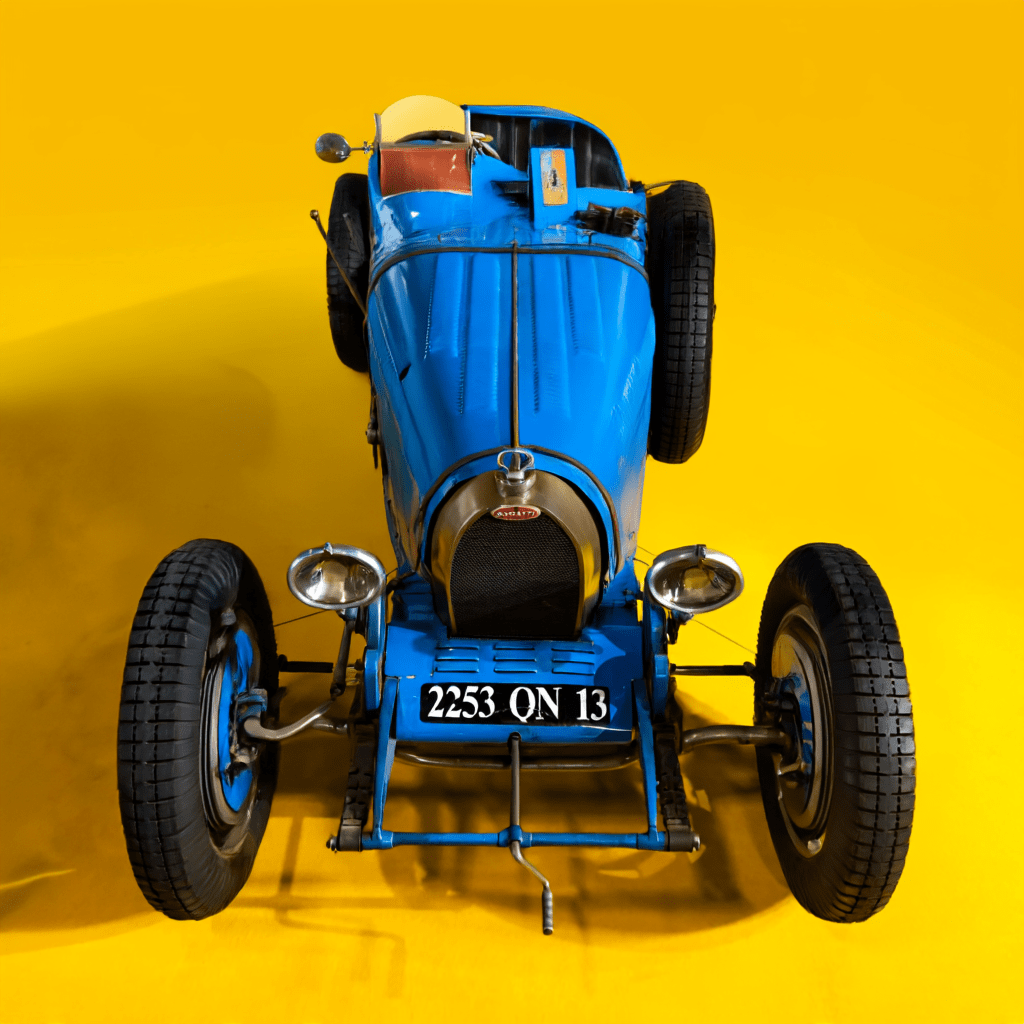} &
    \includegraphics[width=\linewidth]{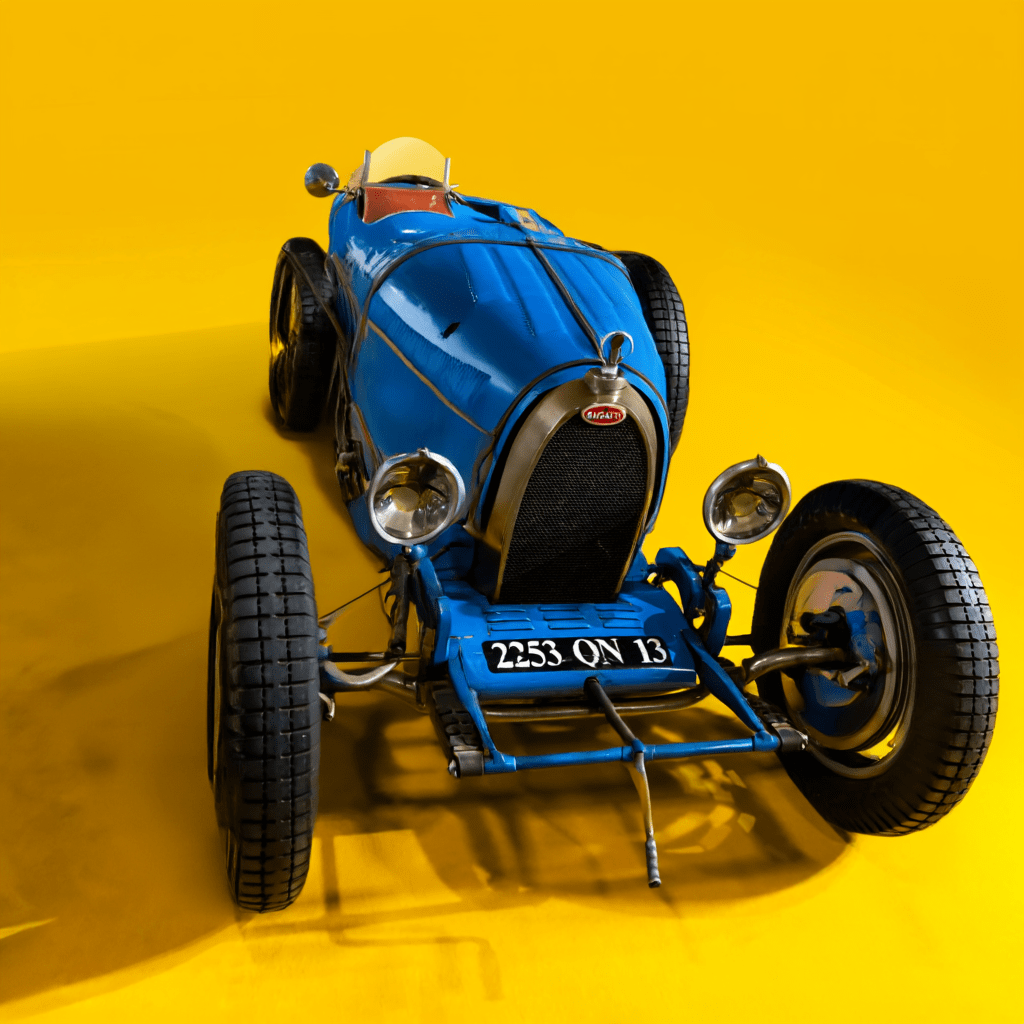} \\
    \includegraphics[width=\linewidth]{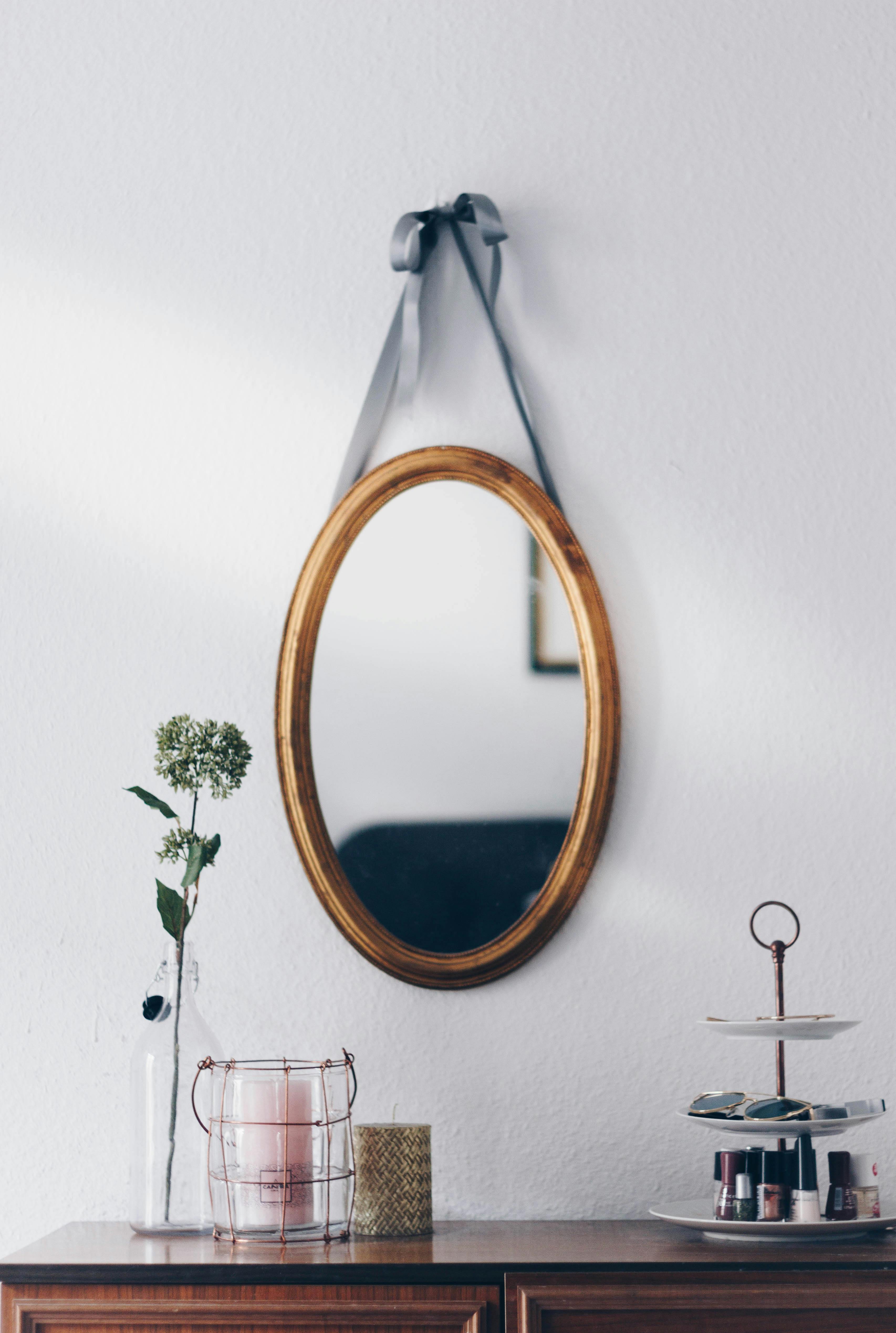} &
    \includegraphics[width=\linewidth]{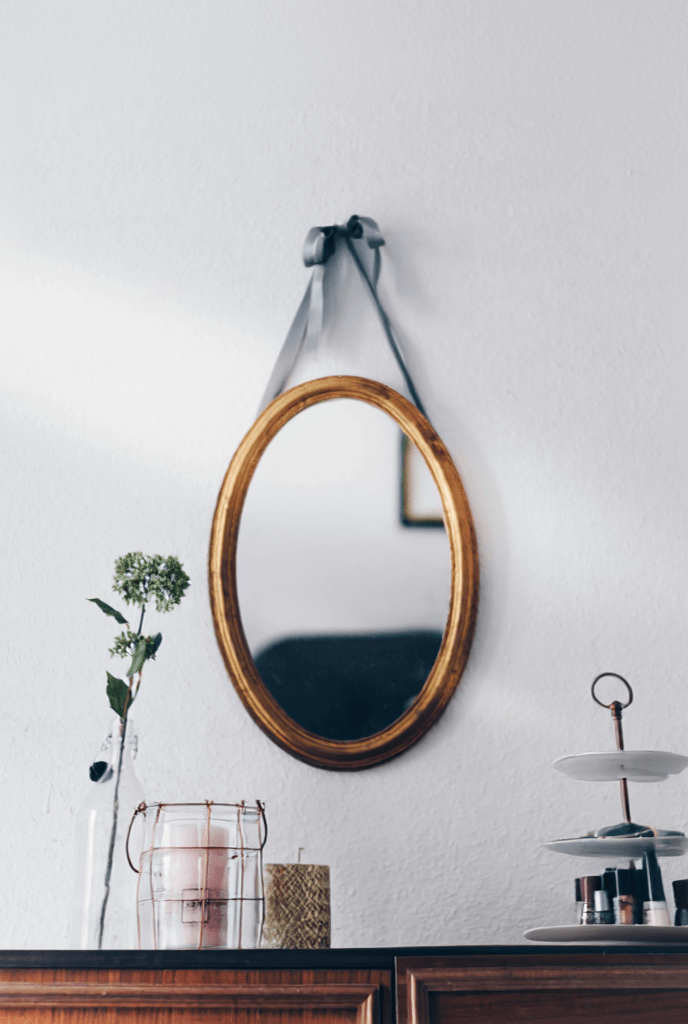} &
    \includegraphics[width=\linewidth]{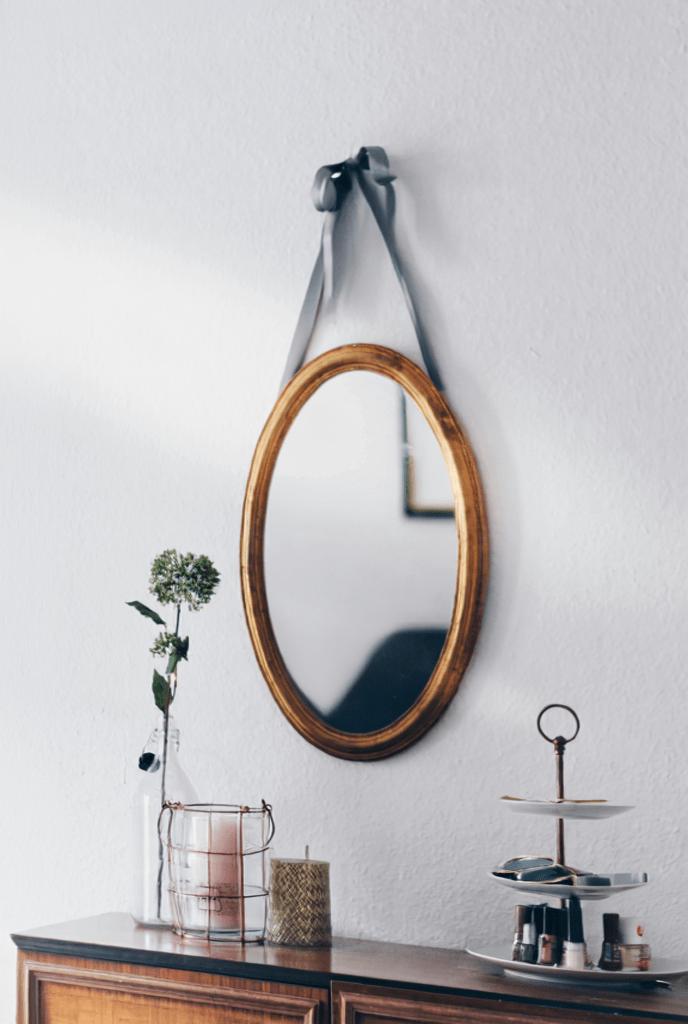} &
    \includegraphics[width=\linewidth]{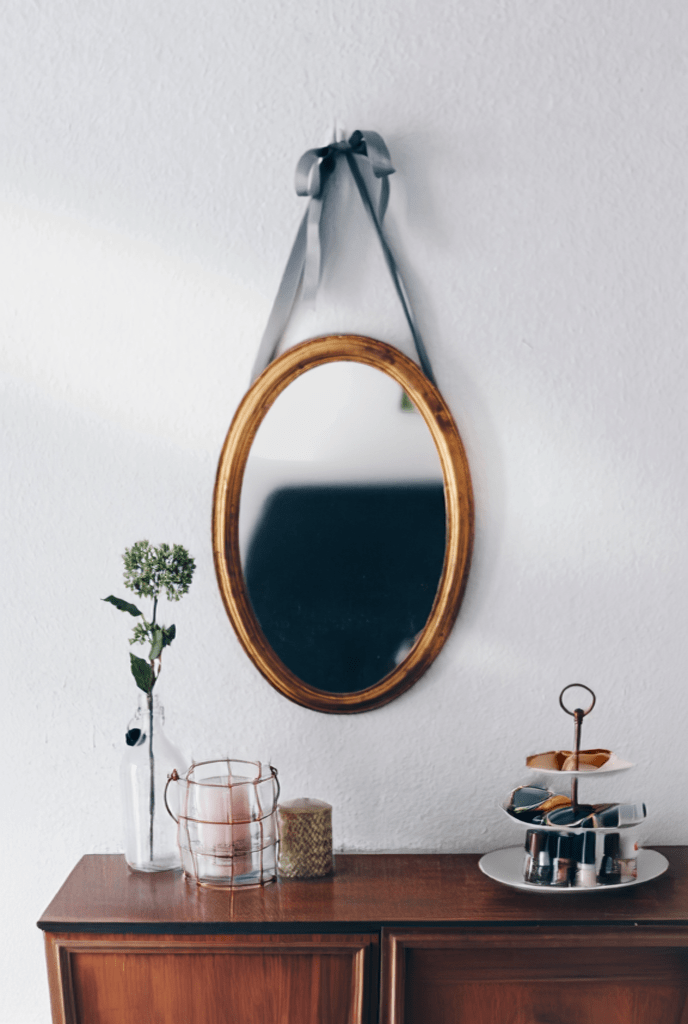} &
    \includegraphics[width=\linewidth]{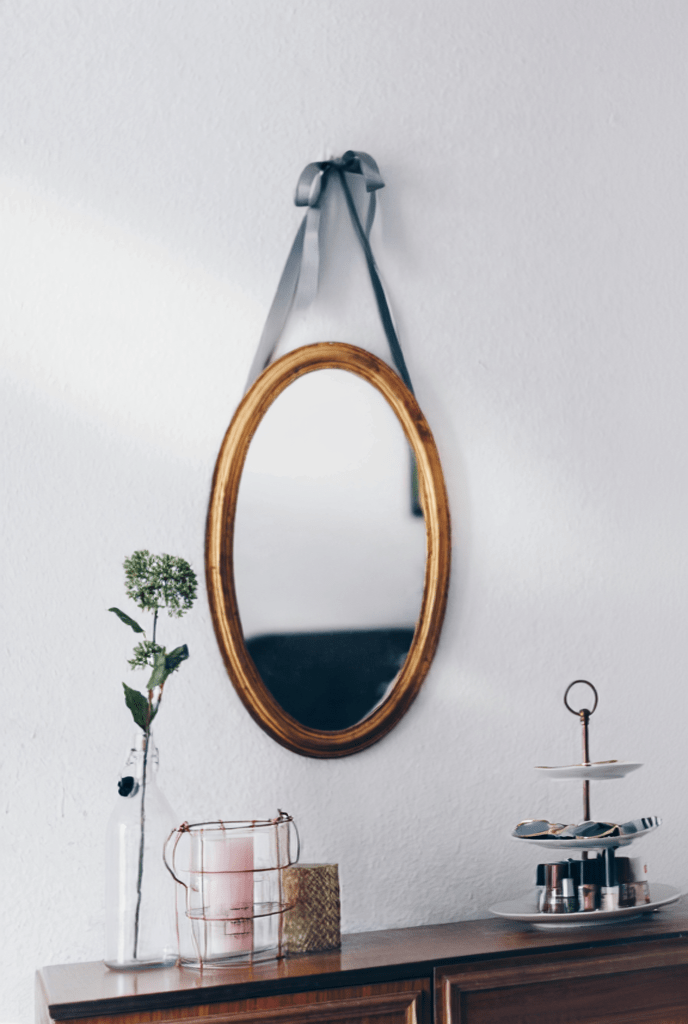} \\
    \end{tabular}
    \caption{\textbf{In-the-Wild Single-View Rendering with Kaleido.} We showcase Kaleido's zero-shot generative capabilities on challenging in-the-wild images. From a single input view (first column), Kaleido generates a sequence of photorealistic novel views along a circular, object-centric camera trajectory. The examples feature complex scenes with diverse objects and structures, demonstrating Kaleido's remarkable generalisation and high-fidelity rendering quality. }
    \label{fig:nvs_wild}
\end{figure}

A full breakdown of SSIM and LPIPS metrics is provided in Appendix \ref{app:nvs_results}. For additional qualitative results, we present high-resolution (1024px) single-view conditioned generations on in-the-wild images in Fig.~\ref{fig:nvs_wild}.

\paragraph{Compared to Per-Scene Optimisation Methods} 
Next, we evaluate the upper bound of Kaleido's rendering precision when provided with many reference views. For this analysis, we compare against two state-of-the-art scene-specific optimisation methods: \textbf{Instant-NGP} \citep{muller2022instant} and \textbf{3D Gaussian Splatting (3DGS)} \citep{kerbl20233gaussian}. As these methods are optimised per-scene, they represent a strong performance ceiling\footnote{For both Instant-NGP and 3DGS, we follow the best practices validated in NVS benchmarking pipelines \citep{kulhanek2024nerfbaselines}, applying hand-tuned pose transformations for each scene and in each dataset to obtain the optimal performance.}. We also include our generative NVS baselines that can accept a flexible number of reference views: \textbf{EscherNet}, evaluated on the \textbf{NeRF-Synthetic} dataset \citep{mildenhall2020nerf} with 256px resolution; and \textbf{SEVA}, evaluated on the \textbf{LLFF} and \textbf{Mip-NeRF 360} datasets \citep{mildenhall2019llff,barron2022mipnerf360} with 512px resolution.

\begin{figure}[ht!]
    \centering
    \begin{subfigure}[b]{0.33\linewidth}
      \centering
        \resizebox{\linewidth}{!}{{
\begin{tikzpicture}


\definecolor{kaleido}{RGB}{242,96,119}
\definecolor{eschernet}{RGB}{65,183,196}
\definecolor{baseline1}{RGB}{120,198,121}
\definecolor{baseline2}{RGB}{140,150,198}

\begin{axis}[
    every axis y label/.style={at={(current axis.north west)},above=2mm},
    width=9cm, height=5.5cm,
    legend cell align={left},
    legend style={fill opacity=0.0, draw opacity=1, text opacity=1, draw=none, legend columns=2, column sep=2pt, very thick, anchor=south east, at={(0.98,0.02)}},
    tick align=outside,
    tick pos=left,
    ymajorgrids,
    major y grid style={dashed},
    xmin=-1, xmax=106,
    ymin=7, ymax=41,
    xtick={1, 3, 5, 10, 20, 35, 60, 100},
    xticklabels={1, 2, 3, 5, 10, 20, 50, 100},
    ytick={8, 12, 16, 20, 24, 28, 32, 36},
    xlabel={\# Reference Views},
    ylabel={PSNR},
]
\addplot [ultra thick, mark=*, mark options={scale=1,solid},opacity=0.4, baseline2]
table {%
1 10.51
3 12.73
5 15.26
10 18.18
20 23.60
35 28.30
60 33.58
100 36.16
};
\addlegendentry{3DGS}
\addplot [ultra thick, mark=*, mark options={scale=1,solid},opacity=0.4, eschernet]
table {%
1 13.36
3 14.95
5 16.19
10 17.16
20 17.74
35 17.91
60 18.05
100 18.15
};
\addlegendentry{EscherNet}
\addplot [ultra thick, mark=*, mark options={scale=1,solid},opacity=0.4, baseline1]
table {%
1 10.44
3 12.20
5 14.04
10 18.79
20 23.55
35 26.42
60 28.91
100 30.61
};
\addlegendentry{Instant-NGP}
\addplot [ultra thick, mark=*, mark options={scale=1,solid}, kaleido]
table {%
1 15.30
3 18.82
5 20.22
10 21.55
20 23.20
35 24.87
60 26.48
100 27.66
};
\addlegendentry{Kaleido}
\draw[-, thick, shorten <=2pt] (axis cs:100,36.16) -- (axis cs:100,37.16) node[pos=1, anchor=south] {\footnotesize \textcolor{baseline2}{\bf 36.16}};
\draw[-, thick, shorten <=2pt] (axis cs:100,30.61) -- (axis cs:100,31.61) node[pos=1, anchor=south] {\footnotesize \textcolor{baseline1}{\bf 30.61}};
\draw[-, thick, shorten <=2pt] (axis cs:100,27.66) -- (axis cs:99,27.66) node[pos=1, anchor=south east] {\footnotesize \textcolor{kaleido}{\bf 27.66}};
\draw[-, thick, shorten <=2pt] (axis cs:100,18.15) -- (axis cs:100,19.16) node[pos=1, anchor=south] {\footnotesize \textcolor{eschernet}{\bf 18.15}};
\end{axis}

\end{tikzpicture}}}
      \caption{NeRF Synthetic [256 Res.]}
    \end{subfigure}\hfill
    \begin{subfigure}[b]{0.33\linewidth}
      \centering
        \resizebox{\linewidth}{!}{{
\begin{tikzpicture}

\definecolor{kaleido}{RGB}{242,96,119}
\definecolor{seva}{RGB}{65,183,196}
\definecolor{baseline1}{RGB}{120,198,121}
\definecolor{baseline2}{RGB}{140,150,198}

\begin{axis}[
    every axis y label/.style={at={(current axis.north west)},above=2mm},
    width=9cm, height=5.5cm,
    major y grid style={dashed},
    ymajorgrids,
    major y grid style={dashed},
    legend cell align={left},
    legend style={fill opacity=0.0, draw opacity=1, text opacity=1, draw=none, legend columns=2, column sep=2pt, very thick, anchor=south east, at={(0.98,0.02)}},
    tick align=outside,
    tick pos=left,
    xmin=0, xmax=21.2,
    ymin=9, ymax=28.2,
    xtick={1, 2, 3, 5, 10, 20},
    ytick={10, 12, 14, 16, 18, 20, 22, 24, 26},
    xlabel={\# Reference Views},
    ylabel={PSNR},
]
\addplot [ultra thick, mark=*, mark options={scale=1,solid}, opacity=0.4, baseline2]
table {%
1 11.86
2 14.72
3 16.62
5 18.43
10 22.26
20 25.31
};
\addlegendentry{3DGS}
\addplot [ultra thick, mark=*, mark options={scale=1,solid}, opacity=0.4, seva]
table {%
1 14.03
2 16.45
3 18.60
5 20.15
10 20.97
20 21.72
};
\addlegendentry{SEVA}
\addplot [ultra thick, mark=*, mark options={scale=1,solid}, opacity=0.4, baseline1]
table {%
1 13.02
2 14.50
3 15.29
5 16.44
10 19.85
20 23.82
};
\addlegendentry{Instant-NGP}
\addplot [ultra thick, mark=*, mark options={scale=1,solid}, kaleido]
table {%
1 13.96
2 19.29
3 20.13
5 21.42
10 23.16
20 24.06
};
\addlegendentry{Kaleido}
\draw[-, thick, shorten <=2pt] (axis cs:20,25.31) -- (axis cs:20,25.81) node[pos=1, anchor=south] {\footnotesize \textcolor{baseline2}{\bf 25.31}};
\draw[-, thick, shorten <=2pt] (axis cs:20,23.82) -- (axis cs:20,23.82) node[pos=1, anchor=north] {\footnotesize \textcolor{baseline1}{\bf 23.82}};
\draw[-, thick, shorten <=2pt] (axis cs:20,24.12) -- (axis cs:19.7,24) node[pos=1, anchor=south east] {\footnotesize \textcolor{kaleido}{\bf 24.06}};
\draw[-, thick, shorten <=2pt] (axis cs:20,21.72) -- (axis cs:20,21.22) node[pos=1, anchor=north] {\footnotesize \textcolor{seva}{\bf 21.72}};
\end{axis}

\end{tikzpicture}

%


}}
      \caption{LLFF [512 Res.]}
    \end{subfigure}\hfill
    \begin{subfigure}[b]{0.33\linewidth}
      \centering
        \resizebox{\linewidth}{!}{{
\begin{tikzpicture}

\definecolor{kaleido}{RGB}{242,96,119}
\definecolor{seva}{RGB}{65,183,196}
\definecolor{baseline1}{RGB}{120,198,121}
\definecolor{baseline2}{RGB}{140,150,198}

\begin{axis}[
    every axis y label/.style={at={(current axis.north west)},above=2mm},
    width=9cm, height=5.5cm,
    ymajorgrids,
    major y grid style={dashed},
    legend cell align={left},
    legend style={fill opacity=0.0, draw opacity=1, text opacity=1, draw=none, legend columns=2, column sep=2pt, very thick, anchor=south east, at={(0.98,0.02)}},
    tick align=outside,
    tick pos=left,
    xmin=-1, xmax=106,
    ymin=3, ymax=31,
    xtick={1, 3, 5, 10, 20, 35, 60, 100},
    xticklabels={1, 2, 3, 5, 10, 20, 50, 100},
    ytick={4, 8, 12, 16, 20, 24, 28},
    xlabel={\# Reference Views},
    ylabel={PSNR},
]
\addplot [ultra thick, mark=*, mark options={scale=1,solid}, opacity=0.4, baseline2]
table {%
1 7.68
3 9.26
5 10.30
10 11.51
20 15.02
35 18.89
60 24.09
100 27.16
};
\addlegendentry{3DGS}
\addplot [ultra thick, mark=*, mark options={scale=1,solid},  opacity=0.4, seva]
table {%
1 12.93
3 14.65
5 15.25
10 16.62
20 17.47
35 18.89
60 19.70
100 19.79
};
\addlegendentry{SEVA}
\addplot [ultra thick, mark=*, mark options={scale=1,solid}, opacity=0.4, baseline1]
table {%
1 8.89
3 9.78
5 9.78
10 10.44
20 10.57
35 12.07
60 15.62
100 21.15
};
\addlegendentry{Instant-NGP}
\addplot [ultra thick, mark=*, mark options={scale=1,solid}, kaleido]
table {%
1 12.86
3 14.51
5 15.36
10 16.59
20 18.64
35 19.95
60 21.48
100 23.06
};
\addlegendentry{Kaleido}
\draw[-, thick, shorten <=2pt] (axis cs:100,27.16) -- (axis cs:100,27.86) node[pos=1, anchor=south] {\footnotesize \textcolor{baseline2}{\bf 27.16}};
\draw[-, thick, shorten <=2pt] (axis cs:100,19.79) -- (axis cs:100,19.09) node[pos=1, anchor=north] {\footnotesize \textcolor{seva}{\bf 19.79}};
\draw[-, thick, shorten <=2pt] (axis cs:100,21.15) -- (axis cs:98,21.15) node[pos=1, anchor=east] {\footnotesize \textcolor{baseline1}{\bf 21.15}};
\draw[-, thick, shorten <=2pt] (axis cs:100,23.06) -- (axis cs:100,23.59) node[pos=1, anchor=south] {\footnotesize \textcolor{kaleido}{\bf 23.06}};
\end{axis}

\end{tikzpicture}

}}
      \caption{Mip-NeRF 360 [512 Res.]}
    \end{subfigure}\hfill
    \caption{\textbf{PSNR Performance with Per-Scene Optimisation Methods.} Kaleido's performance scales consistently with more reference views, demonstrating strong zero-shot generalisation despite being trained on 12 fixed total frames. It significantly outperforms other generative NVS baselines, with the performance gap widening as more views are provided. Notably, when given all available reference views, Kaleido surpasses Instant-NGP on both scene-level datasets. }
    \label{fig:many_shot_nvs}
\end{figure}
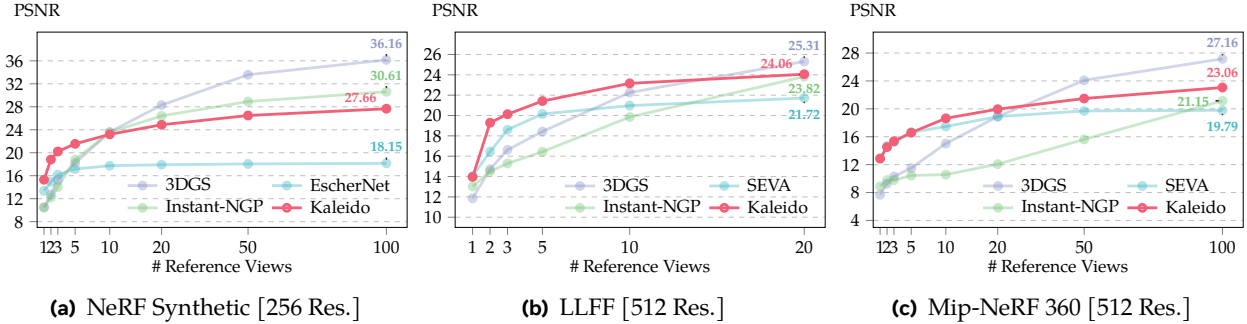

In Fig.~\ref{fig:many_shot_nvs}, we can observe that Kaleido and the other generative baselines (EscherNet, SEVA) initially outperform the scene-specific methods when given fewer than 10 reference views. However, a key difference emerges as more views are added: the performance of the other generative baselines quickly plateaus, while Kaleido shares the same positive scaling trend as the per-scene optimisation methods, with its performance continuing to improve. This superior zero-shot view generalisation, despite being trained on only 12 fixed total views, highlights the advantages of Kaleido's design and creates a widening performance gap over other generative models.

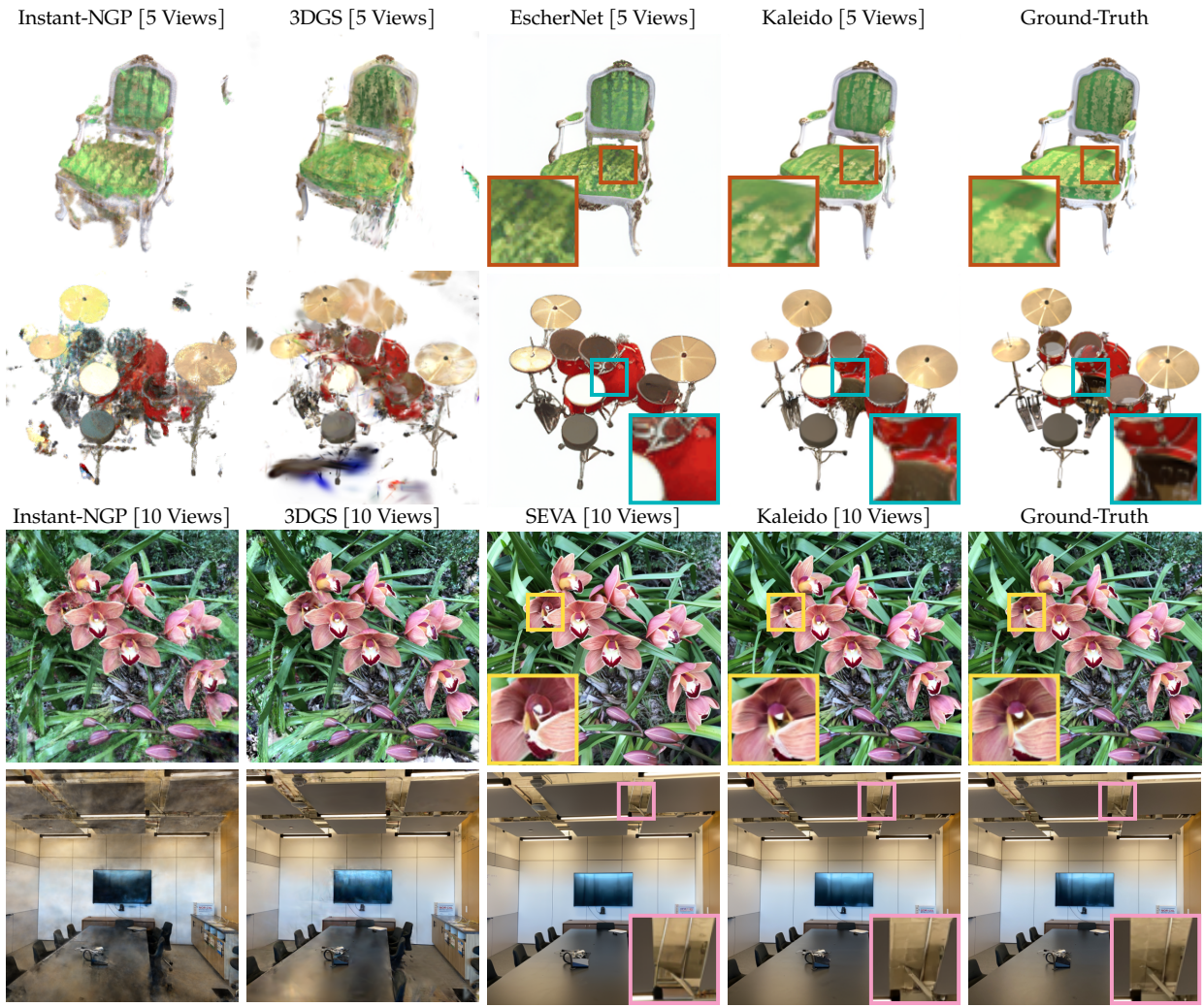
\begin{figure}[t!]
    \newcommand\cleanspy[1]{%
    \begin{tikzpicture}[]
       \centering
        \vspace{-0.2cm}
        \node[inner sep=0pt, outer sep=0pt] {\includegraphics[width=\linewidth]{#1}};
       \end{tikzpicture}
     }
    \newcommand\makespy[1]{%
    \begin{tikzpicture}[spy using outlines={rectangle, magnification=2.5, height=1.2cm, width=1.2cm, every spy on node/.append style={line width=1.5}}]
      \centering
      \vspace{-0.2cm}
        \node[inner sep=0pt, outer sep=0pt] {\includegraphics[width=\linewidth]{#1}};
        \spy[color=Bittersweet] on (0.2, -0.2) in node[line width=1.5,anchor=north east] at (-0.37,-0.37);
       \end{tikzpicture}
     }
   \newcommand\makespyb[1]{%
    \begin{tikzpicture}[spy using outlines={rectangle, magnification=2.5, height=1.2cm, width=1.2cm, every spy on node/.append style={line width=1.5}}]
      \centering
      \vspace{-0.2cm}
        \node[inner sep=0pt, outer sep=0pt] {\includegraphics[width=\linewidth]{#1}};
        \spy[color=Aquamarine] on (0.08, 0.18) in node[line width=1.5,anchor=north west] at (0.37,-0.34);
       \end{tikzpicture}
     }
    \newcommand\makespyc[1]{%
    \begin{tikzpicture}[spy using outlines={rectangle, magnification=2.5, height=1.2cm, width=1.2cm, every spy on node/.append style={line width=1.5}}]
      \centering
      \vspace{-0.2cm}
        \node[inner sep=0pt, outer sep=0pt] {\includegraphics[width=\linewidth]{#1}};
        \spy[color=Goldenrod] on (-0.8, 0.5) in node[line width=1.5,anchor=north east] at (-0.37,-0.37);
       \end{tikzpicture}
     }
    \newcommand\makespyd[1]{%
    \begin{tikzpicture}[spy using outlines={rectangle, magnification=2.5, height=1.2cm, width=1.2cm, every spy on node/.append style={line width=1.5}}]
      \centering
      \vspace{-0.2cm}
        \node[inner sep=0pt, outer sep=0pt] {\includegraphics[width=\linewidth]{#1}};
        \spy[color=Lavender] on (0.45, 1.2) in node[line width=1.5,anchor=north west] at (0.37,-0.37);
       \end{tikzpicture}
     }
    \centering
    \footnotesize
    \setlength{\tabcolsep}{0.2em}  
    \renewcommand{\arraystretch}{0.2}
    \begin{tabular}{C{0.192\textwidth} C{0.192\textwidth} C{0.192\textwidth} C{0.192\textwidth} C{0.192\textwidth}}
    Instant-NGP [5 Views]& 3DGS [5 Views] & EscherNet [5 Views] & Kaleido [5 Views] & Ground-Truth \\
    \cleanspy{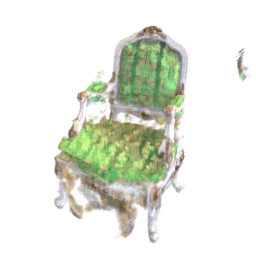} &
    \cleanspy{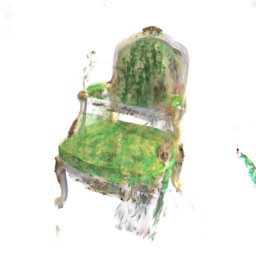} &
    \makespy{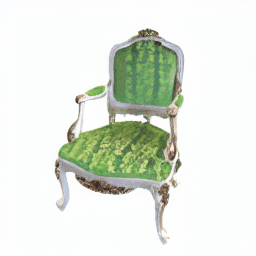} &
    \makespy{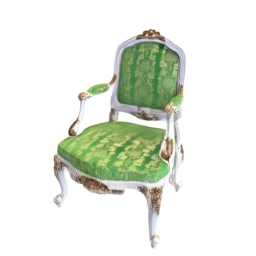} &
    \makespy{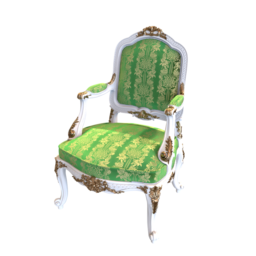} \\  
    \cleanspy{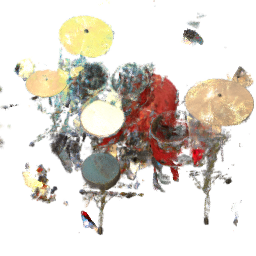} &
    \cleanspy{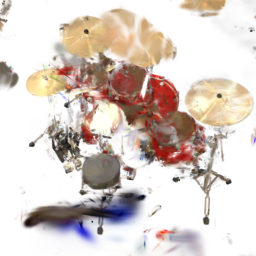} &
    \makespyb{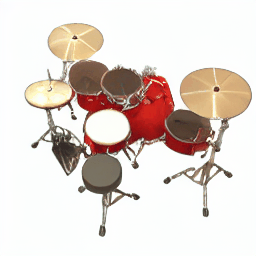} &
    \makespyb{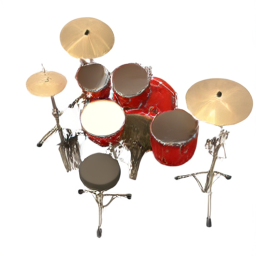} &
    \makespyb{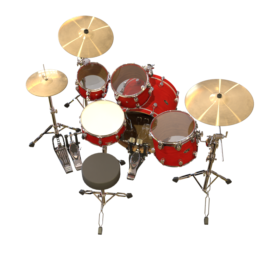} \\ 
    Instant-NGP [10 Views]& 3DGS [10 Views] & SEVA [10 Views] & Kaleido [10 Views] & Ground-Truth \\
    \cleanspy{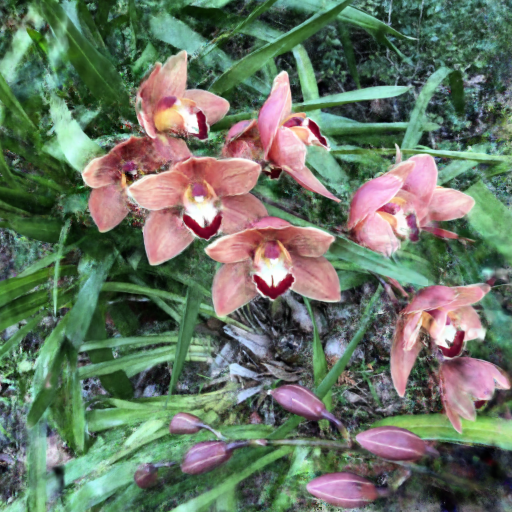} &
    \cleanspy{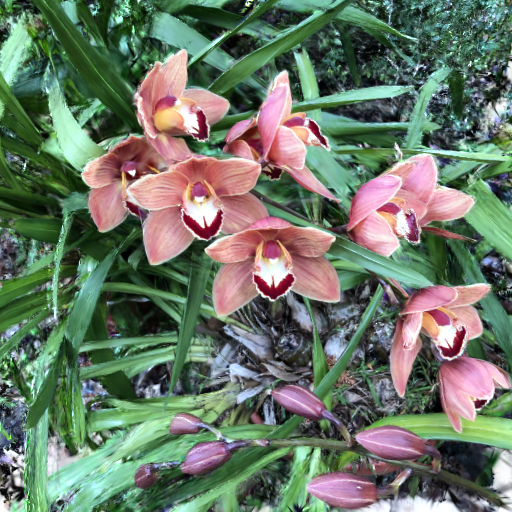} &
    \makespyc{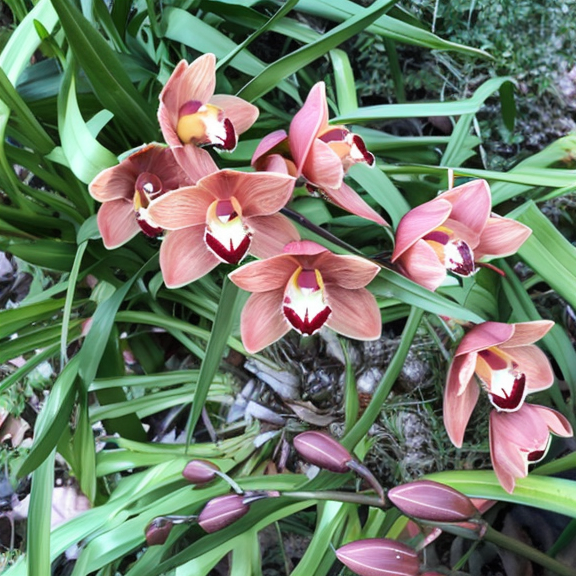} &
    \makespyc{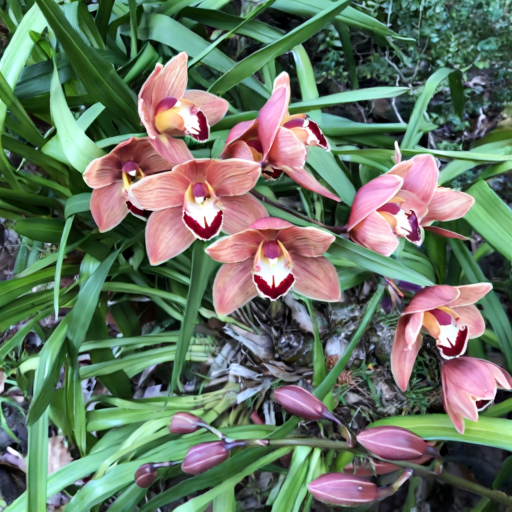} &
    \makespyc{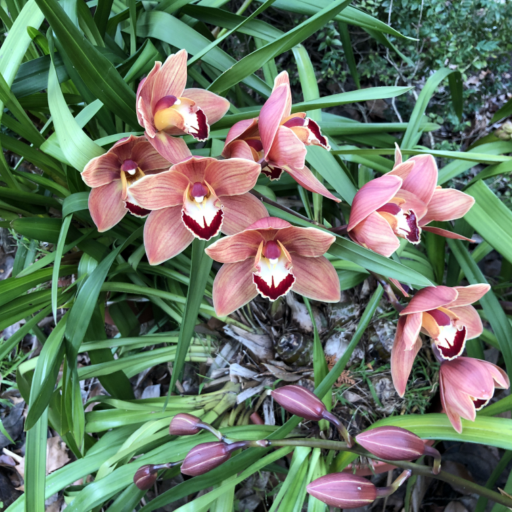} \\  
    \cleanspy{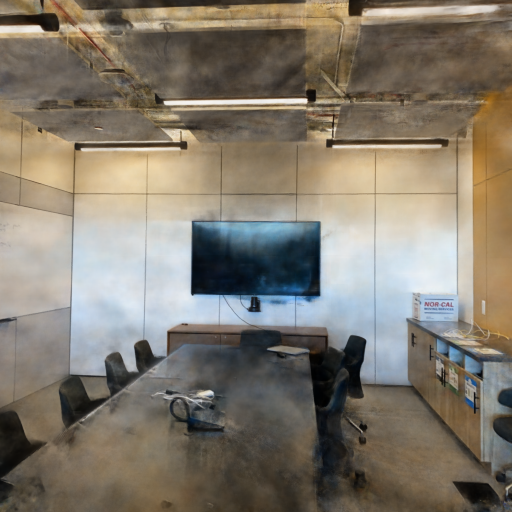} &
    \cleanspy{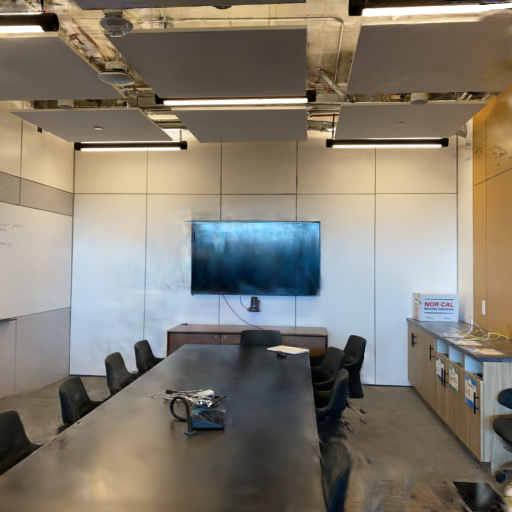} &
    \makespyd{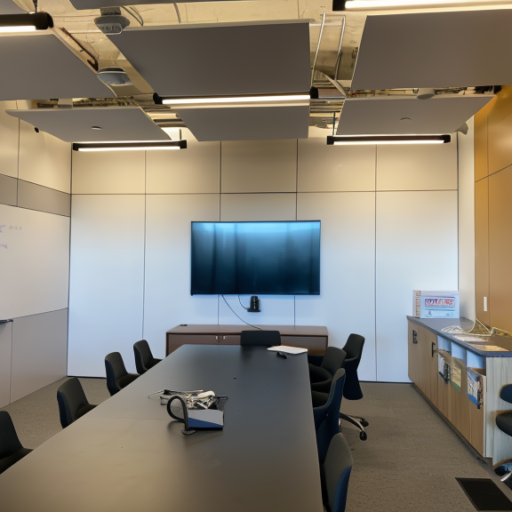} &
    \makespyd{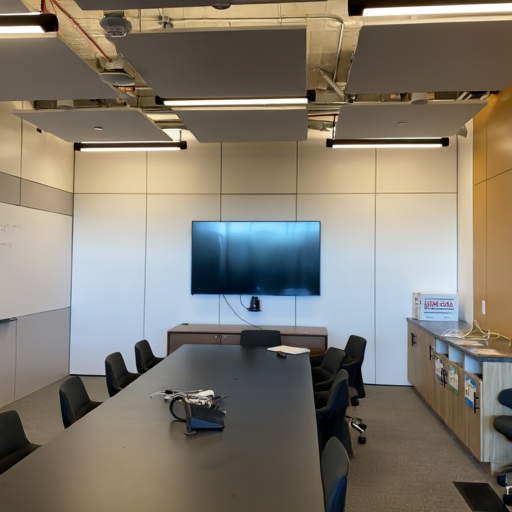} &
    \makespyd{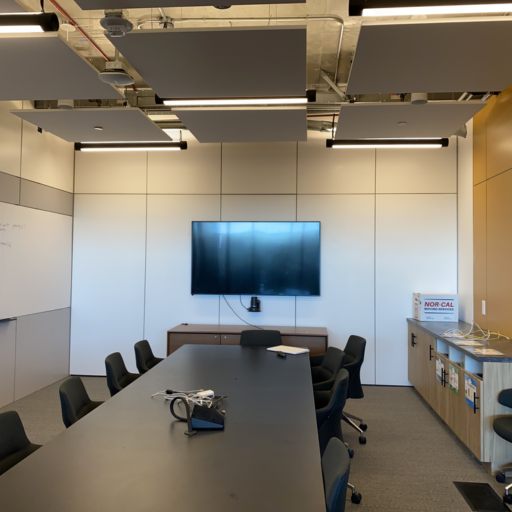} \\ 
    \end{tabular}
    \caption{\textbf{Qualitative Comparison on NeRF-Synthetic (256px, top) and LLFF (512px, bottom).} With more reference views, Kaleido demonstrates superior rendering precision compared to other generative baselines, with more accurate texture details and pose alignment. Furthermore, it avoids the representation-based artefacts sometimes present in the per-scene optimisation methods, highlighting the robustness of its learned, data-driven prior. }
    \vspace{-0.2cm}
    \label{fig:nvs_many}
 \end{figure}

When all available reference views are used, Kaleido's performance on the NeRF-Synthetic dataset is nearly on par with Instant-NGP. On the more complex LLFF and Mip-NeRF 360 scene datasets, Kaleido surpasses Instant-NGP, marking the first time a zero-shot generative model has matched the quality of a state-of-the-art, per-scene optimisation method. Qualitative comparisons are provided in Fig.~\ref{fig:nvs_many}.

Given that per-scene methods can be (very) sensitive to camera coordinate systems and sometimes fail to converge, Kaleido's robust, data-driven performance highlights the immense potential of zero-shot solutions for general-purpose rendering. A detailed analysis of the computational trade-offs between  generative and scene-specific methods is provided in Appendix~\ref{app:memory}.

\subsection{Results on Camera-Conditioned Video Generation}
\label{subsec:camera}

Kaleido leverages large-scale video data to improve novel view synthesis, a strategy that is conceptually related to recent camera-conditioned video generation models~\citep{he2025cameractrl,he2025cameractrlii,bahmanivd3d}. While Kaleido is designed for the distinct problem of ``any-to-any`` spatial view synthesis, there is an overlap in capabilities when the task is restricted to a specific setting: generating a video sequence along a {\it continuous target trajectory} from a {\it single} reference view. In this section, we compare Kaleido against video generation models within this shared domain.

\paragraph{Baselines and Datasets} In Table~\ref{tab:camera}, we compare Kaleido with a range of state-of-the-art camera-conditioned video models: \textbf{Wonderland}~\citep{liang2025wonderland}, \textbf{ViewCrafter}~\citep{yu2024viewcrafter}, \textbf{VD3D}~\citep{bahmanivd3d}, and \textbf{MotionCtrl}~\citep{wang2024motionctrl}.  We evaluate on two challenging in-the-wild datasets: \textbf{DL3DV-140}~\citep{ling2024dl3dv} and \textbf{Tanks and Temples}~\citep{knapitsch2017tandt}, strictly following the evaluation protocol described in Wonderland:
\begin{itemize}
    \item \textbf{DL3DV-140}: We randomly select 300 video clips from the test set. For each video, the first frame serves as the reference image, and the subsequent $n$ camera poses are used as conditions. We set $n=48$ to align with the Wonderland's setting.
    \item \textbf{Tanks and Temples}: We randomly sample 100 video clips from all 14 scenes using the same strategy.
\end{itemize}
Since both datasets lack dense pose annotations for every frame, we use COLMAP~\citep{schoenberger2016colmap1,schoenberger2016colmap2} to generate ground-truth poses for the source videos.

\paragraph{Evaluation Strategy} We evaluate performance using two criteria, following Wonderland's setting:
\begin{itemize}
    \item  \textbf{Camera-Guidance Precision:} We measure \textbf{Rotation error ($R_{err}$)} and \textbf{Translation error ($T_{err}$)}. To compute this, we estimate the camera poses of the generated videos using COLMAP, align the coordinate system relative to the first frame, and normalise the scale for comparison against the input condition.
    \item  \textbf{Visual Similarity:} We assess image quality using \textbf{PSNR, SSIM, and LPIPS} between the generated frames and the ground-truth views. For reliable comparison, we evaluate these metrics over the first 14 frames. This is due to the generated content naturally diverges from the ground truth as the scene progresses, rendering pixel-wise metrics less indicative of generation quality for longer sequences.
\end{itemize}

\paragraph{Results} As shown in the Table~\ref{tab:camera}, Kaleido clearly outperforms all camera-conditioned video models. The improvement is particularly significant in {\it camera precision}. Despite the baseline models being specialised for video generation along continuous trajectories, Kaleido has been shown to adhere to the target camera path with much higher accuracy. This highlights Kaleido's generality and superior rendering precision, even when applied to tasks outside its primary design scope.

\begin{table}[ht!]
    \centering
    \footnotesize
    \renewcommand{\arraystretch}{0.9}
    \setlength{\tabcolsep}{1.5em}
    \vspace{-0.1cm}
    \begin{tabular}{lccccc}
    \toprule
     {\it Dataset} & \multicolumn{5}{c}{Metrics} \\
     \cmidrule(l){2-6}
    Method & $R_{err}$\textdownarrow & $T_{err}$\textdownarrow & LPIPS\textdownarrow & PSNR\textuparrow & SSIM\textuparrow \\
    \midrule
     {\it DL3DV-140} \\
     MotionCtrl & 0.467 & 1.114 & 0.309 & 14.35 & 0.385 \\
     VD3D  & 0.094 & 0.237 & 0.259 & 16.28 & 0.487\\
     ViewCrafter & 0.092 & 0.243 & 0.237 & 17.10 & 0.519 \\
     Wonderland & 0.061 & 0.130 & \textbf{0.218} & 17.56 & 0.543 \\
     \textbf{Kaleido} & \textbf{0.011} & \textbf{0.026} & 0.232 & \textbf{18.09} & \textbf{0.544} \\
     \midrule
     {\it Tanks and Temple} \\
     MotionCtrl & 0.834 & 1.501 & 0.312 & 14.58 & 0.386 \\
     VD3D & 0.117 & 0.292 & 0.284 & 15.35 & 0.467 \\
     ViewCrafter & 0.125 & 0.306 & 0.245 & 16.20 & 0.506 \\
     Wonderland & 0.094 & 0.172 & 0.221 & 16.87 & \textbf{0.529} \\
     \textbf{Kaleido} & \textbf{0.016} & \textbf{0.086} & \textbf{0.197} & \textbf{17.97} & 0.528 \\
    \bottomrule
    \end{tabular}
    \vspace{-0.2cm}
    \caption{\textbf{Comparison with Camera-Conditioned Video Models.} We evaluate camera precision ($R_{err}$, $T_{err}$) and visual similarity (LPIPS, PSNR, SSIM) against video generation baselines. Kaleido significantly outperforms all video models in camera precision, indicating superior geometric consistency, while maintaining visual quality on par with the state-of-the-art.}
  \label{tab:camera}
    \vspace{-0.3cm}
\end{table}

\subsection{Results on 3D Reconstruction}
\label{subsec:3d_recon}

Given Kaleido's precise multi-view rendering capabilities, high-quality 3D reconstruction can be achieved by applying an off-the-shelf reconstruction framework to its generated views. In this section, we evaluate this capability on the \textbf{GSO-30} dataset. We compare Kaleido against a diverse set of generative models designed specifically for image-to-3D tasks. These include methods for direct 3D generation like \textbf{Point-E} (point clouds) \citep{nichol2022pointe} and \textbf{Shape-E} (NeRFs) \citep{jun2023shape}; optimisation-based methods like \textbf{DreamGaussian} \citep{tang2023dreamgaussian}; and view-synthesis-based methods like \textbf{One-2-3-45} \citep{liu2023one} and \textbf{SyncDreamer} \citep{liu2023syncdreamer}.

\paragraph{Evaluation Strategy} the evaluation protocol of SyncDreamer and EscherNet, we perform reconstruction by first using Kaleido to generate a set of views from pre-defined, object-centric camera poses, and then fitting a surface with an off-the-shelf surface reconstruction framework. Specifically, we adopt the camera setup from EscherNet, rendering 36 views by varying the azimuth from  0$^{\circ}$ to 360$^{\circ}$ (in 30$^{\circ}$ increments) at three fixed elevations (-30$^{\circ}$, 0$^{\circ}$, 30$^{\circ}$). These generated views then serve as input for the NeuS2 reconstruction \citep{wang2023neus2}. For a fair comparison, all baseline methods and Kaleido are evaluated at 256px resolution. We also provide results for Kaleido at 1024px resolution to showcase its high-resolution generation capabilities.

\begin{table}[ht!]
    \centering
    \footnotesize
    \renewcommand{\arraystretch}{0.9}
    \setlength{\tabcolsep}{0.6em}
    \begin{tabular}{lccccccccccc}
    \toprule
     & 
    \multicolumn{2}{c}{1 View} & 
    \multicolumn{2}{c}{2 Views} & 
    \multicolumn{2}{c}{3 Views} & 
    \multicolumn{2}{c}{5 Views} & 
    \multicolumn{2}{c}{10 Views} \\
    \cmidrule(lr){2-3} \cmidrule(lr){4-5} \cmidrule(lr){6-7} \cmidrule(lr){8-9} \cmidrule(lr){10-11}
     & CD\textdownarrow & VIoU\textuparrow & CD\textdownarrow & VIoU\textuparrow & CD\textdownarrow & VIoU\textuparrow & CD\textdownarrow & VIoU\textuparrow & CD\textdownarrow & VIoU\textuparrow\\
    \midrule
    Point-E        & 0.0447 & 0.2503 & --     & --     & --     & --     & --     & --     & --     & --     \\
    Shape-E        & 0.0448 & 0.3762 & --     & --     & --     & --     & --     & --     & --     & --     \\
    One-2-3-45        & 0.0667 & 0.4016 & --     & --     & --     & --     & --     & --     & --     & --     \\
    DreamGaussian  & 0.0459 & 0.4531 & --     & --     & --     & --     & --     & --     & --     & --     \\
    SyncDreamer    & 0.0400 & 0.5220 & --     & --     & --     & --     & --     & --     & --     & --     \\
    \midrule
    NeuS           & --     & --     & --     & --     & 0.0366 & 0.5352 & 0.0245 & 0.6742 & 0.0195 & 0.7264 \\
    EscherNet      & 0.0314 & 0.5974 & 0.0215 & 0.6868 & 0.0190 & 0.7189 & 0.0175 & 0.7423 & 0.0167 & 0.7478 \\
    \midrule
    \textbf{Kaleido} & 0.0214 & 0.6800 & 0.0120 & 0.7785 & 0.0113 & 0.7960 & 0.0104 & 0.8082 & 0.0100 & 0.8118 \\
    \textbf{Kaleido [1024 Res.]} & \textbf{0.0183} & \textbf{0.7006} & \textbf{0.0118} & \textbf{0.7851} & \textbf{0.0104} & \textbf{0.8053} & \textbf{0.0091} & \textbf{0.8290} & \textbf{0.0086} & \textbf{0.8418} \\
    \bottomrule
    \end{tabular}
    \caption{\textbf{3D Reconstruction Performance on GSO-30.} We measure reconstruction quality using Chamfer Distance (CD, lower is better) and Volumetric IoU (VIoU, higher is better). Kaleido clearly surpasses EscherNet by a large margin, demonstrating 5x greater view efficiency: Kaleido achieves a better reconstruction quality with just two views than EscherNet does with ten. The quality is further improved when using higher-resolution renderings from Kaleido.}
    \label{tab:3drecon}
\end{table}

\begin{figure}[t!]
    \centering
    \footnotesize
    \setlength{\tabcolsep}{0.6em}  
    \renewcommand{\arraystretch}{0.0}
    \begin{tabular}{C{0.2\textwidth} | C{0.18\textwidth} C{0.18\textwidth} C{0.18\textwidth} C{0.18\textwidth}}
    Referece Images & EscherNet & Kaleido & Kaleido [1024 Res.] & Ground-Truth \\
    \includegraphics[height=1\linewidth]{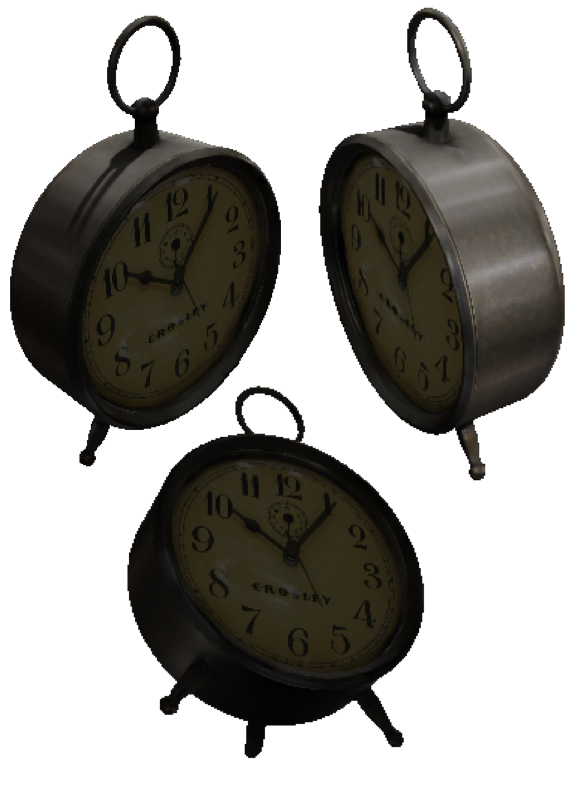} & 
    \includegraphics[trim={0cm 2cm 0cm 3cm}, clip, width=\linewidth]{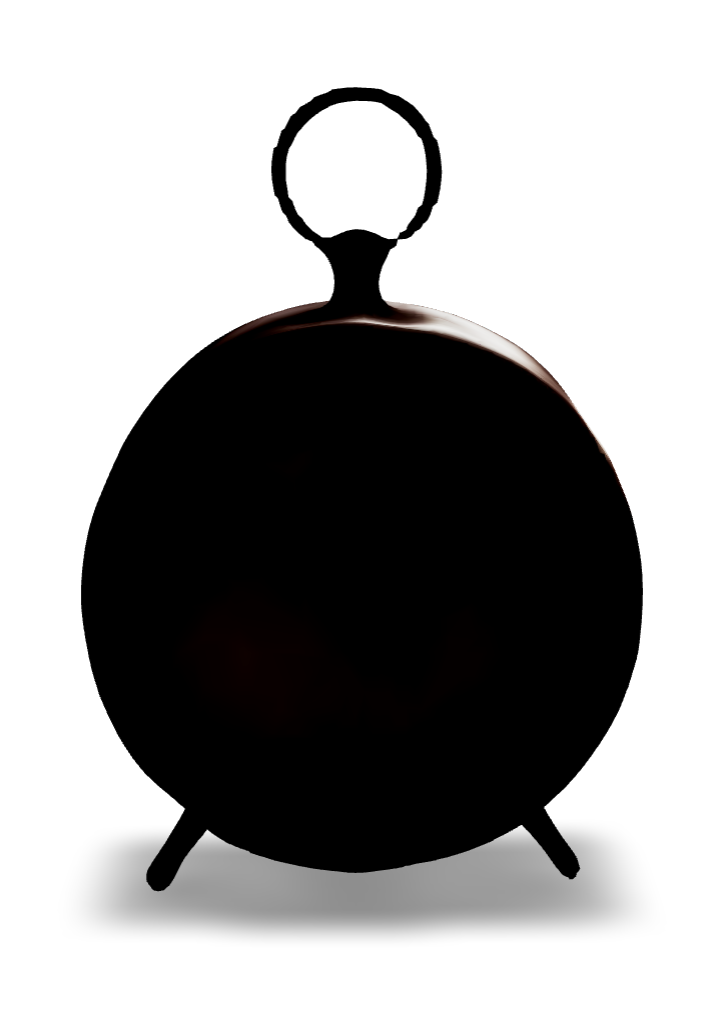} &
    \includegraphics[trim={0cm 2cm 0cm 3cm}, clip, width=\linewidth]{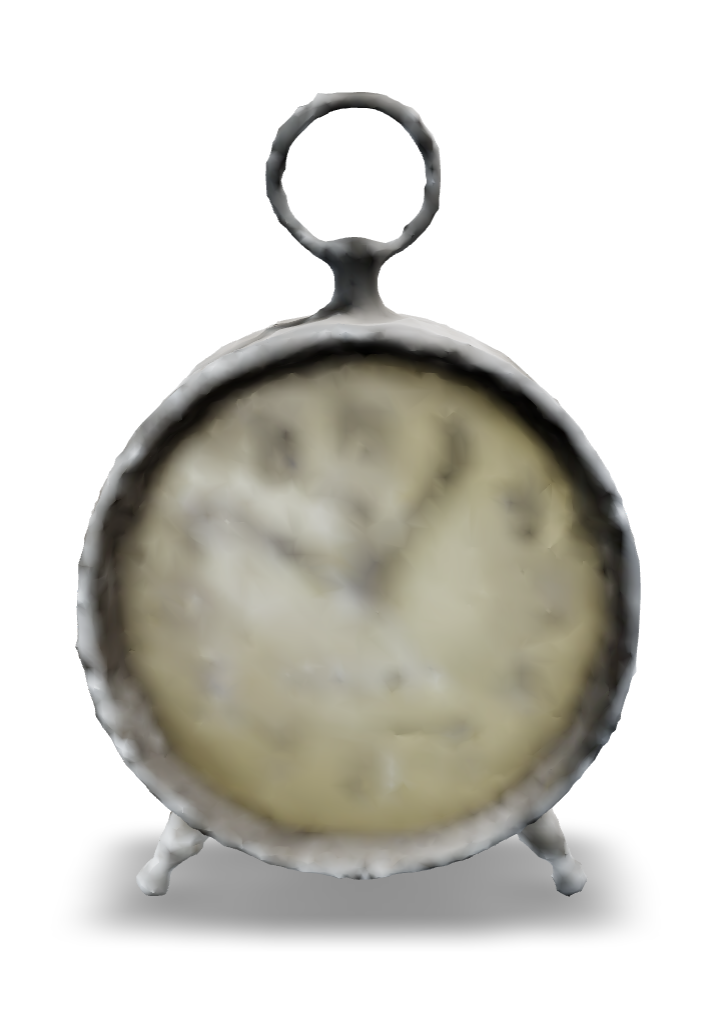} &
    \includegraphics[trim={0cm 2cm 0cm 3cm}, clip, width=\linewidth]{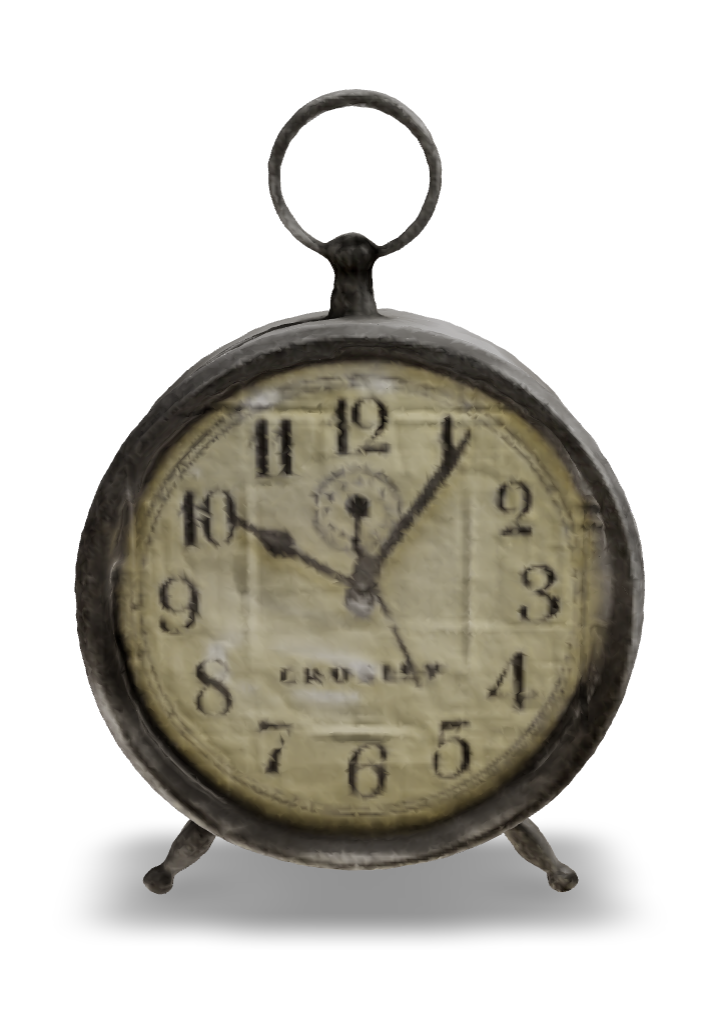} &
    \includegraphics[trim={0cm 2cm 0cm 3cm}, clip, width=\linewidth]{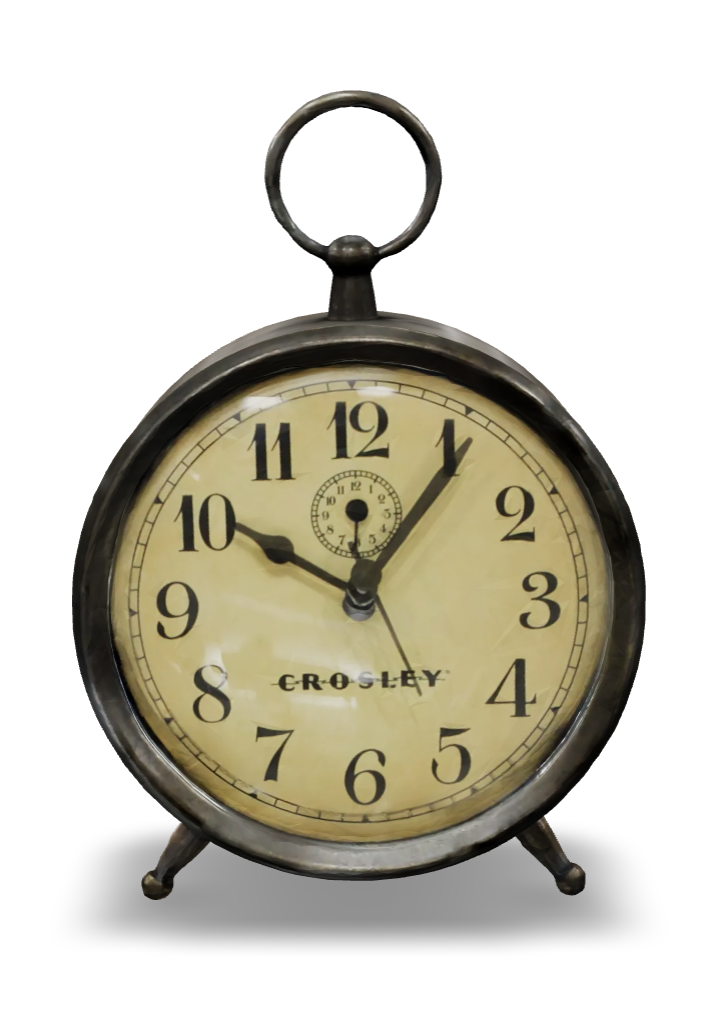} \\ 
    \includegraphics[height=\linewidth]{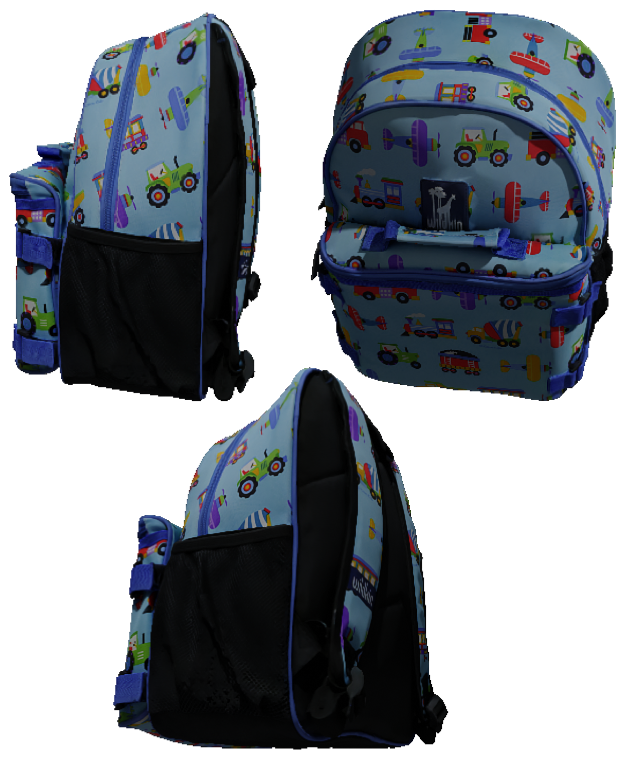} & 
    \includegraphics[trim={1cm 3cm 1cm 3cm}, clip, width=\linewidth]{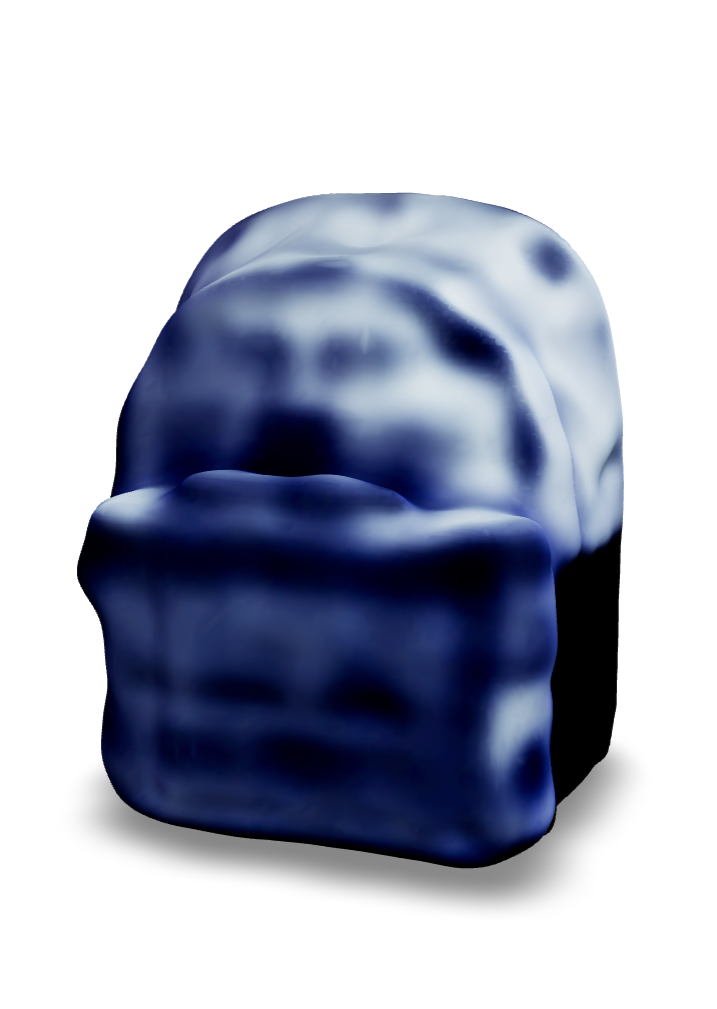} &
    \includegraphics[trim={1cm 3cm 1cm 3cm}, clip, width=\linewidth]{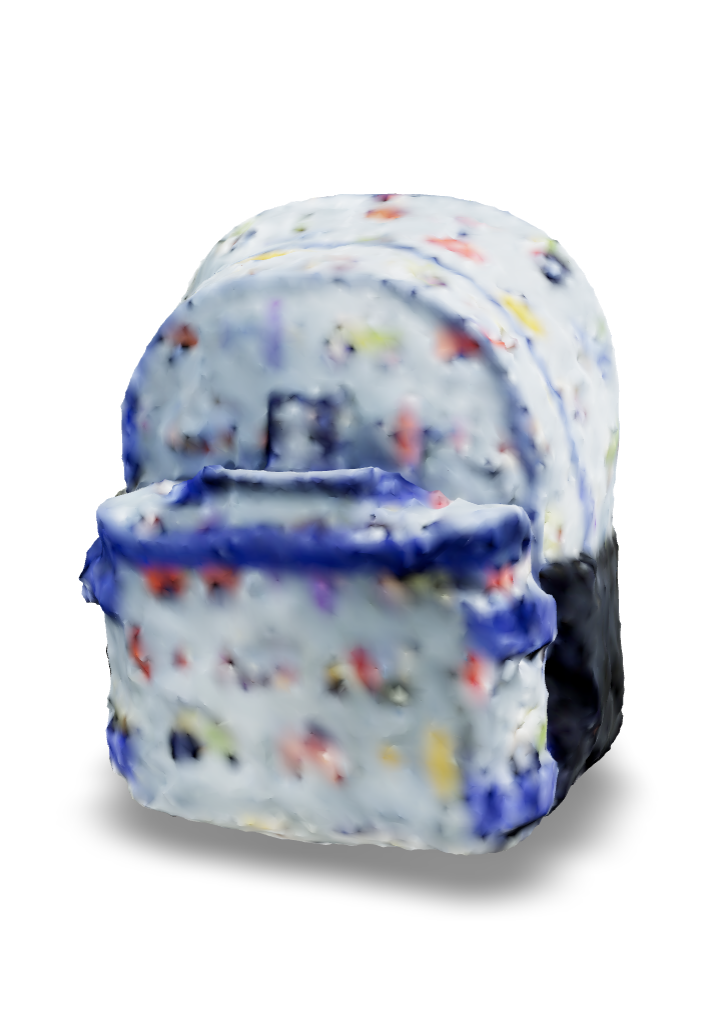} &
    \includegraphics[trim={1cm 3cm 1cm 3cm}, clip, width=\linewidth]{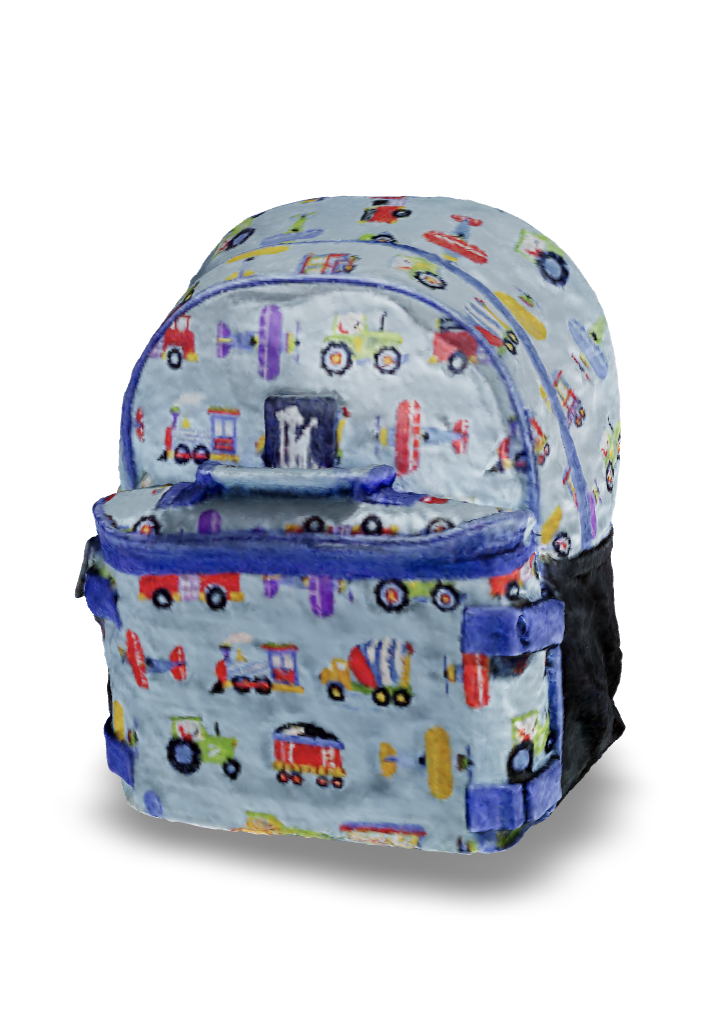} &
    \includegraphics[trim={1cm 3cm 1cm 3cm}, clip, width=\linewidth]{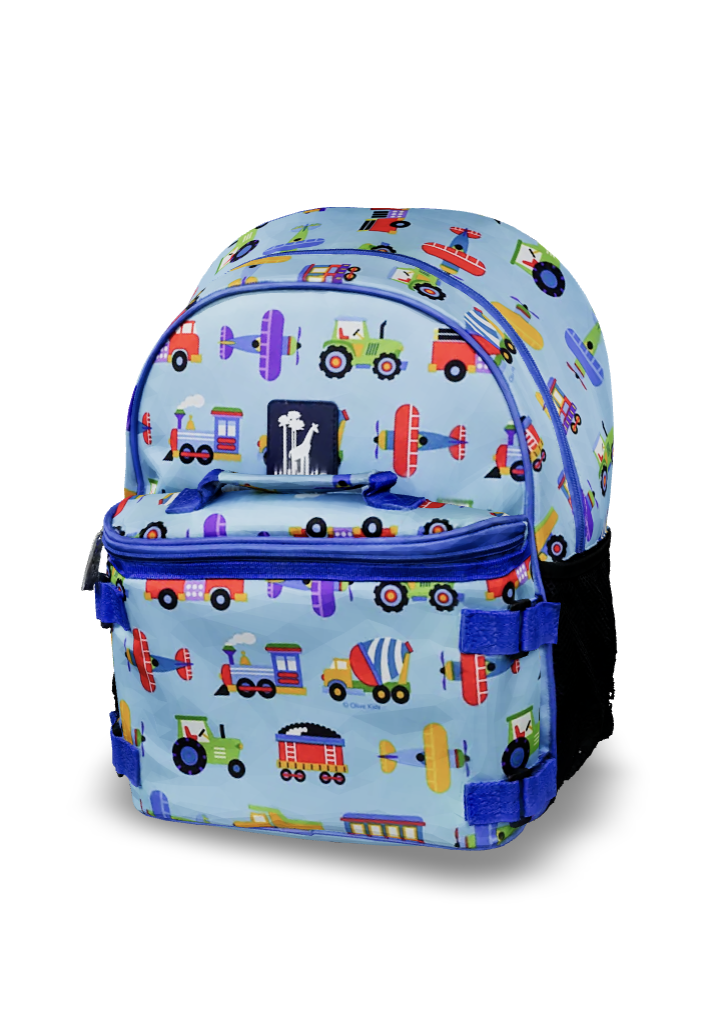} \\
    \includegraphics[height=\linewidth]{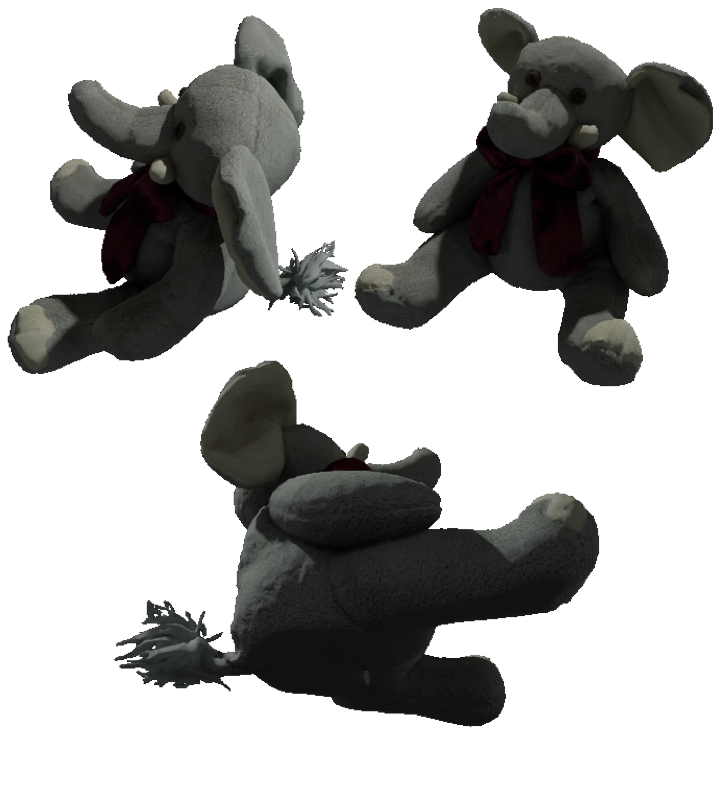} & 
    \includegraphics[trim={3cm 6cm 2cm 8cm}, clip, width=\linewidth]{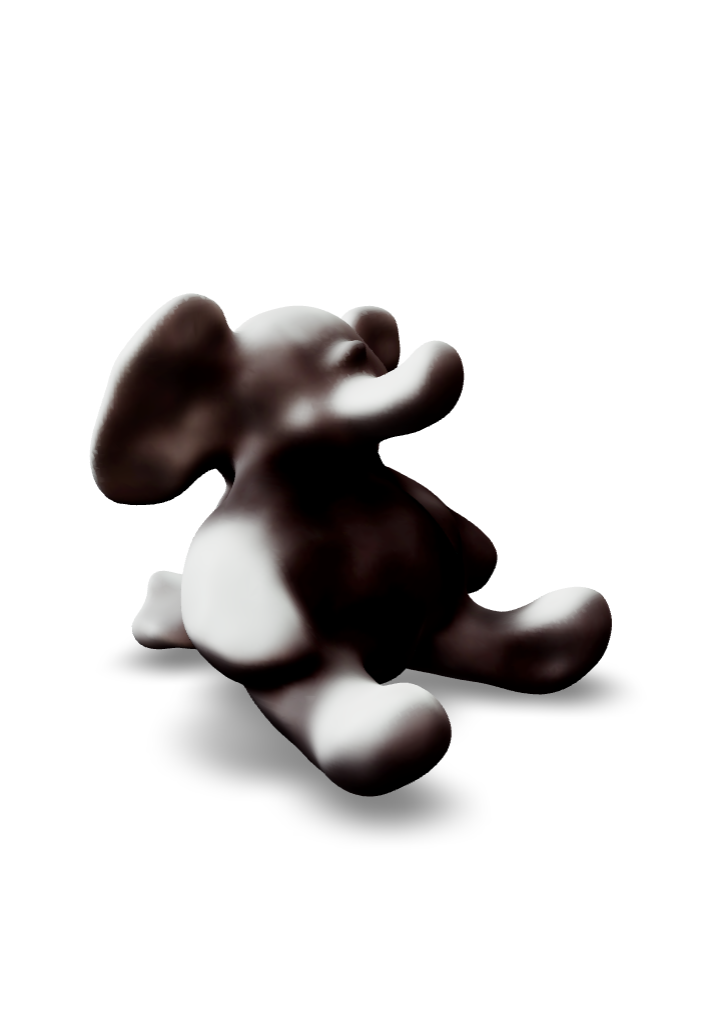} &
    \includegraphics[trim={3cm 6cm 2cm 8cm}, clip, width=\linewidth]{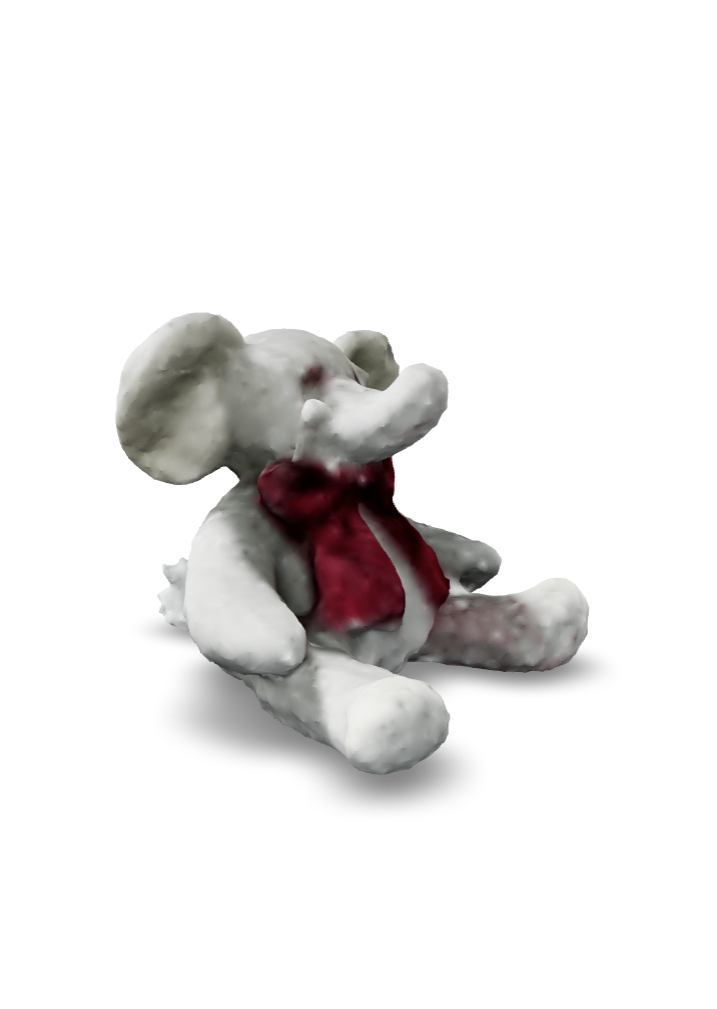} &
    \includegraphics[trim={3cm 6cm 2cm 8cm}, clip, width=\linewidth]{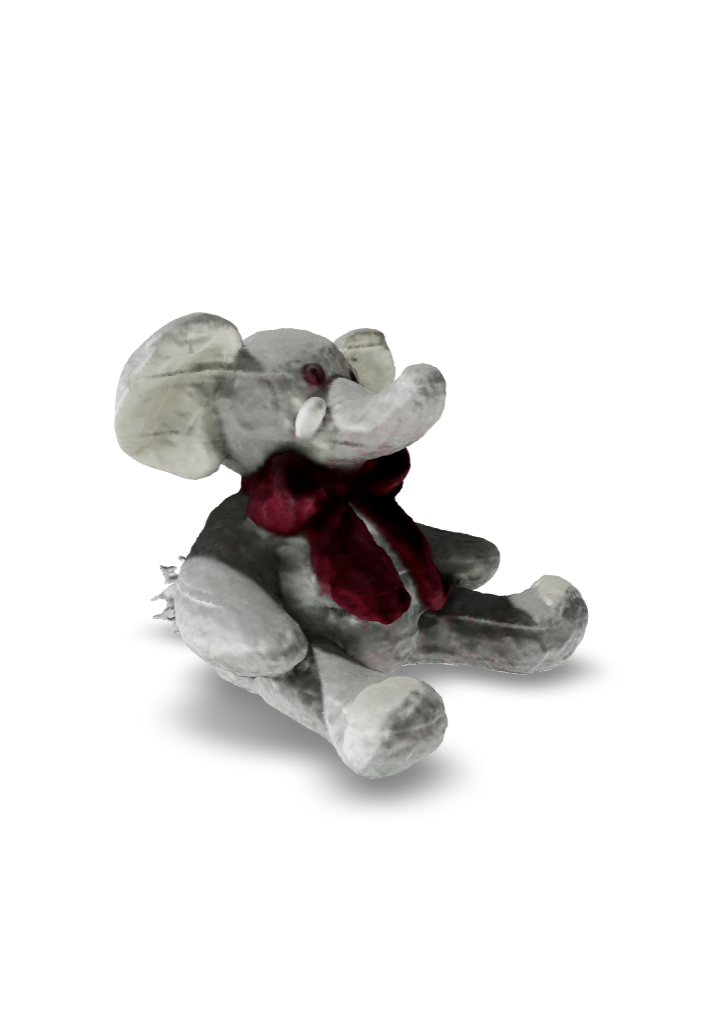} &
    \includegraphics[trim={3cm 6cm 2cm 8cm}, clip, width=\linewidth]{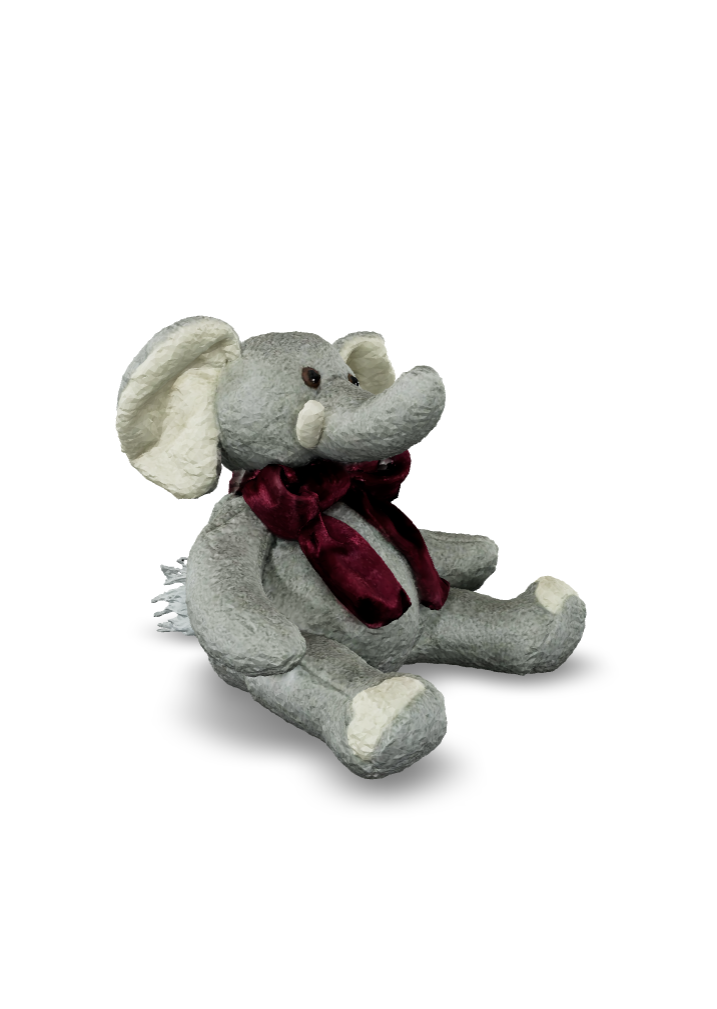} \\
    \includegraphics[height=\linewidth]{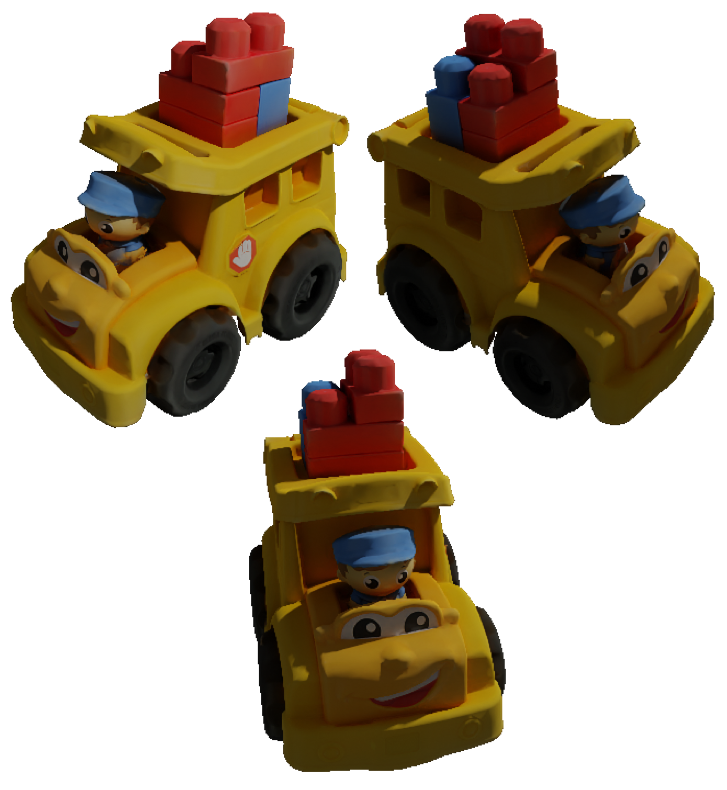} & 
    \includegraphics[trim={3cm 4cm 3cm 8cm}, clip, width=\linewidth]{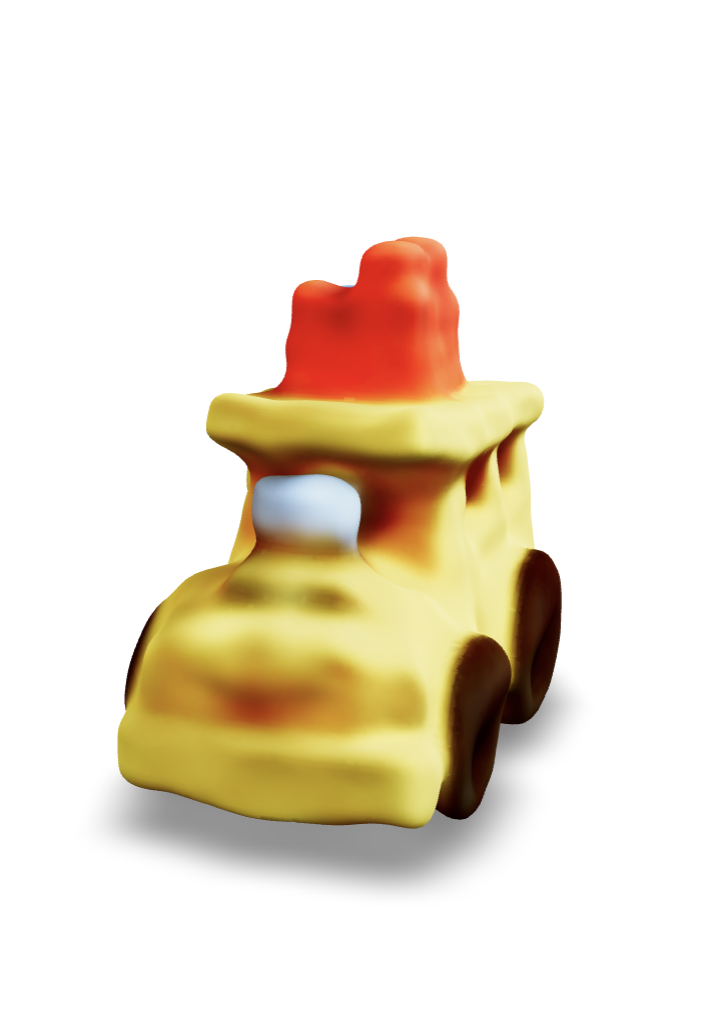} &
    \includegraphics[trim={3cm 4cm 3cm 8cm}, clip, width=\linewidth]{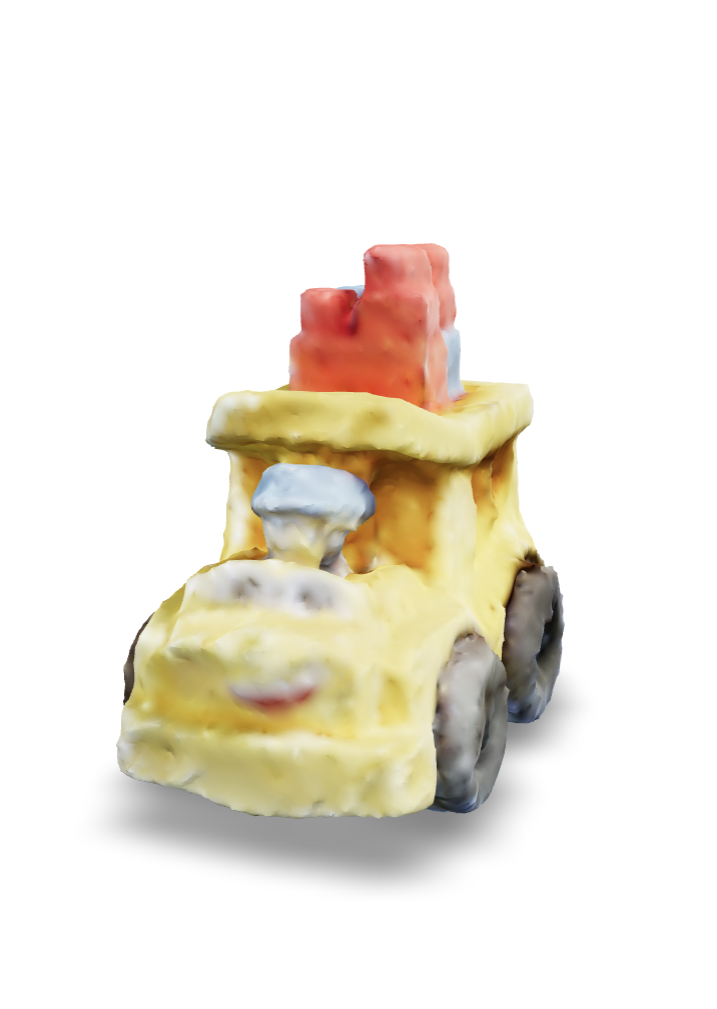} &
    \includegraphics[trim={3cm 4cm 3cm 8cm}, clip, width=\linewidth]{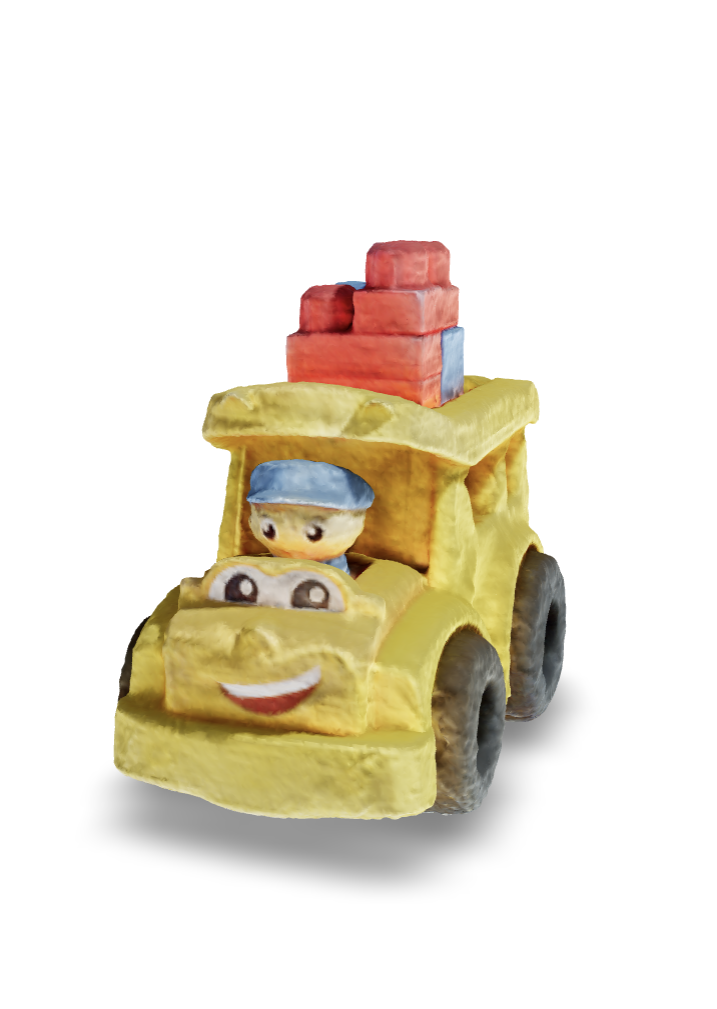} &
    \includegraphics[trim={3cm 4cm 3cm 8cm}, clip, width=\linewidth]{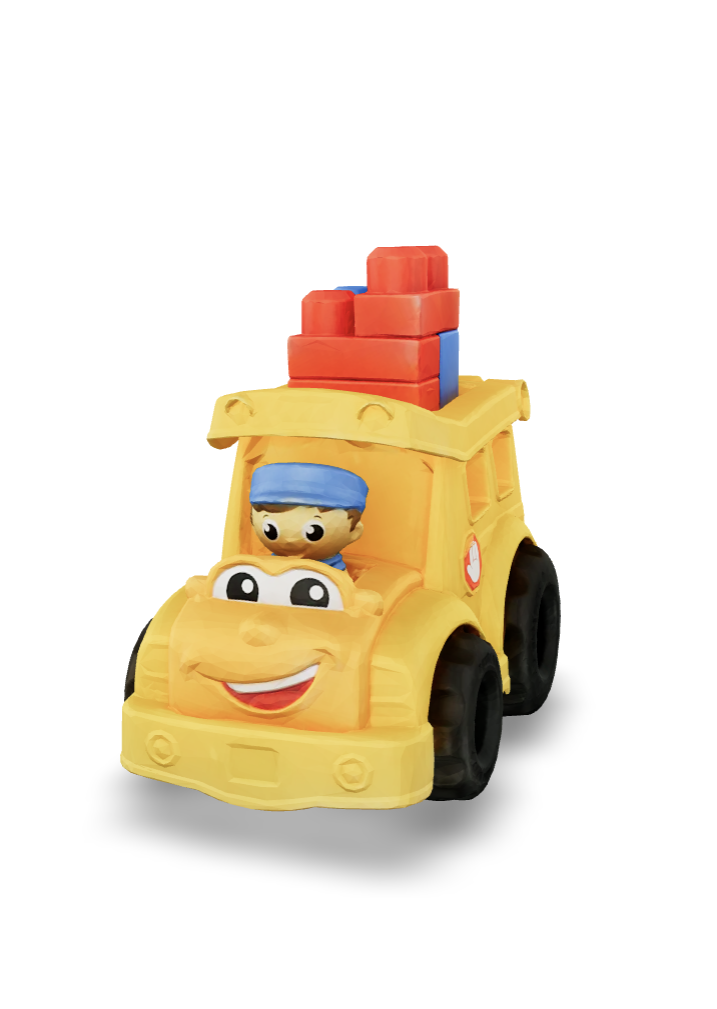} \\
    \end{tabular}
    \caption{\textbf{Visualisation of 3D Reconstructions with 3 Reference Views.}  Kaleido's precise renderings enable high-fidelity 3D mesh reconstruction using NeuS2. When leveraging 1024px renderings, the resulting meshes exhibit incredibly detailed textures, accurately capturing fine features like the numbers on the clock and the intricate patterns on the backpack.}
    \label{fig:3drecon}
 \end{figure}

\paragraph{Results} In Table \ref{tab:3drecon}, Kaleido again achieves state-of-the-art performance in 3D reconstruction, significantly outperforming direct image-to-3D models, our NeuS baseline, and EscherNet.\footnote{We also attempted to evaluate NeuS2 with the same limited input views, but the reconstruction failed to converge for most objects.} The results highlight Kaleido's remarkable rendering efficiency and precision. With just 2 reference views, our model has surpassed the reconstruction quality that EscherNet achieves with 10 views. This superiority is more evident qualitatively in Fig.~\ref{fig:3drecon}. Given the same 256px resolution, Kaleido's generated meshes are significantly better. At 1024px resolution, the reconstructed textures are incredibly detailed and sharp, appearing close to the ground truth and suggesting exciting new applications for high-fidelity, few-shot 3D reconstruction.

\clearpage
\newpage
\section{Conclusions, Limitations and Future Works}
In this paper, we introduced Kaleido, a new family of generative models that redefines neural rendering as a pure sequence-to-sequence problem, unifying 3D and video modelling. Through extensive ablations, we progressively modernised the architecture and training strategies, resulting in a model with exceptional rendering precision and spatial consistency.

Kaleido exhibits strong scaling properties and achieves state-of-the-art performance across a wide range of view synthesis and 3D reconstruction benchmarks. Most notably, it is the first generative rendering model to match the quality of per-scene optimisation methods in a zero-shot setting, representing a significant step towards a universal, general-purpose rendering engine.

Despite its strong performance, Kaleido has several limitations that open exciting avenues for future research:

\paragraph{Texture Flickering and Sticking.} In certain challenging scenarios, we observe two main types of visual artefacts in Kaleido's generations. Texture flickering can occur in scenes with high-frequency details (e.g., the LLFF Fern scene), particularly at lower resolutions or when conditioned on very few reference views, i.e. 1 view. We also occasionally observe texture sticking, where the generated sequence exhibits a non-continuous jump between frames. Improving spatial consistency in these most challenging settings remains an important direction for future work.

\paragraph{Fixed Camera Intrinsics.} Kaleido currently does not model camera intrinsics, which prevents it from generating effects like dolly-zooms, a capability present in models like SEVA \citep{zhou2025seva}. Future work could explore incorporating intrinsic parameterisation, potentially through another form of RoPE-based positional encoding designs \citep{li2025cameras}, to allow for more flexible camera control.

\paragraph{Degraded Generations with Large Viewpoint Changes.} While Kaleido often maintains excellent spatial consistency, its generated views can sometimes lack semantic plausibility when the viewpoint change is extreme. This suggests that while video pre-training builds a strong geometric foundation, it may not provide the diverse semantic knowledge required for high-fidelity single-image realism. Integrating priors from large-scale text-to-image/video models could be a promising direction to address this limitation.

\paragraph{Towards Faster Rendering.} Kaleido's generation time scales with the number of input views, and it is far from real-time. To fully bridge the gap with efficient and fast scene-specific methods like 3D Gaussian Splatting, future work will focus on improving inference speed through techniques like step distillation or architectural optimisations.

\paragraph{Towards 4D Generation.} Our unified positional encoding for space and time provides a natural foundation for true 4D generation. A promising future direction is to extend Kaleido to precisely control scenes across both space and time, enabling generative modelling of dynamic, four-dimensional worlds.

\section*{Acknowledgement}
We would like to thank Jianyuan Wang for his support on VGGT and for his insightful suggestions on our early paper draft; and to Takeru Miyato for his helpful discussions on GTA. We are also grateful to Tom Monnier for his support with the uCO3D and ShutterStock3D datasets, and to Zaiwei Zhang and Marc Mapeke for their support with the monocular depth and normal model benchmarking. Finally, we extend our gratitude to Mengmeng Xu for his assistance with training data preparation and filtering.

\newpage

\bibliographystyle{assets/plainnat}
\bibliography{references}

\newpage

\appendix
\section{Kaleido Design Ablative Quantitative Results}
\label{app:ablation}

Table~\ref{tab:comparison} presents the full quantitative results supporting the ablation study in Fig.~\ref{fig:ablation}, including all alternative design decisions we explored. To provide a comprehensive comparison, we report performance across three settings: one-to-five, five-to-one, and five-to-five reference-to-target views.

\begin{table}[ht!]
\centering
\small
\setlength{\tabcolsep}{0.2em}  
\renewcommand{\arraystretch}{0.9}
\begin{tabular}{lccccccc}
\toprule
 & \multicolumn{3}{c}{{\bf Objaverse}} & \multicolumn{3}{c}{{\bf uCO3D}} & \multirow{2}{*}{\makecell{Training \\  Throughput}} \\
 & $1\to5$ & $5\to1$ & $5\to5$ & $1\to5$ & $5\to1$ & $5\to5$ & \\
\cmidrule(r){1-1} \cmidrule(lr){2-4} \cmidrule(lr){5-7} \cmidrule(l){8-8}
Objaverse -  Single Baseline & 14.56 & 19.53 & 21.23 & - & - & - & - \\
uCO3D - Single Baseline & - & - & - & 14.89 & 19.19 & 16.97 & - \\
Objaverse + uCO3D - Joint Baseline & 12.17 & 19.10 & 20.02 & 14.66 & 18.71 & 16.77 & 160 \\ \midrule
\textbf{(i) Architecture Design [Vanilla DiT]} & & & & & & & \\
{\textcolor{red}{\bf DiT + Llama3 (SwiGLU + GQA)}} & 13.02 & 20.18 & 21.23 & 14.63 & 19.31 & 17.27 & 160 \\ \midrule
\textbf{(ii) Spatial Positional Encoding [2D RoPE + 3D CaPE]} & & & & & & & \\
RoPE + Plucker & 12.18 & 20.00 & 20.17 & 14.75 & 19.43 & 17.61 & 160 \\
{\textcolor{red}{\bf GTA [2D RoPE + 3D CaPE]}} & 11.93 & 21.05 & 22.03 & 13.65 & 20.84 & 17.54 & 148 \\ \midrule
\textbf{(iii) View Sampling Strategies [Fixed 6->6]} & & & & & & & \\
Uniform Sampling w/o Masking & 12.18 & 22.22 & 21.65 & 14.22 & 21.36 & 18.34 & 150 \\
Uniform Sampling w/ Masking & 14.51 & 21.00 & 20.22 & 14.58 & 20.37 & 17.60 & 150 \\
{\textcolor{red}{\bf Exponential Sampling w/ Masking}} & 15.13 & 21.41 & 21.11 & 15.16 & 20.25 & 17.83 & 148 \\ \midrule
\textbf{(iv) Temporal Attention Design [Temporal Attention (K=1)]} & & & & & & & \\
Full Attention & 14.54 & 22.62 & 22.40 & 15.00 & 20.75 & 18.28 & 58 \\
Temporal Window Attention (K = 2) & 15.45 & 21.60 & 21.04 & 15.40 & 20.43 & 18.15 & 146 \\
{\textcolor{red}{\bf  Temporal Window Attention (K = 4)}} & 15.67 & 22.25 & 21.79 & 15.55 & 20.81 & 18.45 & 142 \\
Temporal Window Attention (K = 8) & 15.73 & 22.60 & 22.54 & 15.85 & 21.39 & 19.15 & 103 \\ \midrule
\textbf{(v) Auxiliary Features [None]} & & & & & & & \\
{\textcolor{red}{\bf DiNOv2 [DiT-B]}} & 15.86 & 22.28 & 21.90 & 15.81 & 21.09 & 18.81 & 138 \\ 
DiNOv2 [DiT-L] & 16.34 & 22.65 & 22.43 & 15.94 & 21.24 & 18.75 & 135 \\ 
MetaDepth [DiT-L] & 15.82 & 22.28 & 22.22 & 15.76 & 21.26 & 18.65 & 135 \\ 
MetaNormals [DiT-L] & 15.77 & 22.32 & 22.11 & 15.59 & 21.20 & 18.85 & 135 \\ \midrule
\textbf{(vi) Timestep Conditioning Design \textcolor{red}{[AdaLN-Zero, Top 1 Act.: 15192] }} & & & & & & & \\
Shift Only [Top 1 Act.: 6820] & 16.07 & 21.56 & 21.51 & 15.97 & 20.56 & 18.28 & 144 \\ 
No Timestep [Top 1 Act.: 4312.] & 16.26 & 20.85 & 21.00 & 16.15 & 20.38 & 18.38 & 148 \\ \midrule
\textbf{(vii) Attention Registers [No Registers, Top 1 Act.: 15192]} & & & & & & & \\
\textcolor{red}{{\bf 1 Register [Top 1 Act.: 397.75]}} & 15.93 & 22.27 & 22.12 & 15.77 & 21.02 & 19.03 & 138 \\ 
8 Registers [Top 1 Act.: 279.75] & 15.26 & 22.04 & 21.94 & 15.72 & 20.75 & 18.55 & 138 \\ 
32 Registers [Top 1 Act.: 238.75] & 15.07 & 21.88 & 21.54 & 15.66 & 20.59 & 18.27 & 138 \\ \midrule
\textbf{(viii) Timestep Sampling Training Strategy [LogitNorm [0,1]]} & & & & & & & \\
Uniform [Shift = 1] & 17.58 & 22.01 & 21.96 & 15.90 & 20.49 & 18.57 & 138 \\ 
Uniform [Shift = 3] & 18.27 & 23.64 & 23.28 & 16.39 & 21.74 & 19.21 & 138 \\ 
Uniform [Shift = 5] & 18.43 & 23.38 & 23.06 & 15.90 & 21.42 & 18.94 & 138 \\ 
Mode [Scale = 0.8] & 17.39 & 22.06 & 22.08 & 15.99 & 20.75 & 18.68 & 138 \\ 
\textcolor{red}{{\bf Mode [Scale = 0.8, Shift = 3]}} & 18.19 & 24.06 & 23.75 & 16.03 & 21.76 & 19.11 & 138 \\ \midrule
\textbf{(ix) Timestep Sampling Inference Sampling [Linspace [1, 999]]} & & & & & & & \\
Trailing [1, 980] & 17.95 & 23.66 & 23.51 & 16.70 & 21.83 & 19.43 & 138 \\
\textcolor{red}{{\bf LinearQuadratic [1, 999]}} & 18.09 & 23.87 & 23.95 & 17.03 & 22.15 & 19.79 & 138 \\ \midrule
\textbf{(x) with Video Pre-training [No video Pre-training]} & & & & & & & \\
Video Pre-training 100K Steps (1.3x Eff.) & 18.16 & 24.22 & 24.30 & 17.11 & 22.23 & 20.10 & 138 \\
\textcolor{red}{{\bf Video Pre-training 200K Steps (2x Eff.)}} & 18.28 & 24.55 & 24.60 & 17.18 & 22.43 & 20.15 & 138 \\ 
\bottomrule
\end{tabular}
\caption{\textbf{Quantitative Results for Kaleido Design Ablations.} We report the complete quantitative results (PSNR, higher is better) corresponding to the ablation study in Fig.~\ref{fig:ablation}. Performance is evaluated in one-to-five, five-to-one, and five-to-five reference-to-target view settings. Our final design choice for each component is marked in red. }
\label{tab:comparison}
\end{table}
\newpage

\section{Additional Details of Kaleido Training Strategies}
\label{app:training}

All Kaleido model variants are trained using the same datasets detailed in Sec.~\ref{subsec:training}, with the AdamW optimiser \citep{loshchilov2018adamw} and a weight decay of $0.01$. In each training iteration, we randomly sample a total of 12 frames per sequence. These frames are then partitioned into reference and target views according to the view sampling strategy in Sec.~\ref{subsec:kaleido_design}.

The learning rate is chosen based on the training stage. For the initial video pre-training and the first stage of 3D fine-tuning (both at 256px resolution), we apply a learning rate of $10^{-4}$. For the subsequent high-resolution 3D fine-tuning stages (512px and 1024px resolution), we decrease the learning rate to $10^{-5}$. To train our larger models at high resolutions, we incorporate FSDP sharding and activation checkpointing.

Across all stages, we use {\tt fp16} mixed-precision training, as we find it crucial for stable training convergence; while {\tt bf16} consistently leads to unstable training. Our largest Kaleido model is trained for two weeks on 512 NVIDIA H100 GPUs. Additional hyper-parameters are listed in Table \ref{tab:kaleido_training}.

\begin{table}[ht!]
    \centering
    \footnotesize
    \begin{tabular}{lcccccccc}
    \toprule
     & \multicolumn{2}{c}{\textbf{Stage 1 (Video data)}} 
     & \multicolumn{2}{c}{\textbf{Stage 2 (3D data)}} 
     & \multicolumn{2}{c}{\textbf{Stage 3 (3D data)}} 
     & \multicolumn{2}{c}{\textbf{Stage 4 (3D data)}} \\ 
     & \multicolumn{2}{c}{[$256\times256$]} 
     & \multicolumn{2}{c}{[$256\times256$]} 
     & \multicolumn{2}{c}{[$512\times512$]} 
     & \multicolumn{2}{c}{[$1024$ mixed AR]} \\ 
     \cmidrule(lr){2-3} \cmidrule(lr){4-5} \cmidrule(lr){6-7} \cmidrule(lr){8-9}
     & Batch Size & \# Steps 
     & Batch Size & \# Steps 
     & Batch Size & \# Steps 
     & Batch Size & \# Steps \\
     \midrule
    \textbf{Kaleido-Small}  & 1024 & 700K & 1024 & 300K & 256 & 100K & 256 & 100K \\ 
    \textbf{Kaleido-Medium} & 1024 & 700K & 1024 & 300K & 256 & 100K & 256 & 100K \\ 
    \textbf{Kaleido}  & 2048 & 700K & 2048 & 500K & 256 & 100K & 256 & 100K \\ 
    \bottomrule
    \end{tabular}
    \caption{\textbf{Kaleido Training Pipeline.} Kaleido's training follows a multi-stage curriculum. The model is first pre-trained on a large-scale video dataset and is then fine-tuned on combined multi-view 3D datasets, with the image resolution progressively increased from 256px up to 1024px. In the final stage, we sample images with mixed aspect ratios to enable flexible resolution generation. Larger batch sizes are used for our largest Kaleido model to validate scaling laws.}
    \label{tab:kaleido_training}
\end{table}

\section{Additional Results for Few-shot View Synthesis}
\label{app:nvs_results}

We provide additional quantitative metrics for our few-view NVS benchmarks. Consistent with the PSNR results presented in Table~\ref{tab:few_shot_nvs}, Kaleido achieves state-of-the-art SSIM and LPIPS scores across all object- and scene-level datasets, confirming its superior generative rendering capabilities.

\begin{table}[ht!]
\centering
\scriptsize
\setlength{\tabcolsep}{0.33em}  
\renewcommand{\arraystretch}{0.9}
\begin{tabular}{lcccccccccccccccccccccc}
\toprule
 & {\bf OO3D} &\multicolumn{5}{c}{{\bf GSO-30}} & \multicolumn{5}{c}{{\bf RTMV}} & \multicolumn{2}{c}{{\bf LLFF}} & \multicolumn{3}{c}{{\bf Mip-NeRF 360}} & \multicolumn{4}{c}{{\bf Tanks and Temples}} \\
   \cmidrule(lr){2-2}  \cmidrule(lr){3-7} \cmidrule(lr){8-12} \cmidrule(lr){13-14}  \cmidrule(lr){15-17} \cmidrule(l){18-21}
\# Ref. Views & 1 & 1 & 2 & 3 & 5 & 10 &  1  & 2 & 3 & 5 & 10 & 1 & 3 & 1 &3 & 6  & 1 & 3 & 6 & 9 \\
\cmidrule(r){1-1} \cmidrule(lr){2-2}  \cmidrule(lr){3-7} \cmidrule(lr){8-12} \cmidrule(lr){13-14}  \cmidrule(lr){15-17} \cmidrule(l){18-21}
Eval. Data Type & Object & \multicolumn{5}{c}{Object} & \multicolumn{5}{c}{Multi-Object} & \multicolumn{2}{c}{Scene} & \multicolumn{3}{c}{Scene} & \multicolumn{4}{c}{Scene} \\
  Eval. Resolution & 512 & \multicolumn{5}{c}{256} & \multicolumn{5}{c}{256}  & \multicolumn{2}{c}{512} & \multicolumn{3}{c}{512} & \multicolumn{4}{c}{512}\\
  Eval. Tar. Views & 20 &\multicolumn{5}{c}{15}  & \multicolumn{5}{c}{10} & \multicolumn{2}{c}{5} & \multicolumn{3}{c}{27} & \multicolumn{4}{c}{35}\\
 SoTA Model & SV3D &\multicolumn{5}{c}{EscherNet} & \multicolumn{5}{c}{EscherNet}  & \multicolumn{2}{c}{SEVA} & \multicolumn{3}{c}{SEVA} & \multicolumn{4}{c}{SEVA} \\
 Results (LPIPS\textdownarrow) & 0.158 & 0.095 & 0.064 & 0.052 & 0.043 & 0.036 & 0.410 & 0.301 & 0.258 & 0.222 & 0.185 & 0.389 & 0.181  & 0.573 & 0.364 & 0.319 & 0.571 & 0.463 & 0.387 & 0.328\\
\cmidrule(r){1-1} \cmidrule(lr){2-2}  \cmidrule(lr){3-7} \cmidrule(lr){8-12} \cmidrule(lr){13-14}  \cmidrule(lr){15-17} \cmidrule(l){18-21}
\textbf{Kaleido-Small} & 0.144 & 0.123 & 0.061 & 0.043 & 0.029 & 0.019 & 0.332 & 0.204 & 0.166 & 0.130 & 0.095 & 0.323 & 0.152 & 0.528 & 0.376 & 0.318 & 0.549 & 0.449 & 0.385 & 0.328 \\
\textbf{Kaleido-Medium} & 0.126 & 0.094 & 0.048 & 0.034 & 0.023 & 0.015 & 0.329 & 0.181 & 0.145 & 0.109 & 0.080 & 0.315 & 0.127 & 0.473 & 0.347 & 0.290 & 0.508 & 0.437 & 0.359 & 0.302 \\
\textbf{Kaleido} & 0.121 & 0.086 & 0.044 & 0.030 & 0.021 & 0.013 & 0.289 & 0.171 & 0.137 & 0.105 & 0.074 & 0.301 & 0.123 & 0.530 & 0.344 & 0.286 & 0.541 & 0.465 & 0.363 & 0.288 \\
\midrule
 Results (SSIM\textuparrow) & 0.850 & 0.884 & 0.908 & 0.918 &  0.927 & 0.935 &  0.518 & 0.585 &  0.611 &   0.633 & 0.657 & 0.384 & 0.602 & 0.282 & 0.377 &0.395 & 0.342 &0.385 & 0.427 &0.452 \\
\cmidrule(r){1-1} \cmidrule(lr){2-2}  \cmidrule(lr){3-7} \cmidrule(lr){8-12} \cmidrule(lr){13-14}  \cmidrule(lr){15-17} \cmidrule(l){18-21}
\textbf{Kaleido-Small} & 0.873 & 0.867 & 0.919 & 0.938 & 0.954 & 0.969 & 0.584 & 0.670 & 0.703 & 0.746 & 0.800 & 0.341 & 0.574 & 0.221 & 0.313 & 0.362 & 0.313 & 0.359 & 0.403 & 0.444  \\
\textbf{Kaleido-Medium} & 0.880 & 0.885 & 0.933 & 0.948 & 0.963 & 0.975 & 0.591 & 0.697 & 0.731 & 0.778 & 0.827 & 0.359 & 0.645 & 0.271 & 0.347 & 0.410 & 0.351 & 0.359 & 0.419 & 0.459 \\
\textbf{Kaleido} & 0.884 & 0.895 & 0.938 & 0.954 & 0.966 & 0.978 & 0.610 & 0.704 & 0.738 & 0.781 & 0.836 & 0.375 & 0.659 & 0.248 & 0.361 & 0.433 & 0.333 & 0.368 & 0.429 & 0.479 \\
\bottomrule
\end{tabular}
\caption{\textbf{Zero-shot SSIM/LPIPS Performance with Generative Methods.} Kaleido achieves state-of-the-art performance across all object- and scene-level benchmarks, with SSIM and LPIPS metrics consistent with the superior PSNR performance reported in Table \ref{tab:few_shot_nvs}.}
\label{tab:few_shot_nvs_lpips}
\end{table}

\newpage
\section{Memory Consumption and Running Latencies}
\label{app:memory}

In this section, we compare the memory consumption and inference latency of Kaleido against the generative and deterministic baselines evaluated in Section \ref{subsec:exp_nvs}. In Table \ref{tab:memory}, the per-scene optimisation methods (Instant-NGP and 3DGS) operate in a distinct efficiency tier. They run significantly faster than generative approaches, as they do not require iterative denoising steps; 3DGS, in particular, achieves the fastest rendering speeds by not relying on the neural parameterisation entirely.

However, within the class of generative models, Kaleido demonstrates superior scaling efficiency. Notably, Kaleido-Medium is slightly larger than EscherNet in parameter count but scales much better with resolution (due to the efficient design of factorised spatial and temporal window attention), requiring only 10\% of the rendering time on 1024 resolution. While our current design prioritises rendering quality and model scalability, narrowing the inference speed gap with deterministic, scene-specific methods remains an important direction for future work.

\begin{table}[ht!]
    \centering
    \footnotesize
    \renewcommand{\arraystretch}{0.9}
    \setlength{\tabcolsep}{0.8em}
    \begin{tabular}{cccccccc}
    \toprule
     & 
    \multicolumn{5}{c}{Generative (sec. per frame / max frames per GPU)} & 
    \multicolumn{2}{c}{Deterministic (sec. per frame)} \\
    \cmidrule(r){2-6} \cmidrule(l){7-8}
     & \makecell{EscherNet\\ {[0.9B]}\\ (50 Steps)} & \makecell{SEVA\\ {[1.3B]}\\ (50 Steps)} & \makecell{Kaleido-Small\\ {[0.6B]}\\ (24 Steps)} & \makecell{Kaleido-Medium\\ {[1.2B]}\\ (24 Steps)} & \makecell{Kaleido\\ {[3.1B]}\\ (24 Steps)} & InstantNGP & 3DGS \\
    \midrule
    $[256\times 256]$ & 0.58s / 1601 & 0.42s / 1473 & 0.55s / 231 & 1.1s / 201 & 2.7s / 161 & 0.35s & 0.0025s \\
    $[512\times 512]$ & 6.4s / 401 & 1.5s / 369 & 1.6s / 119 & 3.1s / 103 & 4.6s / 76 & 0.35s & 0.0025s \\
    $[1024\times 1024]$ & 92s / 101 & 8.6s / 93 & 5.4s / 55 & 10.0s / 46  & 17.6s / 35 & 0.67s & 0.0034s \\
    \bottomrule
    \end{tabular}
    \caption{\textbf{Memory Consumption and Inference Latency Comparisons.} We report the averaged per-frame rendering time for a sequence of 25 frames (1 reference + 24 target views) across both generative and deterministic methods at various resolutions. For generative models, the reported times reflect the full inference process based on the required denoising steps specified in the original papers. To evaluate memory efficiency, we additionally report the maximum number of frames for all generative methods that can be generated concurrently on a single GPU. All benchmarks were conducted on a single NVIDIA A100.}
\label{tab:memory}
\end{table}

\end{document}